\documentclass[10pt,twocolumn,letterpaper]{article}

\usepackage{cvpr}              
\usepackage[accsupp]{axessibility} 

\usepackage[dvipsnames]{xcolor}

\definecolor{cvprblue}{rgb}{0.21,0.49,0.74}
\usepackage[pagebackref,breaklinks,colorlinks,citecolor=cvprblue]{hyperref}

\def\paperID{2305} 
\def\confName{CVPR}
\def\confYear{2024}

\title{Total Selfie: Generating Full-Body Selfies}

\author{Bowei Chen 
\qquad Brian Curless
\qquad Ira Kemelmacher-Shlizerman
\qquad Steven M. Seitz
\\
University of Washington\\
\\
{\tt\small \{boweiche, curless, kemelmi, seitz\}@cs.washington.edu}
}

\begin{document}
\twocolumn[{%
\renewcommand\twocolumn[1][]{#1}%
\maketitle
\begin{center}
    \centering
    \captionsetup{type=figure}
    \includegraphics[scale=0.55]{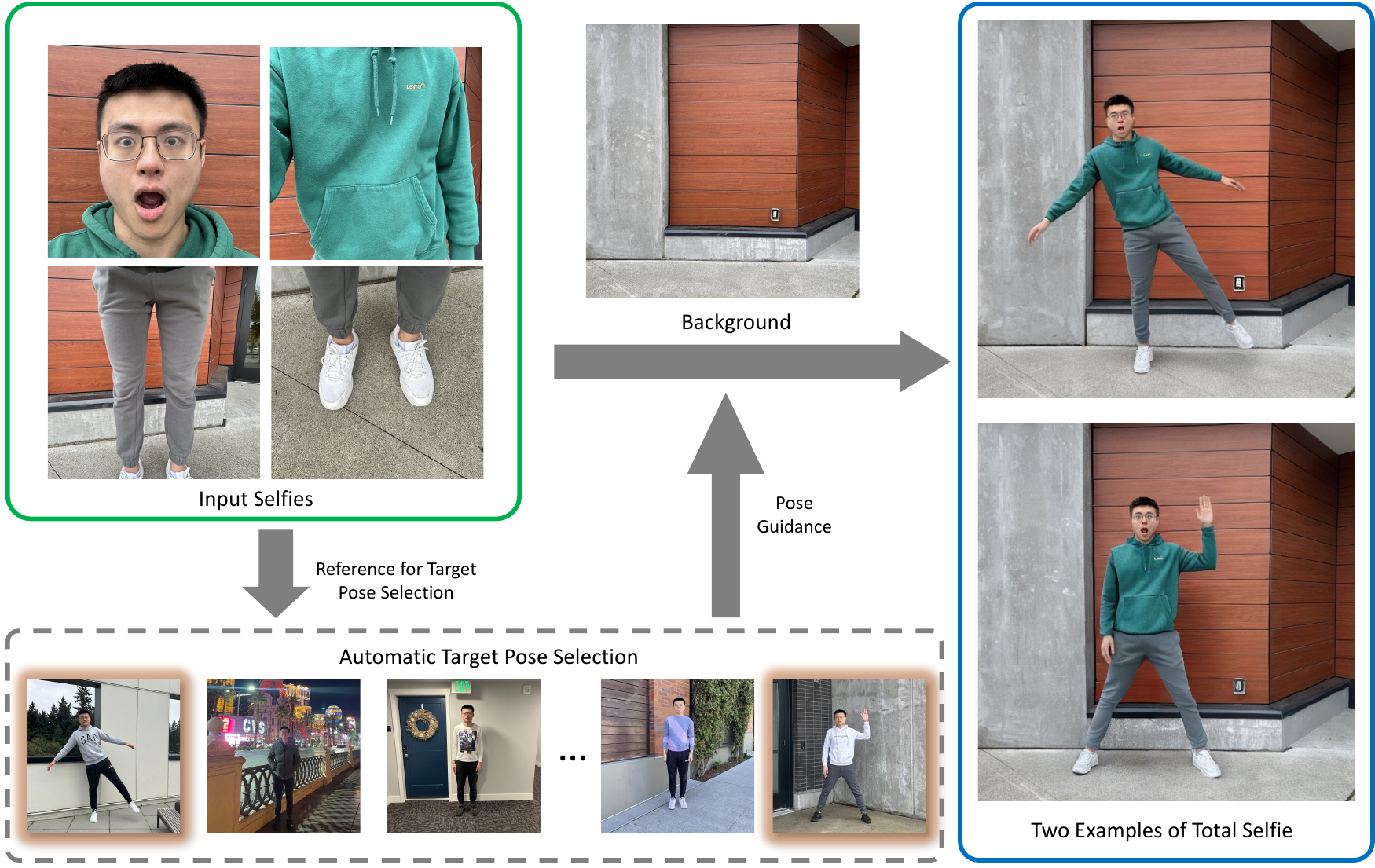}
        \captionof{figure}{We generate full-body selfies of you (right), from self-captured images of your face and body (top left) and background.  You can choose any target pose from a reference photo --- we auto-select a set of good candidates from your photo collection
        (bottom).         
        }
        \label{fig:teaser}
\end{center}%
}]

\begin{abstract}
We present a method to generate full-body selfies from photographs originally taken at arms length. Because self-captured photos are typically taken close up, they have limited field of view and exaggerated perspective that distorts facial shapes. We instead seek to generate the photo some one else would take of you from a few feet away.
Our approach takes as input four selfies of your face and body, a background image, and generates a full-body selfie in a desired target pose.
We introduce a novel diffusion-based approach to combine all of this information into high-quality, well-composed photos of you with the desired pose and background. 
\end{abstract}    
\section{Introduction}
\label{sec:intro}

The prevalence of selfies has skyrocketed in recent years, with an estimated 93 million taken each day. Despite their popularity, they suffer from multiple shortcomings:  (1) they capture only the upper portion of the subject, (2) the close-up camera viewpoint distorts faces and requires awkward poses (e.g., with arm reaching out), and (3) it is difficult to compose a shot that optimally captures both the subject and the scene.

Instead, what if you could capture the full-body image that {\em someone else would take of you} in the scene?  We call this a {\em total selfie}.  As input, we require four selfies to cover different parts of your body, and a photo of the background that you would like to be composited into (Fig.~\ref{fig:teaser}).  
Based on this information, we generate convincing full-body photos of you in a specified target pose in the desired scene (Fig.~\ref{fig:teaser} right).
In practice, we automatically select candidate reference photos from your photo collection, allowing you to choose one or more of them to determine the target pose. 






Solving this problem requires addressing a number of challenges.  First, we must render a complete and accurate image of your body, piecing together separate close-up images of your face, upper torso, legs, and shoes.
Second, we must reproject you to a virtual viewpoint from several feet away -- far enough to compose your full body within the scene.
And third, we need to render you with a desired target pose (where you're not holding the camera), which can be completely different from the one you used to take the selfies.  
The target pose can be specified by any full-body image from your photo collection.  To facilitate target pose selection, we auto-detect photos from your collection where you are wearing similar clothing to the input selfie set, leading to results with more accurate body shapes for a given type of clothing. 
 Most importantly, the resulting composite must retain your identity, expression, and clothing, but be composited realistically into the target scene with the desired pose and appearance.

One approach to this problem would be to collect a 
 paired dataset of selfies and full-body images of many people, 
 and train a generative model on it. 
However, acquiring such a dataset would be time and cost intensive.
Instead, we train a selfie to full-body model using a paired {\em synthetic} dataset, and further perform per-capture 
fine-tuning to narrow the gap between real and synthetic data.
Specifically, we first introduce a diffusion-based inpainting model trained in a self-supervised manner. This model takes four selfies as input and inpaints a full-body subject into a masked background.
Given a set of input images, we first remove perspective distortion of the face selfie, and then fine-tune the trained model on these images to enhance the fidelity of generated full-body photo further. 

Our contributions can be summarized as follows:
\begin{itemize}
    \item We introduce a novel type of self-captured photo -- {\em total selfie} -- that captures your entire body, as if some one was taking a photo of you in the scene.
    \item We propose a diffusion-based full-body generation model, followed by per-capture fine-tuning techniques, to generate {\em total selfie} from four selfies covering the body, a background image, and an auto-selected reference image as target pose.
    \item We demonstrate results for twelve individuals in various scenes (\eg, indoor and sunny outdoor) and clothing (\eg, skirts) with a wide range of poses and expressions.  Our experimental results outperform existing methods in generating realistic and accurate full-body images.
\end{itemize}


\section{Related Work}
\label{sec:related}


\noindent \textbf{Full-Body Image Generation.}
Extensive research has been dedicated to generating full-body images, either without specific conditions~\cite{fruhstuck2022insetgan,fu2022stylegan,weng2023diffusion,fu2023unitedhuman} or with conditions such as pose~\cite{yang20233dhumangan}, shape~\cite{sanyal2023sculpt}, or text prompts~\cite{jiang2022text2human}. 
One related area of research to our task is human reposing~\cite{ren2021flow,fang2023dance,knoche2020reposing,jain2023umfuse,wang2023disco,chen2023open,zhang2022pose,zhou2022casd,ren2022neural,Cui_2021_ICCV,lv2021learning,ren2020deep,karras2023dreampose,bhunia2022person,han2023controllable,li2023collecting,ma2023waveipt,sanyal2021learning}.
These methods transform a full-body (or partial-body) human image from one pose to another, with a target pose provided. 
For example, DisCo~\cite{wang2023disco} designed a diffusion-based framework to achieve this by using a person image (in any pose), a background image, and a target pose.
However, these methods are tailored for single image input and fall short when applied to our task due to the inherent limitation of a single selfie in capturing the full view of the person.


Another related line of work is Virtual Try-On~\cite{zhu2023tryondiffusion,chong2023fashion,lee2022high,xie2023gp,yan2023linking,li2023virtual,choi2021viton,chen2023size,he2022fs_vton}, where the goal is to generate a visualization of how a  garment might appear on a person, given the person image (in different clothing) and the garment image. For instance, LaDI-VTON~\cite{morelli2023ladi} introduced the first Latent Diffusion textual Inversion-Enhanced model to synthesize an image of the person wearing a specified clothing item. 
However, these approaches usually assume simple backgrounds, posing challenges in generating realistic shading within complex backgrounds. They also cannot alter footwear and facial expressions, crucial for our task.


Despite these challenges, both streams of work assume input from third-person view images, not selfies.


\noindent \textbf{Selfie-Related Techniques.}
Numerous studies have explored selfies for applications like reposing~\cite{ma2020unselfie}, face recognition~\cite{kumarapu2023wsd,botezatu2022fun}, style transfer~\cite{li2021adaptive,torbunov2023uvcgan}, novel view synthesis~\cite{bian2021nvss,park2021nerfies,park2021hypernerf,athar2023flame,kania2022conerf}, relighting~\cite{capece2019deepflash}, and video stabilization~\cite{yu2021real,yu2018selfie}.
For example, \cite{ma2020unselfie} proposed a coordinate-based method to transform a typical selfie, mainly focusing on the face, into a neutral-pose portrait. 
Nevertheless, their approach was limited to upper body selfies and could not generate full-body selfies capturing both the subject and the surroundings.
Our work is the first to propose and generate \textit{total selfie} from arm-captured selfies.

\begin{figure*}[!t]
    \centering
\includegraphics[scale=0.53]{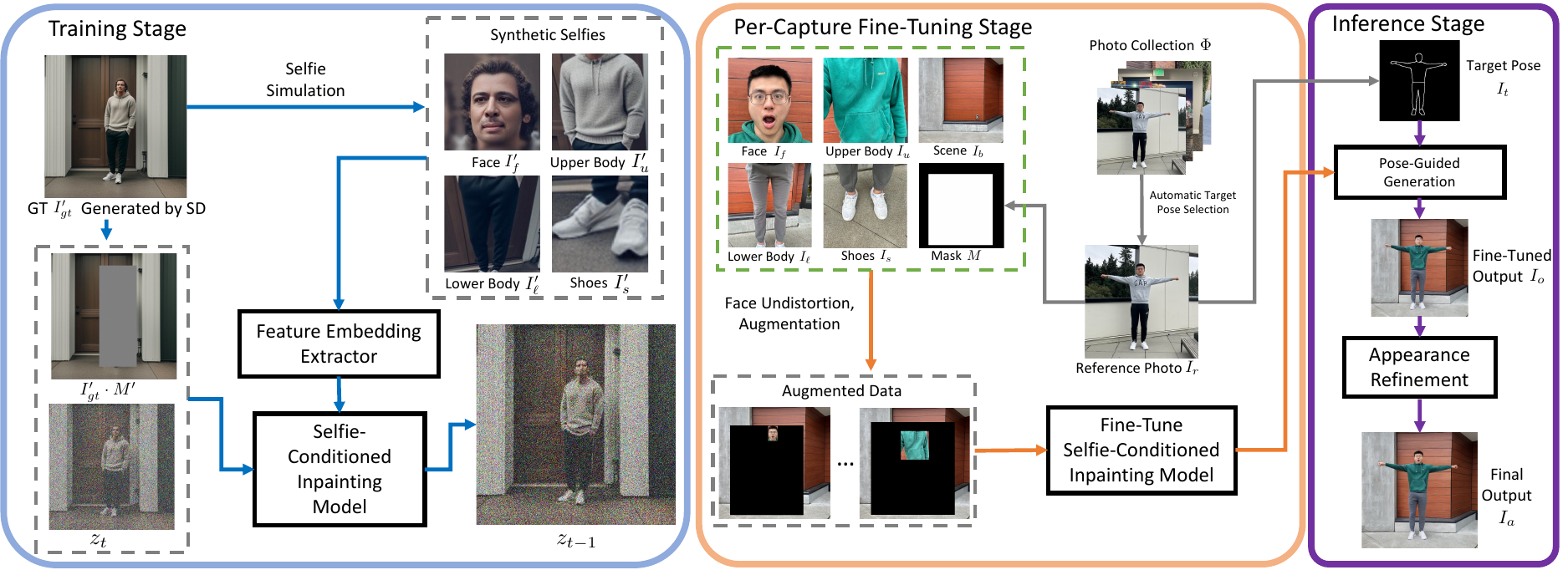}
    \caption{Overview of Total Selfie.  First, we train a selfie-conditioned inpainting model based on  a synthetic selfie to full-body dataset (blue box). Second, we fine-tune the trained model on a specific capture (orange box), and use it to produce a full-body selfie with the help of modified ControlNet (for pose) and appearance refinement (for face and shoes), visualized in the purple box.  Note that,  images in the green dashed box (inside the orange box) serve as input and conditional signals to the inpainting model, arrows omitted for simplicity.
    }
    \vspace{-2mm}
    \label{fig:overview}
\end{figure*}

\noindent \textbf{Diffusion Models.}
Diffusion models have recently demonstrated their success in various tasks such as text-to-image~\cite{rassin2022dalle,saharia2022photorealistic,Rombach_2022_CVPR} and image-to-image translation~\cite{yang2022paint,han2023svdiff}. 
DreamBooth~\cite{ruiz2022dreambooth} was proposed to personalize a text-to-image model by fine-tuning the model on a few reference images.
 RealFill~\cite{tang2023realfill} introduced an image completion technique to outpaint an input photo using reference images from the same scene. It fine-tuned a text-to-image inpainting model using reference images and applied it to complete the input photo. 
However, RealFill assumed third-person view images as input and mainly focused on generating scene content, rather than full-body subjects.
Another relevant work, Paint-By-Example~\cite{yang2022paint}, introduced an image-conditioned inpainting model to inpaint masked scenes with content specified in a reference photo. Similarly, we frame our problem as an exemplar-based inpainting problem. Thus we adapt  this model to suit our  settings.

\section{Total Selfie}
\label{sec:method}

We refer to our pipeline as Total Selfie, and define our task setting more formally.
As input, a user captures four selfies, including face $I_f$, upper body $I_u$, lower body $I_\ell$, and shoes $I_s$, as well as the background image $I_b$.
Total Selfie inpaints the full-body individual into $I_b$, with the target pose $I_t$ and inpainting mask $M$ (where 1 indicates the region to be inpainted) specified by a reference image $I_r$. 
Here, $I_r$ is automatically selected from the user's photo collection $\Phi$. 


Total Selfie has two main steps.
As depicted in Fig.~\ref{fig:overview} left, we first generate a large paired dataset comprising four selfies as input and a corresponding full-body image as ground truth. Then we train a selfie-conditioned inpainting model on this dataset. 
Given an input capture, we perform preprocessing on the input images, including face undistortion and automated pose selection.  These preprocessed images are then used to fine-tune the trained model (Fig.~\ref{fig:overview} middle). The fine-tuned model is employed to generate an initial output $I_o$, which is further refined to produce the final output $I_a$ (Fig. \ref{fig:overview} right).

\subsection{Training Selfie-Conditioned Inpainting Model}
\label{sec:training}
Training the selfie-conditioned inpainting model involves two steps:
(1) generating a large paired dataset and (2) training an 
 image-conditioned diffusion model on this dataset. 

\noindent \textbf{Dataset Generation}.  
We define one training pair as $\{(S', I'_{gt} \cdot M', M'), I'_{gt}\}$, where $S'=\{I'_f, I'_u, I'_\ell, I'_s\}$ is a set of four synthetic selfies for face, upper body, lower body, and shoes respectively. $I'_{gt}$ is the ground-truth full-body image, and $M'$ is the inpainting mask.

To create a pair, we start by generating $I'_{gt}$ using Stable Diffusion~\cite{Rombach_2022_CVPR}, with the pose guided by OpenPose ControlNet~\cite{zhang2023adding}.  
The person's bounding box in $I'_{gt}$ is then scaled up (following \cite{yang2022paint}) to generate the inpainting mask $M'$. We choose a bounding box representation instead of a more detailed shape for the mask since we anticipate changes in nearby regions, like those affected by shadows, when integrating the individual into the scene. 
Simulating selfie set $S'$ from the third-person view $I'_{gt}$ is non-trivial. 
One possible idea is to estimate 3D geometry of $I'_{gt}$ using depth estimation~\cite{Ranftl2022,bhat2023zoedepth,birkl2023midas} or human reconstruction~\cite{xiu2023econ} methods, and  then re-render it from the perspective of a selfie camera.  However, this is not practical due to inaccuracy of the estimated 3D geometry. 
Instead we propose a simpler yet effective way for obtaining $S'$ using homography transformation.
To create $I'_u$, we follow these steps.
(1) Identify the \textit{typical} positions of upper body OpenPose keypoints from pre-captured real upper body selfies (see supplementary). 
(2) Detect the upper body OpenPose keypoints from $I'_{gt}$.
(3) Compute a homography transformation that maps keypoints in $I'_{gt}$ to  their typical positions.
(4) Use this transformation to warp $I'_{gt}$ and obtain $I'_u$.
We apply the same approach to generate $I'_\ell$ and $I'_s$. 
Finally, we use FFHQ face alignment~\cite{karras2019style} to extract the face selfie $I'_f$.
The resultant selfie set $S'$ is visualized in the top right part of Fig. \ref{fig:overview} blue box.
We repeat this process to create multiple training pairs with $I'_{gt}$ in diverse poses.
Although the selfie simulation process primarily focuses on viewpoint correction and does not  consider pose differences (\eg, using one arm to hold camera), we observed that this is sufficient for training an inpainting model and obtaining a reasonable selfie-to-full-body prior.

\noindent \textbf{Training}. 
We start with an existing image-conditioned, diffusion-based inpainting model called Paint-By-Example~\cite{yang2022paint} and tailor our specific task. We refer to our adapted model as  selfie-conditioned inpainting model.


To train our model, we first apply the forward diffusion process to $I'_{gt}$. 
Starting from a clean latent $z_0 = E(I'_{gt})$, the forward process yields a noisy latent $z_t = \alpha_t z_0 + (1-\alpha_t) \epsilon$, where $E$ is the encoder of Variational AutoEncoder used in Latent Diffusion~\cite{Rombach_2022_CVPR}.  Here, $t$ is a randomly sampled timestep, $\epsilon$ represents Gaussian noise, and $\alpha_t$ is a weight parameter determined by $t$.
We then use a diffusion model 
to denoise $z_t$ and update the model parameters by minimizing the following loss function:
\begin{equation}
\mathcal{L} = \mathbb E_{t,z_0,\epsilon} || \epsilon_\theta (z_t, I'_{gt} \cdot M', c', t) - \epsilon||_2^2,
\label{eq:loss}
\end{equation}
where $c'$ is the condition for the diffusion model, which, in our case, is the image embeddings extracted from $S'$.
In Paint-By-Example, $c'$ is implemented as $L\cdot F(I)$, where $I$ is a single image, $F(\cdot)$ is a pretrained CLIP image encoder~\cite{radford2021learning} followed by a multi-layer perceptron (MLP), and $L(\cdot)$ is a linear layer. 
 In our model, we extend this idea and implement it as:
 \begin{equation}
     c'=L([F(I'_f), F(I'_u), F(I'_\ell), F(I'_s)]), 
     \label{eq:c}
 \end{equation}
 where $[\cdot]$ is the operation of concatenation. $L$ is modified to adapt to the dimension of the concatenated embeddings.
 One advantage of $F(\cdot)$ is that it transforms an image into a highly compressed vector. This forces the network to understand the clothing type and color in $S'$, preventing it from reaching optimal results by simply applying homography transformation to $S'$ in training.
 The training process is visualized in the blue box in Fig. \ref{fig:overview}.
In practice, we train our model by updating parameters ($\theta$, weights in $F$'s MLP and $L$) using Eq.~\ref{eq:loss}. We initialize weights (excluding those in layer L) with pretrained weights from Paint-By-Example.
This allows us to leverage the generative prior in this pretrained model and significantly reduce our training time.

\subsection{Per-Capture Preprocessing and Fine-Tuning}
We start by presenting two test time preprocessing steps: \textit{face undistortion} and \textit{automatic target pose selection}.  Then we introduce how to generate results aligning with the target pose, and discuss the issue of generated results to motivate the per-capture fine-tuning.


\begin{figure*}[!t]
\centering
  \subcaptionbox{ Input Selfies}%
{\includegraphics[width=0.195\linewidth]{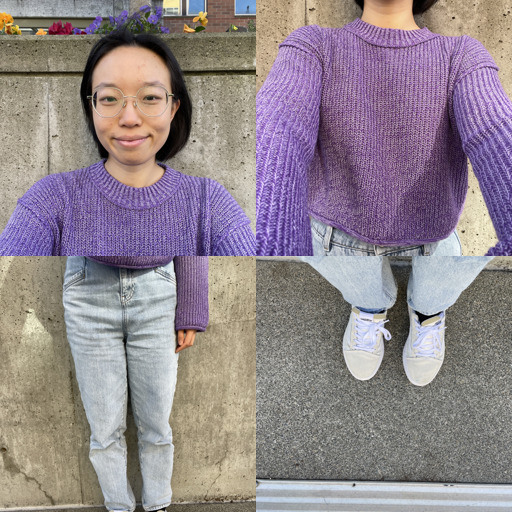}}
  \subcaptionbox{ Target Pose}%
{\includegraphics[width=0.195\linewidth]{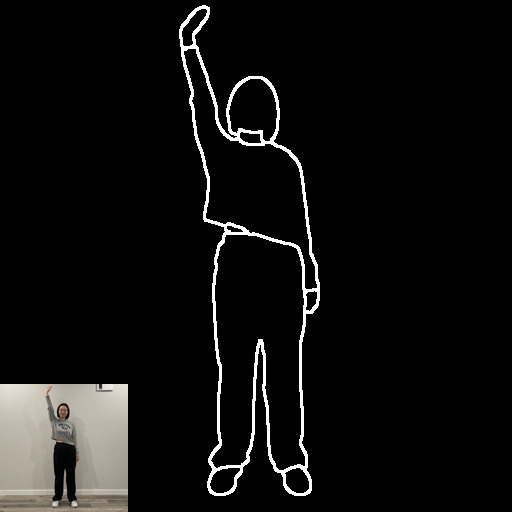}}
    \subcaptionbox{Output w/o Fine-Tuning }%
{\includegraphics[width=0.195\linewidth]{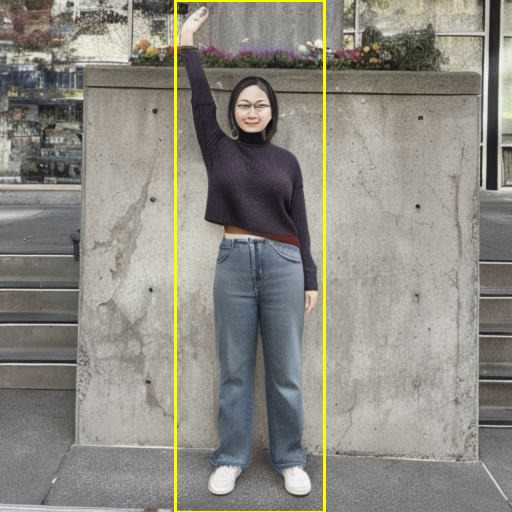}}
    \subcaptionbox{Fine-Tuned Output}%
{\includegraphics[width=0.195\linewidth]{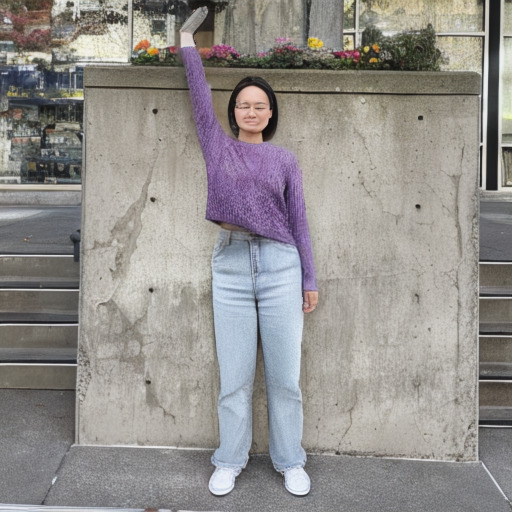}}
    \subcaptionbox{Full Pipeline}%
{\includegraphics[width=0.195\linewidth]{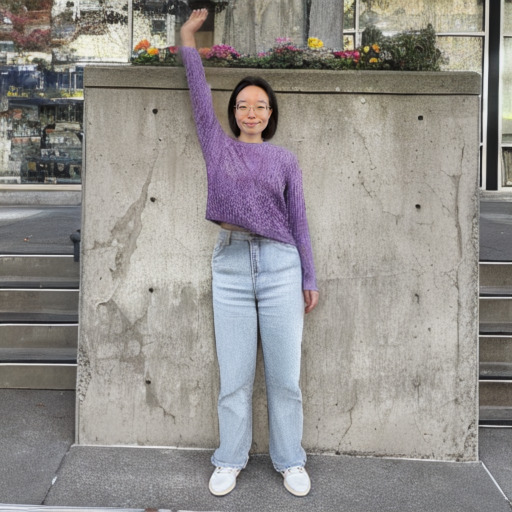}}
    \vspace{-2mm}
\caption{Results for different modules of our pipeline. 
Background image omitted due to space; regions inside bounding box (c) are to be inpainted.
The Canny Edge image in (b) is detected from the reference image, inset.  Generating without fine-tuning (c) produces inaccurate outfit and identity.   Through fine-tuning, the pipeline (d) generates correct outfit with reasonable shading and clothing details (\eg, wrinkles on upper cloth), but with wrong identity.  With appearance refinement, the full pipeline (e) yields  high-quality full-body selfies.   
}
\vspace{-4mm}
  \label{pipeline_ablation}
\end{figure*}

At test time, the set of user-captured selfies $S=\{I_f, I_u, I_\ell, I_s\}$ has a different distribution from the simulated selfies $S'$ in the training set. 
We note two particular differences: (1) real selfie $I_f$ often exhibits significant face distortion, which is not present in the simulated selfie $I'_f$ (obtained from full-body image),
 (2) upper body selfies $I_u$ and $I'_u$ have different poses, arising from the need to hold the camera out front (using an upper body arm) when taking a real selfie (see Fig. \ref{fig:overview}). 
To address the first issue, we propose a face undistortion strategy as a  preprocessing step to help reduce the domain gap. We address the second issue, pose differences, with a fine-tuning step which we discuss later, as it is non-trivial to resolve during preprocessing.

\noindent\textbf{Face Undistortion}.
Existing methods alleviate selfie distortion by either optimizing on a single image~\cite{wang2023disco_undis,shih2019distortion} or by training a model on a combination of an  unrealistic synthetic dataset and a small real dataset~\cite{zhao2019learning}. 
  For test-time efficiency, we follow the latter idea. We render a large paired dataset using a method that generates realistic, textured 3D heads using 3D GANs~\cite{Chan2021}. 
 Then we fine-tune a talking-head synthesis network~\cite{wang2021facevid2vid} to perform perspective undistortion using the rendered dataset.  
 See supplementary for more details and results.
Additionally, we roughly align the shoes selfie $I_s$ with $I'_s$ by cropping it based on the bounding box of the shoes.
For simplicity,  we reuse $I_f$ and $I_s$ to represent corrected face selfie  and cropped shoes selfie, respectively. 

\noindent \textbf{Automatic Target Pose Selection}.
Another preprocessing step is to obtain the target pose $I_t$ to guide the full-body selfie generation. 
We require $I_t$ to convey two types of information: (1) the desired pose and (2) the actual body shape. To achieve this, we represent $I_t$ as the contour of the user's body, derived from a reference photo $I_r$ of the \textit{same} person.
We first discuss how to obtain $I_r$ and address the process of deriving $I_t$ from $I_r$ in the next section. 

We develop an automatic selection strategy to help obtain $I_r$ from the users' photo collection  $\Phi$.
The selection criteria are based on the similarity between the clothing types in the input selfies and a candidate image in $\Phi$.
This is because the more similar the clothing type is, the more accurately the body shape (in this particular type of outfit) can be extracted from $I_r$. Specifically, we first use a pretrained human parsing model~\cite{yang2023humanparsing} to detect the  clothing types (\eg, hoodie) in the selfies $I_u$, $I_\ell$, and $I_s$. We then apply the detection to each full-body photo in $\Phi$.
 A list of candidate reference photos is suggested  based on the number of matched clothing types (higher is preferred) between the selfies and the image from $\Phi$.
 Then the user can choose a reference image $I_r$ from the list. 
 Finally, the inpainting mask $M$ is obtained using the scaled bounding box of the person in $I_r$. 




\noindent\textbf{Pose-Guided Generation}.
After preprocessing, to generate a full-body selfie, we follow the standard diffusion denoising process with the guidance of ControlNet~\cite{zhang2023adding} to ensure that the generated image aligns with the pose in $I_r$.    
The key difference is that we modify the ControlNet architecture to enable it  to possess similar guiding capabilities for our image-conditioned diffusion model as it does for text-conditioned models. 
Specifically, we replace the text embeddings with $c$, which is computed using Eq. \ref{eq:c} by replacing  simulated selfie $I'_{k}$ with real selfie $I_{k}$, where $k \in \{f,u,\ell,s\}$. 
This modification allows any pretrained ControlNet to be plugged into our model, providing the desired guidance. 
To meet our requirements, we specifically opt for a Canny Edge ControlNet with target pose $I_t$ as the control signal. To obtain $I_t$, we segment various body parts in $I_r$ using \cite{yang2023humanparsing}, producing a semantic map. The Canny Edge $I_t$ is then detected from this semantic map (Fig. \ref{fig:overview} top right), not directly from $I_r$ itself.
This  ensures that the pipeline is not influenced by  the outfit details in $I_r$, such as cloth texture.
Note that, as we will discuss, one can always use OpenPose ControlNet, and obtain $I_t$ (skeleton) from any human image, albeit sacrificing accurate body shape.

To obtain the full-body selfie,  starting from a random noise $z_T$ at timestep $T$, we iteratively perform one-step denoising using $\epsilon_\theta (z_t, I_b \cdot M, c, t)$ with modified ControlNet for $T$ steps. This yields the clean latent $z_0$, which is further decoded by decoder $D$ to generate the full-body selfie $I_n$. 
However, as shown in Figure \ref{pipeline_ablation} (c),  directly using trained model in Sec.~\ref{sec:training} produces inaccurate outfit and identity due to the gap between synthetic and real data.



\noindent \textbf{Fine-Tuning}.
To improve results, we fine-tune the selfie-conditioned inpainting model on the specific capture. 
By leveraging the full-body generative prior in the model, this strategy is robust to distribution differences (\eg, pose variations in the upper body) between $S$ and $S'$, enhancing the preservation of clothing texture and generating pose-specific details (wrinkles on upper clothing in Fig. \ref{pipeline_ablation} (d)).


To implement this strategy, we generate a ``ground truth'' for fine-tuning by resizing and placing a randomly selected selfie image from the set $S$ into the masked area of the image $I_b$ (see supplementary for details).  The bottom left part of Fig. \ref{fig:overview} orange box visualizes  examples of two augmented images produced using $I_f$ and $I_u$.  
We create 200 augmented images and employ them to fine-tune $\epsilon_\theta (z_t, I_b \cdot M, c, t)$ supervised by the loss computed similar to Eq. \ref{eq:loss}. 
While fine-tuning on close-up selfies, the full-body generative prior in the trained model helps prevent overfitting, especially for $I_u$ with significant pose differences.
Finally, we utilize the fine-tuned model to produce full-body selfie $I_o$ using the pose-guided generation. 
Fig. \ref{pipeline_ablation} (d) shows an example of fine-tuned output, which has reasonable outfit and shading but  wrong identity. 
The identity problem arises because the VAE used in Latent Diffusion fails to generate details of the small face in the full-body photo, a well-known limitation. 
Similar challenges arise with shoes because they are too small in $I_o$.


\noindent \textbf{Appearance Refinement}.
To further enhance the details of $I_o$, we employ several post-processing strategies.
First, for refining face and shoes, we augment $I_f$ and $I_s$ by resizing them to various resolutions and zero-padding the border. Subsequently, we train a DreamBooth model~\cite{ruiz2022dreambooth} with two concepts (shoes and face) using these augmented images. 
Finally, we crop the face region from $I_o$ and then employ SDEdit~\cite{meng2021sdedit}, based on the trained DreamBooth, to refine the face. We subsequently compose the refined face back into $I_o$. A similar operation is applied to the shoes region. 
Second, we also apply the similar operation to hands region, but with a pretrained Stable Diffusion model~\cite{Rombach_2022_CVPR} since hands are usually not shown in the input selfies. 
As shown in Fig.~\ref{pipeline_ablation} (e) and Fig.~\ref{fig:overview} purple box, these post-processing steps culminate in the creation of the final output, denoted as $I_a$. This final output faithfully preserves the identity and outfit, and exhibits realistic shading, along with the desired pose.  

At the end, the user can switch to a new reference photo and use the current fine-tuned model for generation, provided the person in the new reference photo is within the inpainting mask $M$. 
If not, fine-tuning for the new reference photo is required. See the supplementary for execution time and implementation details, including strategy to help preserve background content inside the inpainting mask.

\section{Experiments}

\begin{figure*}[!h]
\centering
{\includegraphics[width=0.189\linewidth]{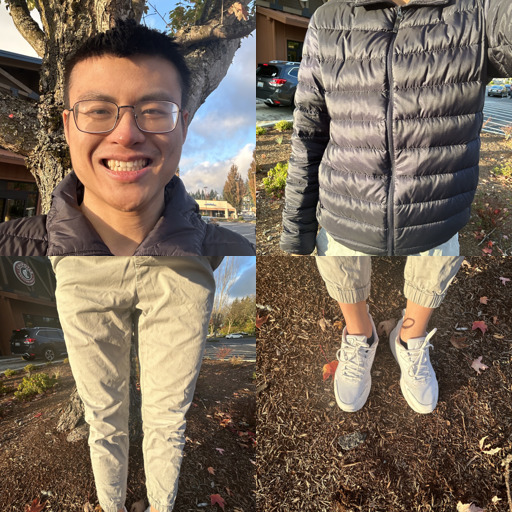}}
{\includegraphics[width=0.189\linewidth]{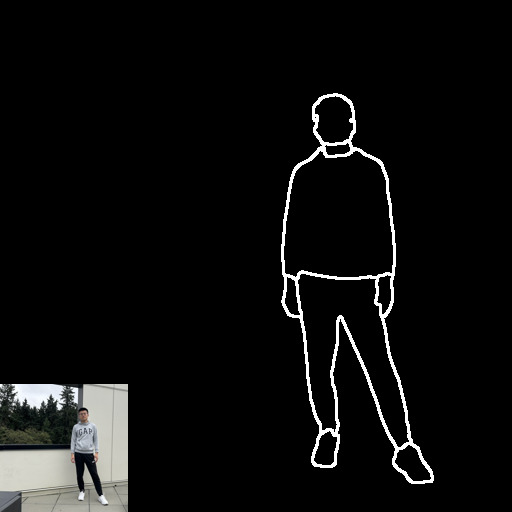}}
{\includegraphics[width=0.189\linewidth]{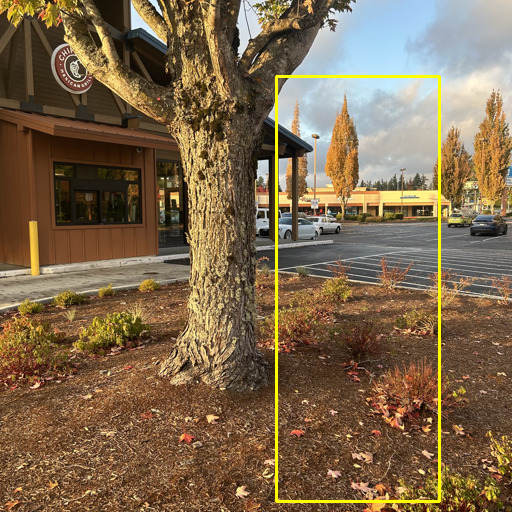}}
{\includegraphics[width=0.189\linewidth]{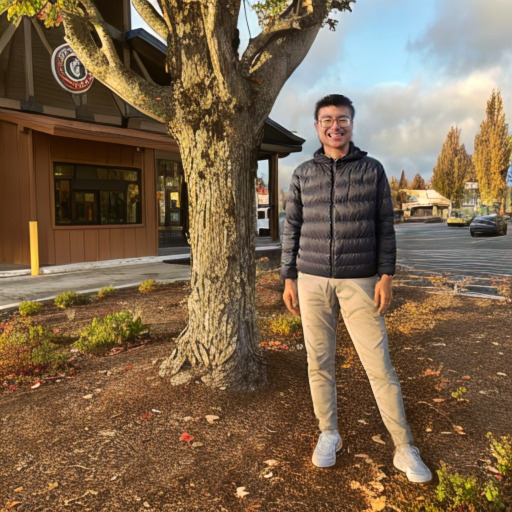}}
\\
{\includegraphics[width=0.189\linewidth]{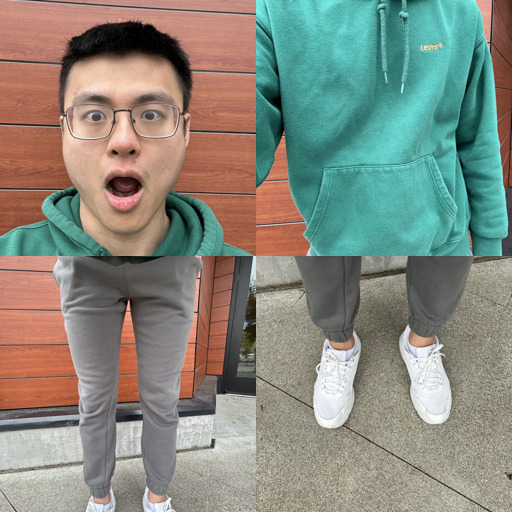}}
{\includegraphics[width=0.189\linewidth]{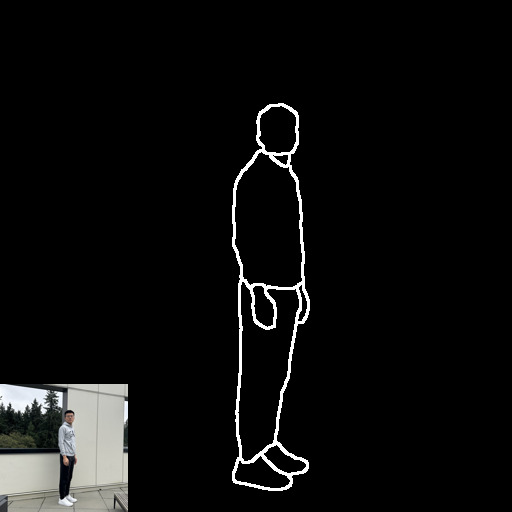}}
{\includegraphics[width=0.189\linewidth]{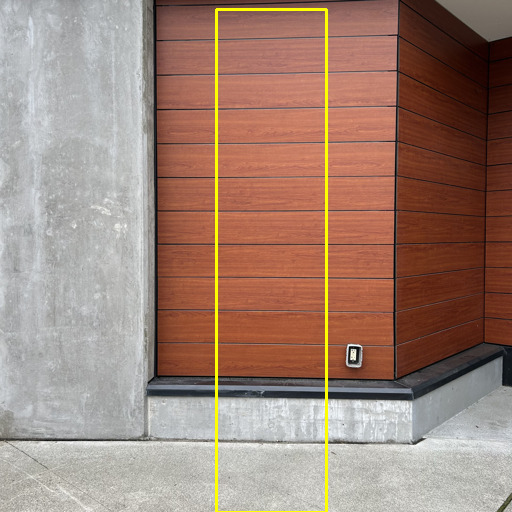}}
{\includegraphics[width=0.189\linewidth]{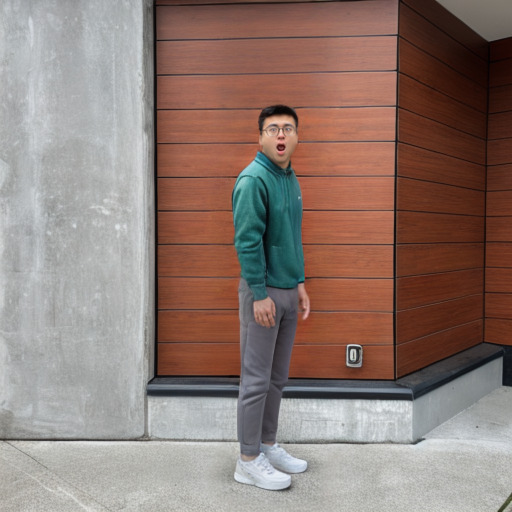}}
\\
{\includegraphics[width=0.189\linewidth]{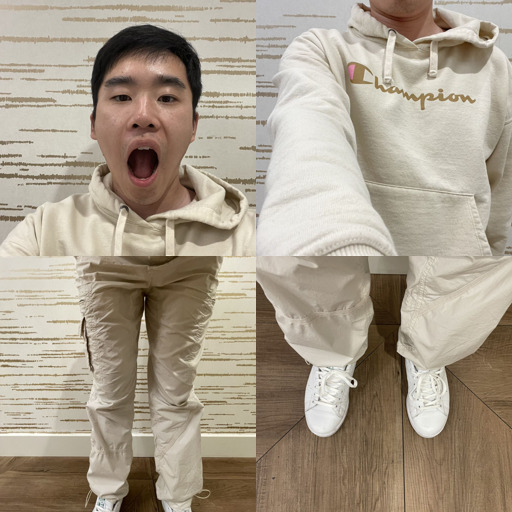}}
{\includegraphics[width=0.189\linewidth]{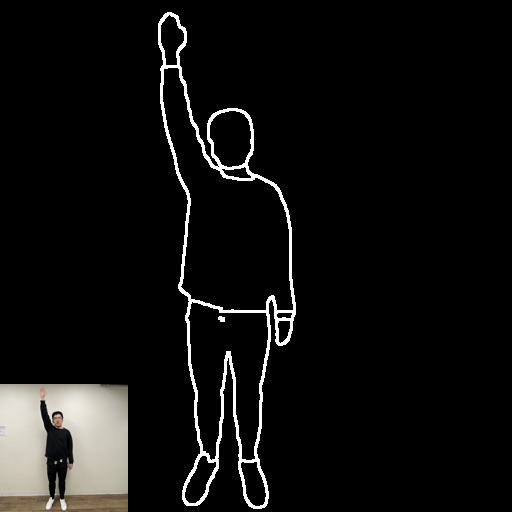}}
{\includegraphics[width=0.189\linewidth]{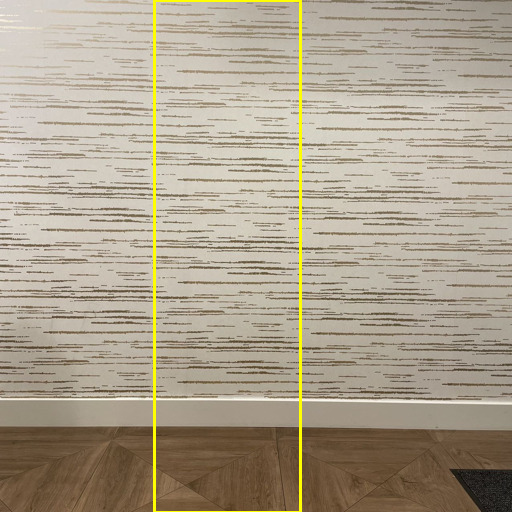}}
{\includegraphics[width=0.189\linewidth]{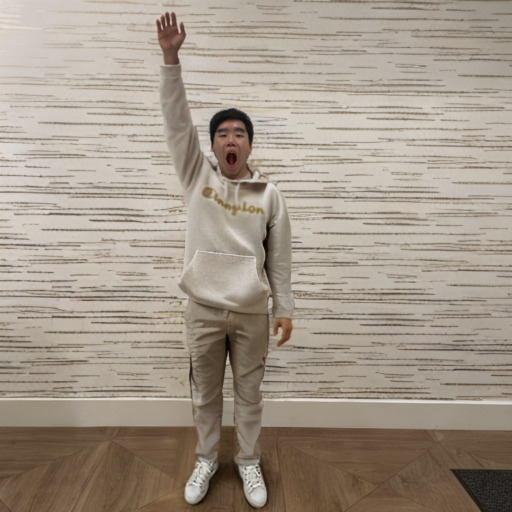}}
\\
{\includegraphics[width=0.189\linewidth]{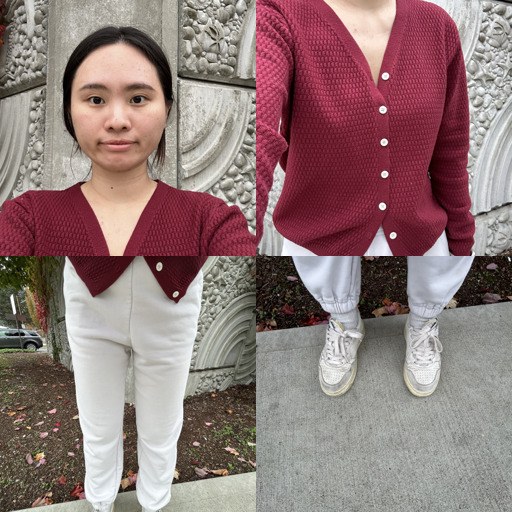}}
{\includegraphics[width=0.189\linewidth]{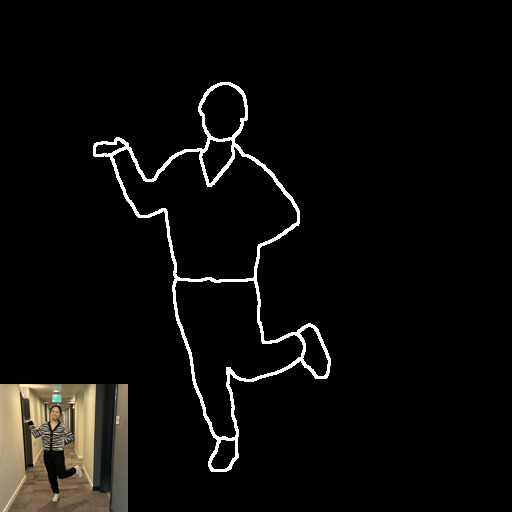}}
{\includegraphics[width=0.189\linewidth]{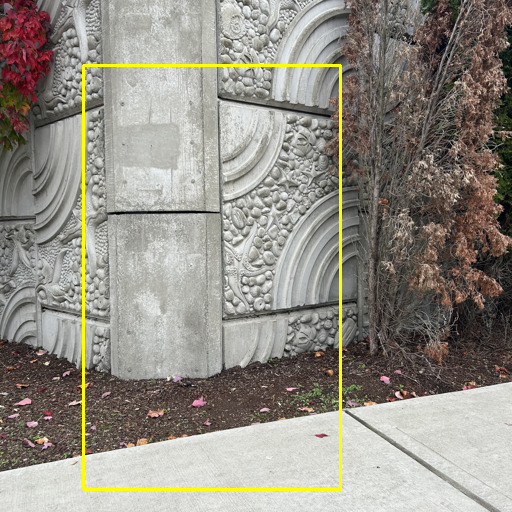}}
{\includegraphics[width=0.189\linewidth]{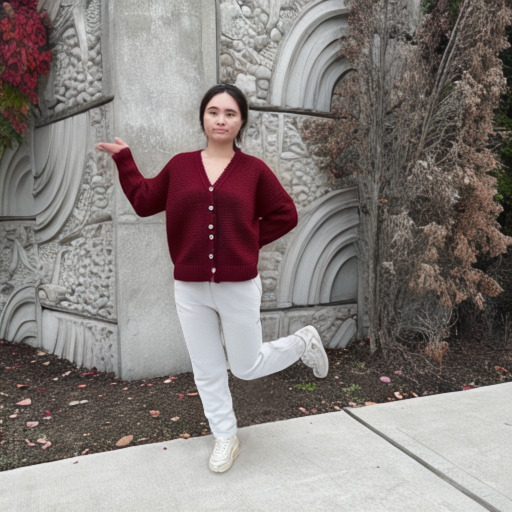}}
\\
{\includegraphics[width=0.189\linewidth]{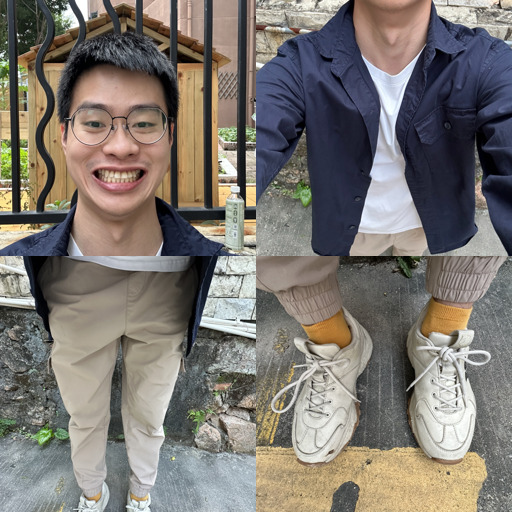}}
{\includegraphics[width=0.189\linewidth]{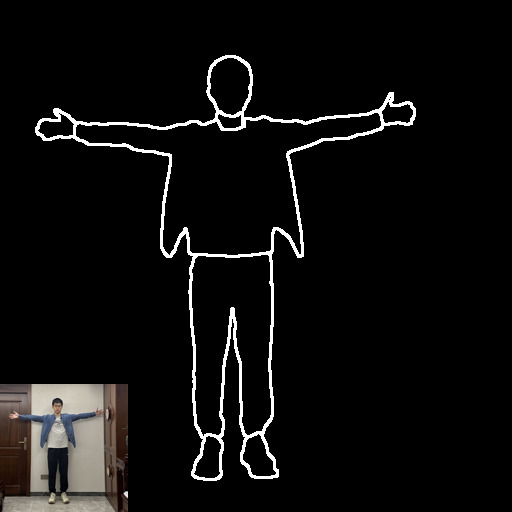}}
{\includegraphics[width=0.189\linewidth]{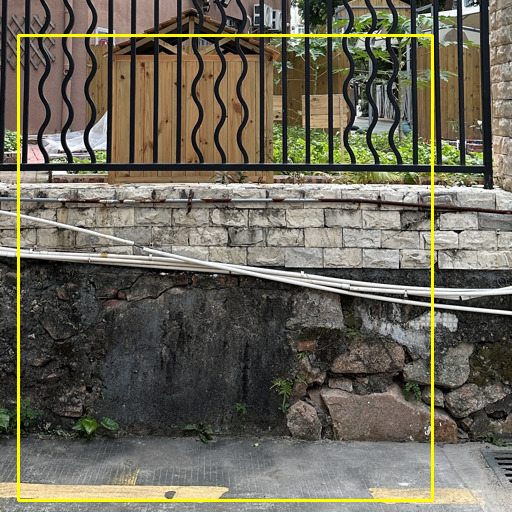}}
{\includegraphics[width=0.189\linewidth]{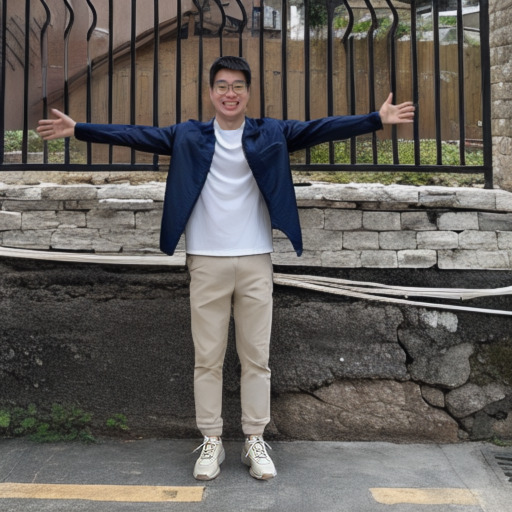}}
\\
  \subcaptionbox{ Input Selfies}%
{\includegraphics[width=0.189\linewidth]{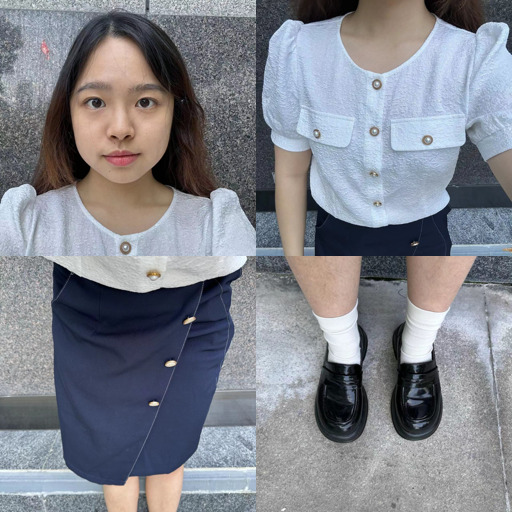}}
    \subcaptionbox{Target Pose }%
{\includegraphics[width=0.189\linewidth]{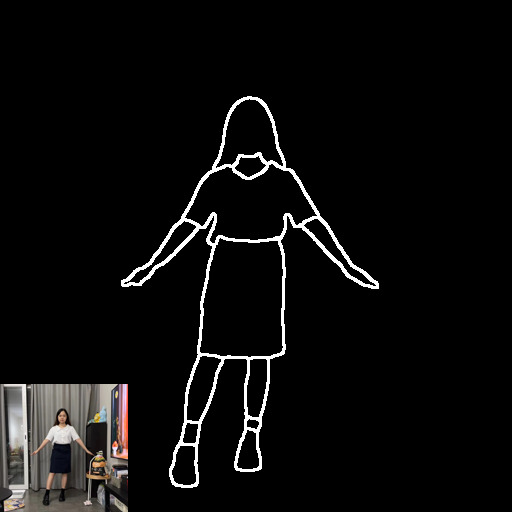}}
  \subcaptionbox{ Masked Background}%
{\includegraphics[width=0.189\linewidth]{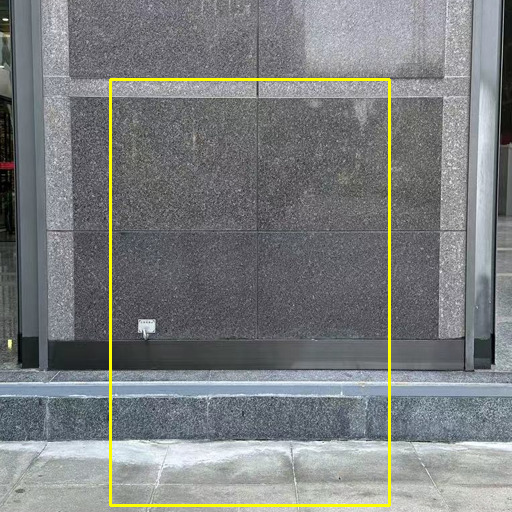}}
    \subcaptionbox{Total Selfie}%
{\includegraphics[width=0.189\linewidth]{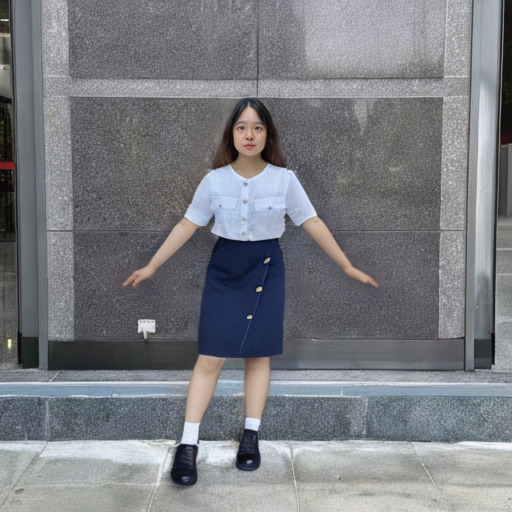}}
\vspace{-2mm}
\caption{Results. The second column shows the Canny Edge images detected from reference images (shown as insets). Regions inside yellow box of (c) are the masked regions.
Total Selfie generates realistic, full-body images of different individuals with diverse poses and expressions against a variety of backgrounds, while preserving facial expression and clothing. 
The results are robust to selfies captured in different ways, such as those with one or two hands involved or from a downward-looking perspective (row 5), and with target pose images in outfits that differ somewhat from the input selfies.
}
  \label{fig:our_results}
\end{figure*}

\noindent
\textbf{Results.}
We begin by showcasing the results of Total Selfie across five different captures, as illustrated in Figure \ref{fig:our_results}. 
The goal is to retain the user's facial expression and clothing, but realistically retargeted onto the background with the desired pose; these results demonstrate this capability.

In particular, 
the results show compelling full-body selfies with a wide range of poses, even if the desired poses significantly deviate from those in the input selfies (row 2 to 5). 
Importantly, the method is able to add realistic wrinkles in the clothing to fit the new pose, with consistent shading (\eg, raised arm in rows 3 to 5).  
Observe that the technique works with a range of clothing, from short sleeves to jackets, and both pants and skirts.
Finally, the method is able to convincingly fill in missing details (\eg, hands, which are often missing in selfies), that fit the individual and are realistically shaded in the target scene.
%
More results are in the supplementary.

\noindent \textbf{Evaluation Data.} 
For evaluation, we collected a dataset of twelve people wearing various outfits with selfies taken in a variety of scenes and lighting conditions, resulting in a total of seventeen  captures. 
Additionally, we captured real, ``ground truth'' full-body photos of each subject, maintaining the same clothing, nearly identical facial expressions and background.

\noindent
\textbf{Ablation Study.}
We study the effects of different parts of the pipeline and design three variants: 
(1) \textit{Ours-FT-AR}:  full pipeline without 
 per-capture fine-tuning and appearance refinement. 
(2) \textit{Ours-AR}:  full pipeline without appearance refinement,
(3) \textit{Ours-FU}:  full pipeline without face undistortion. 
As discussed in Sec. \ref{sec:method}, Fig. \ref{pipeline_ablation} and Table. \ref{tab:main_quantitative} show that the full pipeline outperforms all tested variants.

\noindent
\textbf{Comparison to Baseline.}
To the best of our knowledge, there are no existing papers solving the same task. 
We therefore adapt four existing methods to create four baselines.
(1) 
\textit{Paint-By-Example}~\cite{yang2022paint} was designed to inpaint a masked image using a single source image. Here, we concatenate four input selfies vertically as the source since this works better than only using one selfie. We use Canny Edge ControlNet~\cite{zhang2023adding} to guide the pose.
(2)
\textit{DisCo}~\cite{wang2023disco} reposes an input human image using diffusion models, given a background photo and a target pose. 
We opt for this over other reposing approaches because it can handle complex backgrounds.
Similar as before, we concatenate four input selfies vertically to create the input human image for better performance.
Following the official implementation, we use OpenPose Skeleton as target pose since their provided model does not work well with Canny Edge input. 
(3)
\textit{LaDI-VTON}~\cite{morelli2023ladi} is a diffusion-based Virtual Try-On method. We opt for this method over other similar methods since it can handle both upper- and lower-body garments, and the code is available. 
This method specifically requires a full-body RGB image as input. 
Thus we use Stable Diffusion and Canny Edge ControlNet to generate the full-body input aligning with target pose. Then we apply garment editing on this  input image using upper and lower body selfies, excluding the face and shoes, as this method does not support editing those elements. Finally, we blend the edited image with the input background for the final output.
(4)
\textit{DreamBooth}~\cite{ruiz2022dreambooth} customizes a pretrained text-to-image diffusion model by fine-tuning with a few reference images. Here, we use four selfies (with augmentation) as reference images and generate outputs using Canny Edge ControlNet.

\begin{table}[!t]
\centering
\resizebox{0.45\textwidth}{!}{
\begin{tabular}{|l|llll|}
\toprule
       {Method}  &  {LPIPS} $\downarrow$       & {SSIM} $\uparrow$  & {PSNR} $\uparrow$   & {FID} $\downarrow$  \\
\midrule
 Paint-By-Example       & 0.252     & 0.621    &  12.37   &  184.6  \\
DisCo      &   0.287   &   0.586   &   12.13   &  245.7  \\
LaDI-VTON       &  0.263   &   0.563   &  10.35  &  172.8  \\
DreamBooth      &  0.224  &    0.663  &   13.61  &   159.2 \\
\midrule
Ours-FT-AR      & 0.218  & 0.691 &  14.19  & 154.1 \\
Ours-AR        &  0.190 &    0.703 & 16.89   & 143.4 \\
Ours-FU     & 0.188  &   0.706  &  16.98  & 141.6   \\
\midrule
Ours          &  \textbf{0.187} &  \textbf{0.708} &   \textbf{17.01} & \textbf{139.4}   \\
\bottomrule
\end{tabular}
}
\vspace{-2mm}
\caption{Comparison with baselines and our ablation variants. The metrics are evaluated only in the foreground. 
We employ ground truth Canny Edge to define the target pose for all rows. 
}
\vspace{-6mm}
\label{tab:main_quantitative}
\end{table}

\begin{figure*}[!t]
\centering
{\includegraphics[width=0.162\linewidth]{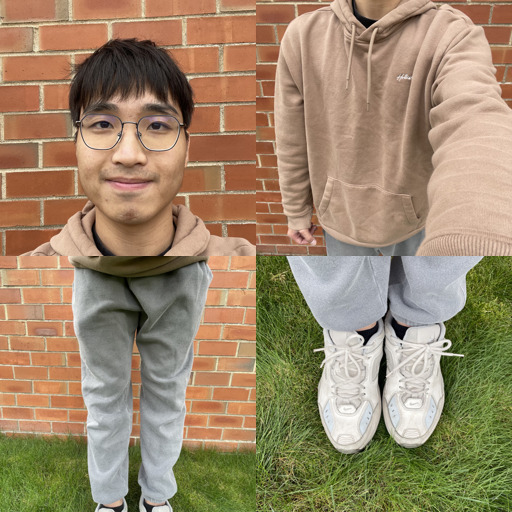}}
{\includegraphics[width=0.162\linewidth]{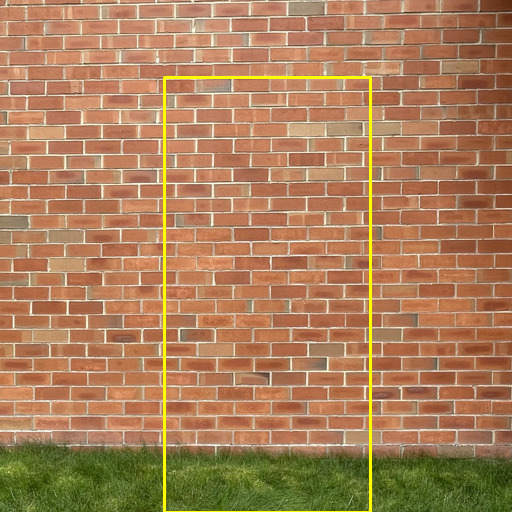}}
{\includegraphics[width=0.162\linewidth]{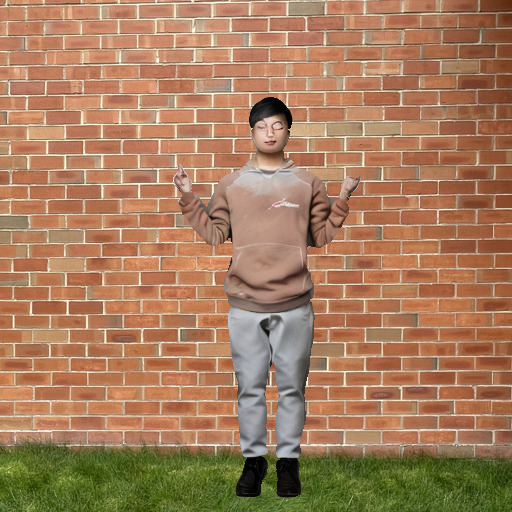}}
{\includegraphics[width=0.162\linewidth]{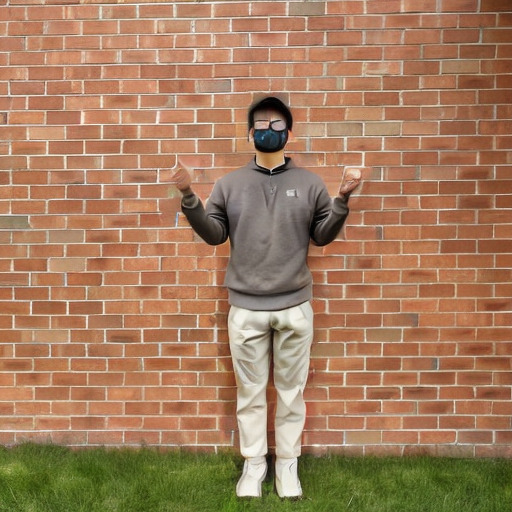}}
{\includegraphics[width=0.162\linewidth]{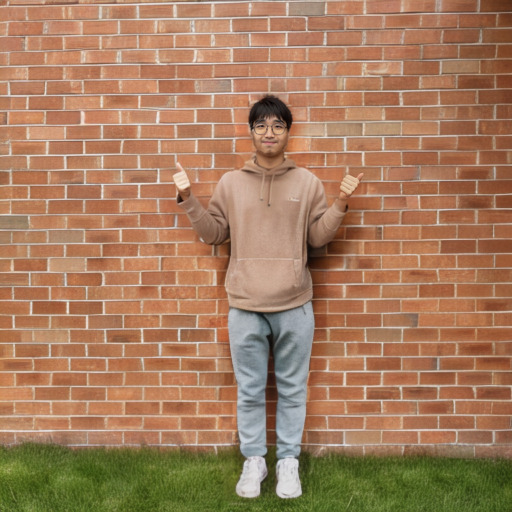}}
{\includegraphics[width=0.162\linewidth]{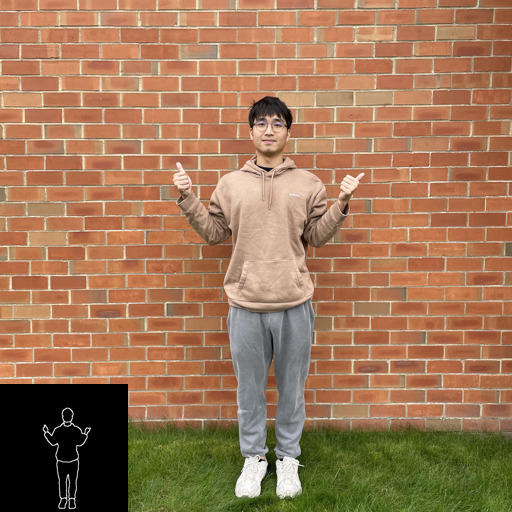}}
\\
  \subcaptionbox{ Input Selfies}%
{\includegraphics[width=0.162\linewidth]{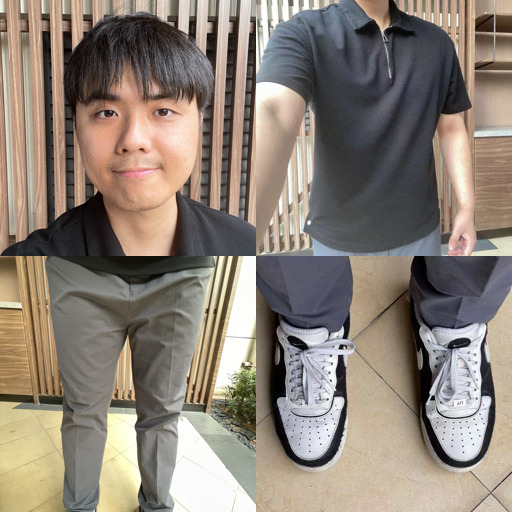}}
  \subcaptionbox{ Background}%
{\includegraphics[width=0.162\linewidth]{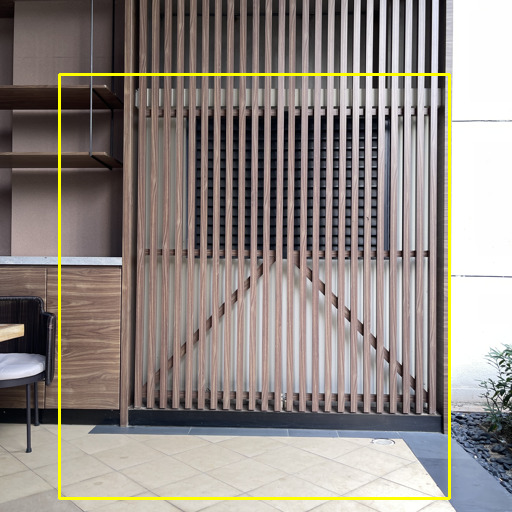}}
    \subcaptionbox{LaDI-VTON}%
{\includegraphics[width=0.162\linewidth]{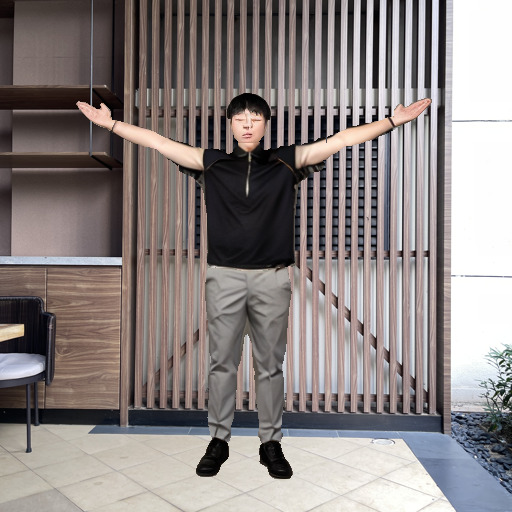}}
    \subcaptionbox{DreamBooth }%
{\includegraphics[width=0.162\linewidth]{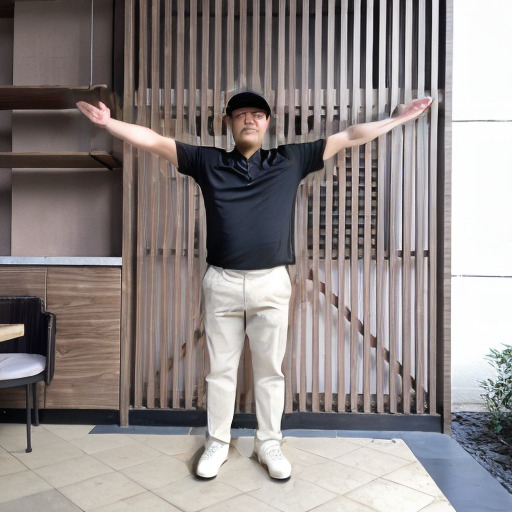}}
    \subcaptionbox{Ours}%
{\includegraphics[width=0.162\linewidth]{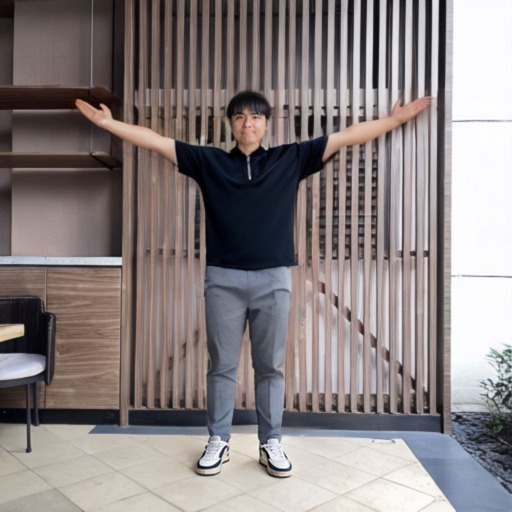}}
    \subcaptionbox{Real Photo}%
{\includegraphics[width=0.162\linewidth]{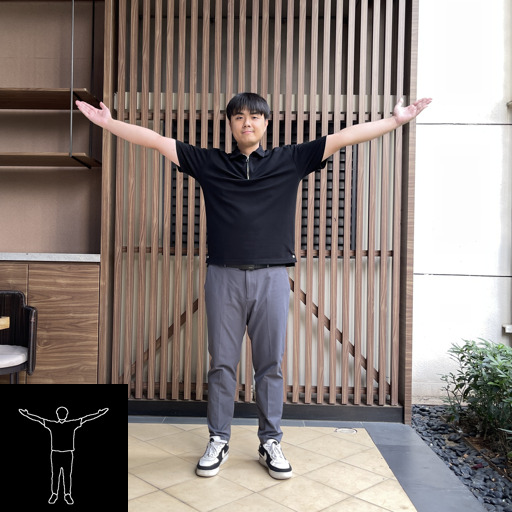}}
\vspace{-2mm}
\caption{Qualitative comparison with two best-performing baselines.  For this comparison, we used Canny Edge of the real photo as target pose (inset of (f)).  Our pipeline clearly outperforms baselines in terms of photorealism and faithfulness (zoom in for details, including faces and shoes). Note that, while the selfies, background image, and real photo were captured in the same session, variations in lighting conditions, auto exposure, white balance, and other factors may result in intensity and color tone differences. 
}
  \label{fig:comparison}
  \vspace{-2mm}
\end{figure*}

\begin{figure*}[!t]
\centering
{\includegraphics[width=0.161\linewidth]{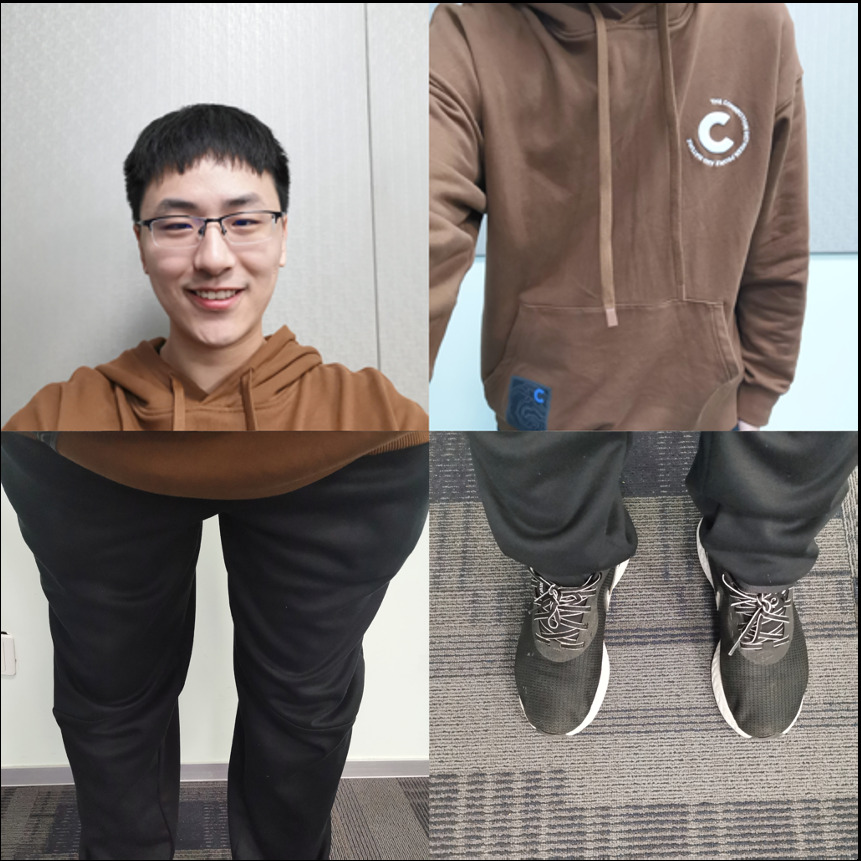}}
{\includegraphics[width=0.161\linewidth]{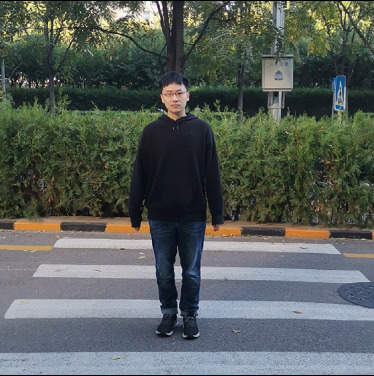}}
{\includegraphics[width=0.161\linewidth]{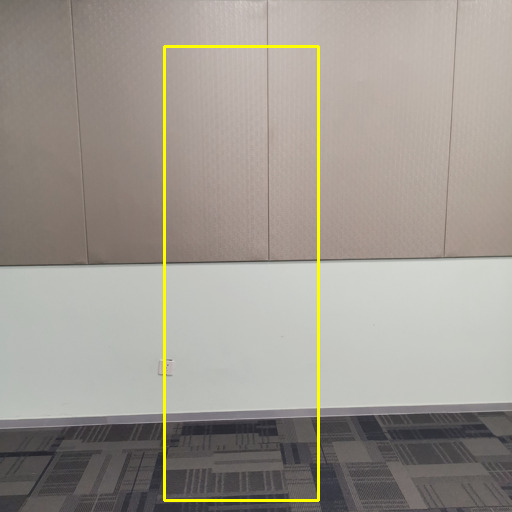}}
{\includegraphics[width=0.161\linewidth]{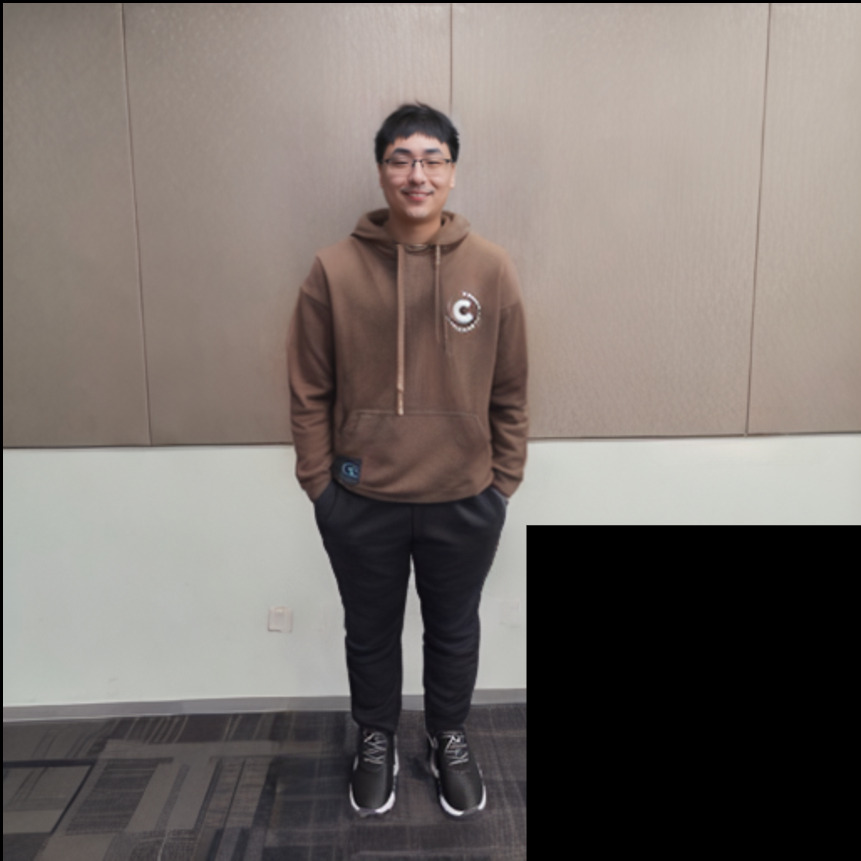}}
{\includegraphics[width=0.161\linewidth]{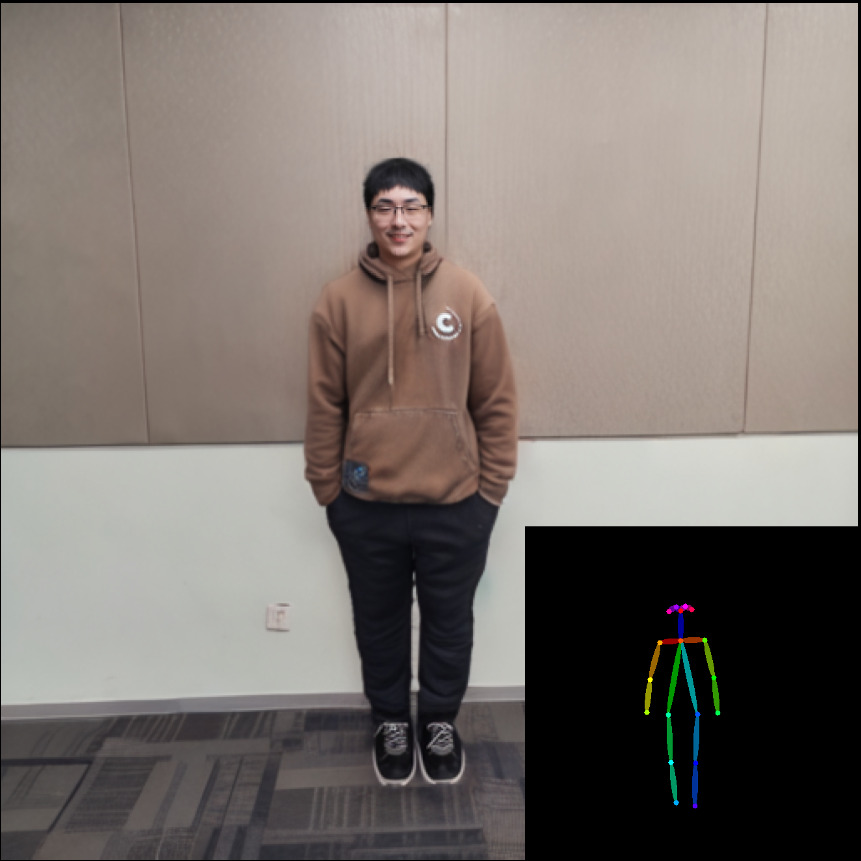}}
{\includegraphics[width=0.161\linewidth]{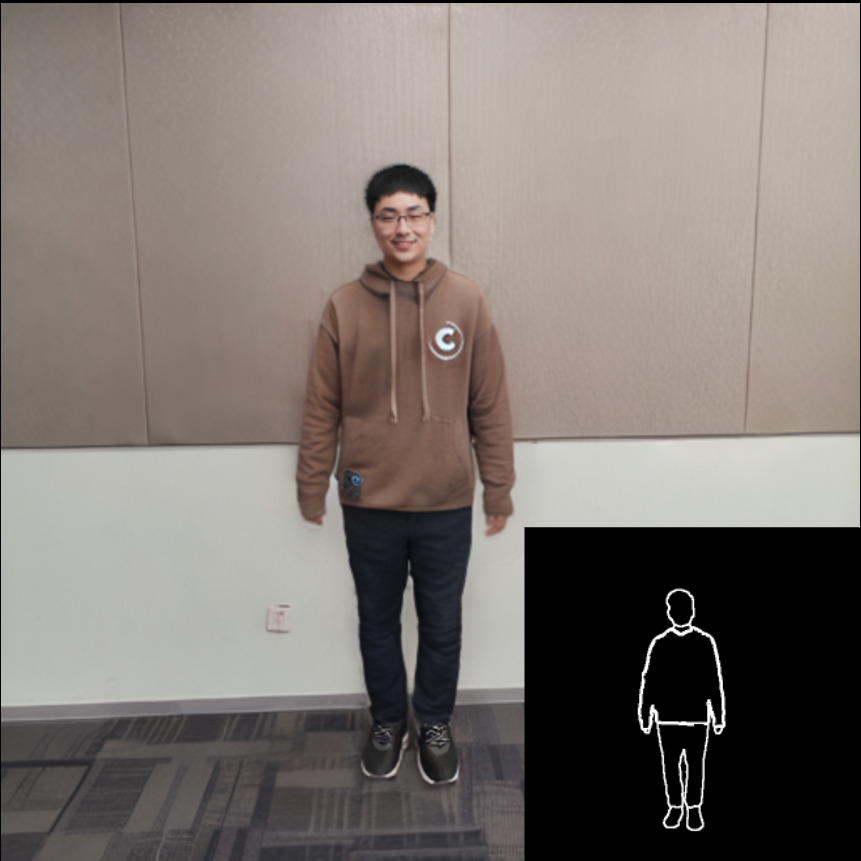}}
\\
  \subcaptionbox{ Input Selfies}%
{\includegraphics[width=0.161\linewidth]{fig/experiments/main_results/bohan_ex1_pose2/selfies_combined.jpg}}
  \subcaptionbox{ Reference Photo}%
{\includegraphics[width=0.161\linewidth]{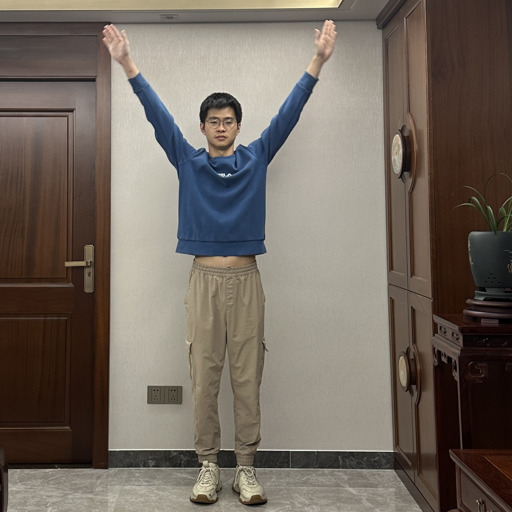}}
  \subcaptionbox{Background}%
{\includegraphics[width=0.161\linewidth]{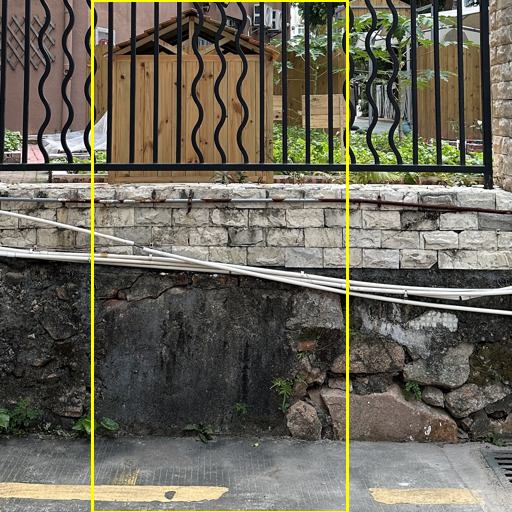}}
    \subcaptionbox{No Condition }%
{\includegraphics[width=0.161\linewidth]{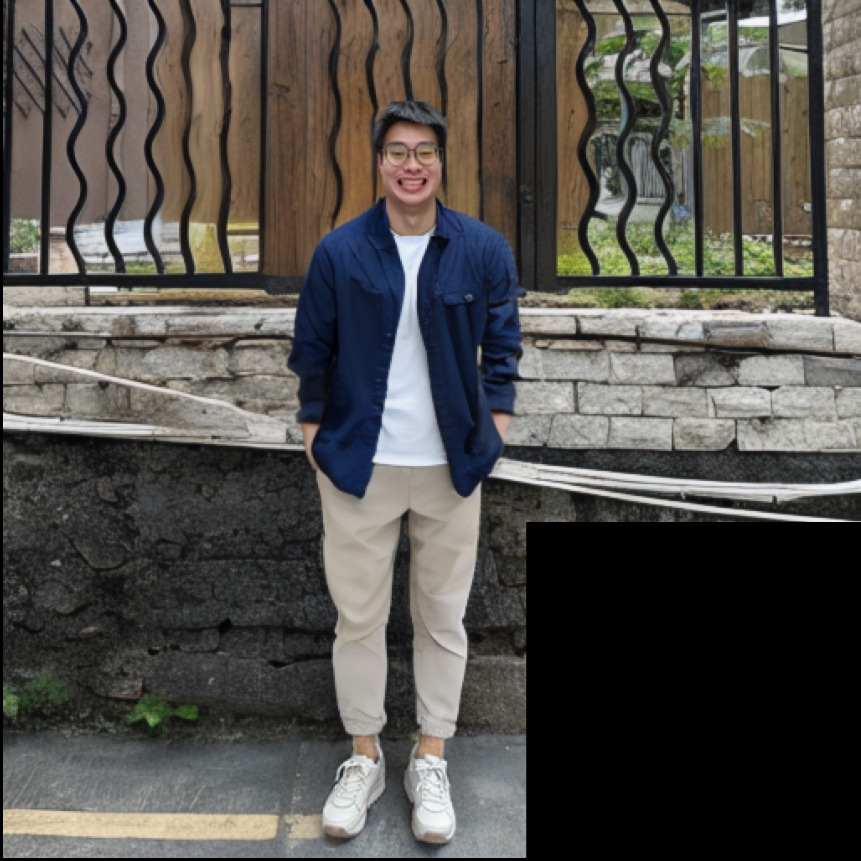}}
    \subcaptionbox{OpenPose Skeleton}%
{\includegraphics[width=0.161\linewidth]{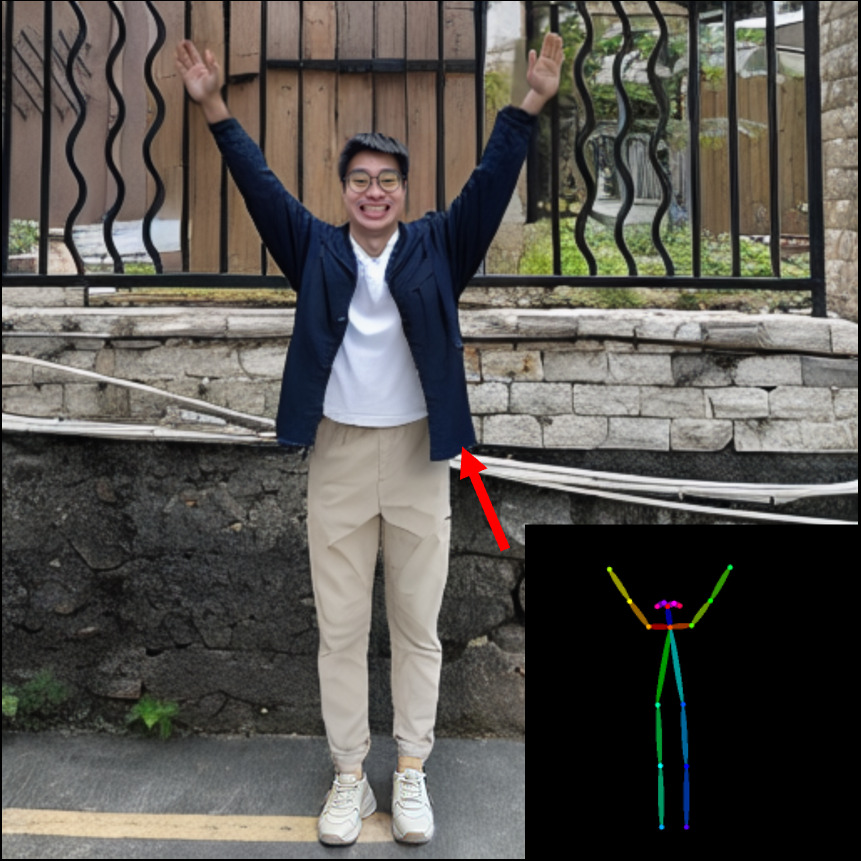}}
    \subcaptionbox{Canny Edge}%
{\includegraphics[width=0.161\linewidth]{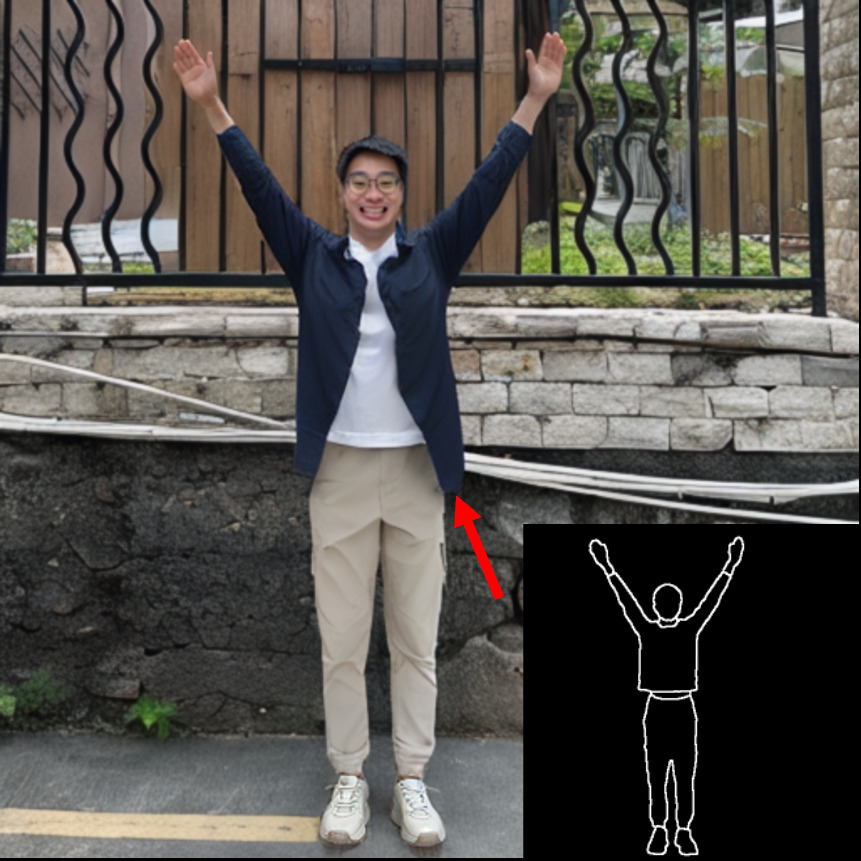}}
\vspace{-2mm}
\caption{Discussion of target pose options.  Insets of (d)-(f) show the conditional signals.  
In the first row, Canny Edge (f) exhibits the closest body shape to the reference photo (b) compared to (d) and (e).
In the second row, when the reference photo has a different clothing type, Canny Edge (f) produces an unnatural hem of the jacket (highlighted by a red arrow). In such cases, the OpenPose skeleton (e) may offer a more natural result despite a slight fattening of the waist compared to reference.
}
\vspace{-3mm}
  \label{pose_discussion}
\end{figure*}

Fig. \ref{fig:comparison} and Table. \ref{tab:main_quantitative} show the qualitative and quantitative results of the compared methods, respectively.
 Please note that real photos may have contrast and color tone variations (e.g., due to auto-exposure and white balance), leading to differences between real and generated outputs. 
Among the compared methods, \textit{DisCo} performs worst because it is guided by skeleton and assumes a third-person view input.  \textit{Paint-By-Example} is also ineffective since it is not designed specifically for full-body generation. 
Further, both methods can only consider one reference image, leading to sub-optimal results. 
\textit{LaDI-VTON} exhibits artifacts on the clothes (Fig. \ref{fig:comparison} (c)), which may be attributed to the reference garment images being selfies, distinct from those in its training set.
\textit{DreamBooth} produces inaccurate outfits (Fig. \ref{fig:comparison} (d)). This is due to the pretrained text-to-image model lacking strong priors for understanding clothes in selfies and linking them together as a coherent full-body image.
Conversely, Total Selfie, which is initially trained on a synthetic dataset and then fine-tuned per capture, excels in realistically generating full-body selfies.


\noindent
\textbf{Discussion of Target Pose Options.}
We explore the trade-offs of different target pose options, as shown in Fig. \ref{pose_discussion}.  
\textit{No Condition} is the simplest option, requiring no target pose input. Our pipeline automatically determines pose, body shape, and scale from the masked background and input selfies. However, it generates inaccurate body shape and scale (row 1(d)), and lacks pose controllability.
\textit{OpenPose Skeleton} enables users to specify the target pose (skeleton) using any human photo. Our pipeline integrates this control signal using a modified OpenPose ControlNet. While this option produces reasonable poses, it struggles with correct body shapes (row 1(e)).
\textit{Canny Edge} offers accurate body shape and pose but may yield sub-optimal results when the reference pose photo has different clothing than the input selfies. For instance, the hem of the jacket in row 2(f) appears less natural.  In such cases, \textit{OpenPose Skeleton} (row 2(e)) can do a better job with clothing contours.
In summary, each option has its advantages and limitations, and there is no one-size-fits-all choice.
The selection of the target pose type depends on users'  needs and available data.

\begin{figure}[!b]
\vspace{-5mm}
\centering
  \subcaptionbox{ Input Selfies}%
{\includegraphics[width=0.3\linewidth]{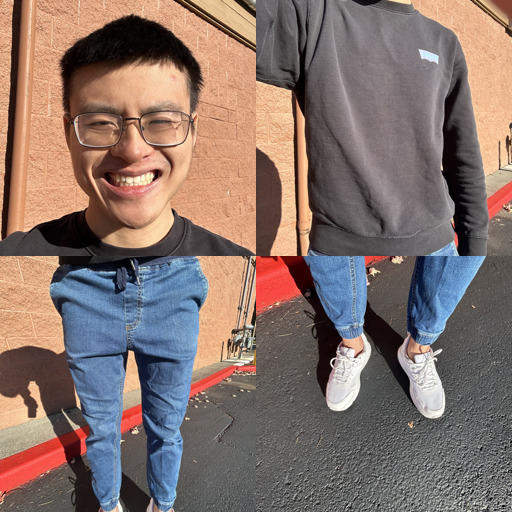}}
  \subcaptionbox{ Ours}%
{\includegraphics[width=0.3\linewidth]{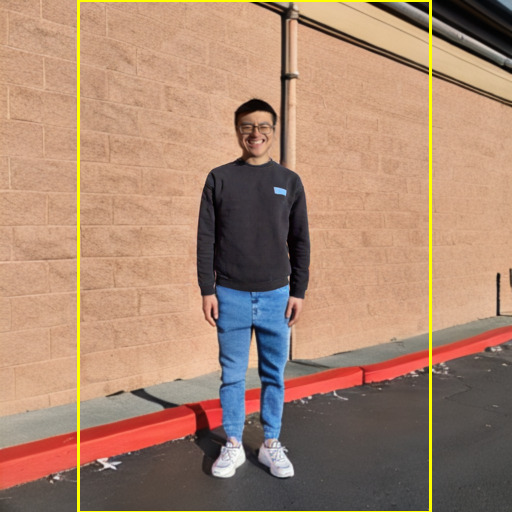}}
  \subcaptionbox{ Real Photo}%
{\includegraphics[width=0.3\linewidth]{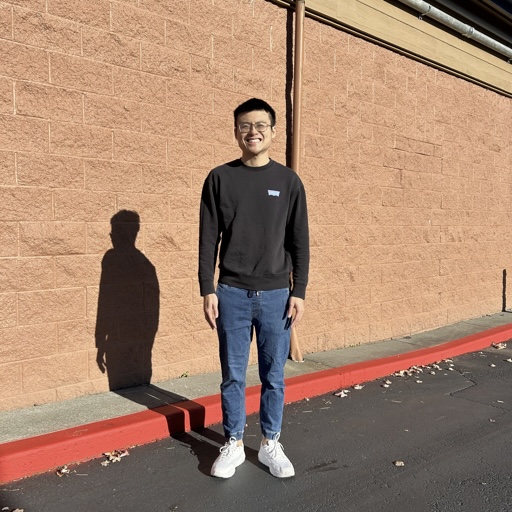}}
\vspace{-2mm}
\caption{Failure case of shadow generation under strong sunlight. Masked regions are shown in (b), and target pose is from (c).
}
\vspace{-5mm}
  \label{fig:limitations}
\end{figure}

\noindent
\textbf{Limitations and Future Work.}
Fig. \ref{fig:limitations} shows two limitations of Total Selfie:
(1) 
While our method generally yields a harmonized output (b), the shading of the body may not precisely align with that in the actual photo (c).
(2)
Our method cannot accurately generate  hard shadows of a person under strong sunlight since inferring the sun's direction and scene geometry solely from the background is difficult.

Topics of future work include explore how to effectively infer body shape and scale from input selfies, and automatically suggesting good target poses.

\noindent
\textbf{Conclusions.}
We introduce a new selfie type called \textit{total selfie}, and propose a diffusion-based framework to generate it from four selfies, a background image, and a target pose. 
Our method generates faithful and realistic full-body selfies, outperforming  existing techniques.

\noindent
\textbf{Acknowledgement.} This work was supported by the UW Reality Lab, Meta, Google, OPPO, and Amazon.

{
    \small
    \bibliographystyle{ieeenat_fullname}
    \bibliography{main}
}


\end{document}


\twocolumn[{%
\renewcommand\twocolumn[1][]{#1}%
\maketitlesupplementary
\begin{center}
    \centering
        \captionsetup{type=figure}
\begin{subfigure}[b]{0.19\linewidth}
\includegraphics[width=\linewidth]{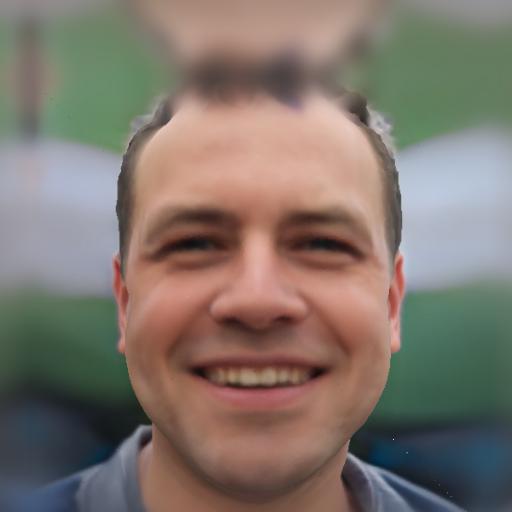} %
\caption{$d=1$}
\end{subfigure}
\begin{subfigure}[b]{0.19\linewidth}
\includegraphics[width=\linewidth]{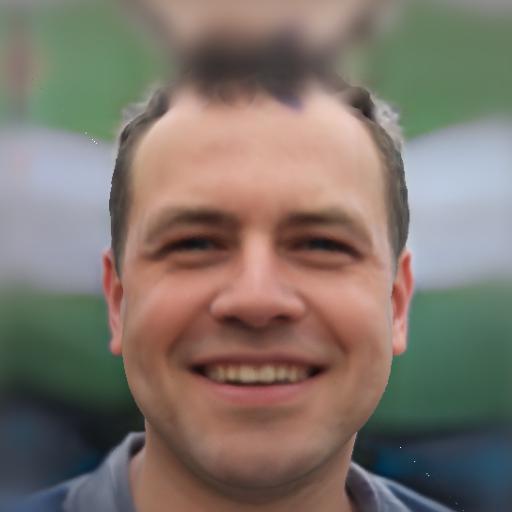} %
\caption{$d=1.3$}
\end{subfigure}
\begin{subfigure}[b]{0.19\linewidth}
\includegraphics[width=\linewidth]{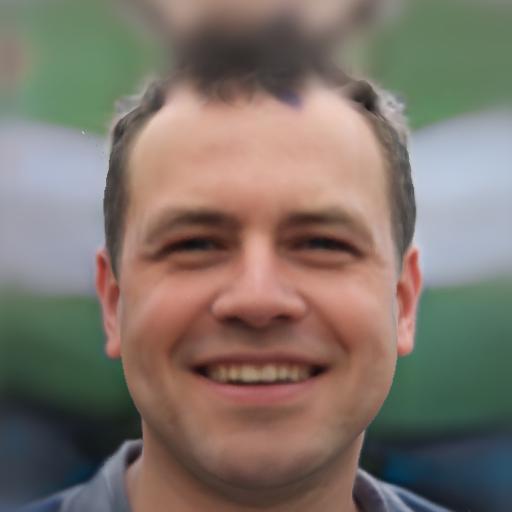} %
\caption{$d=1.6$}
\end{subfigure}
\begin{subfigure}[b]{0.19\linewidth}
\includegraphics[width=\linewidth]{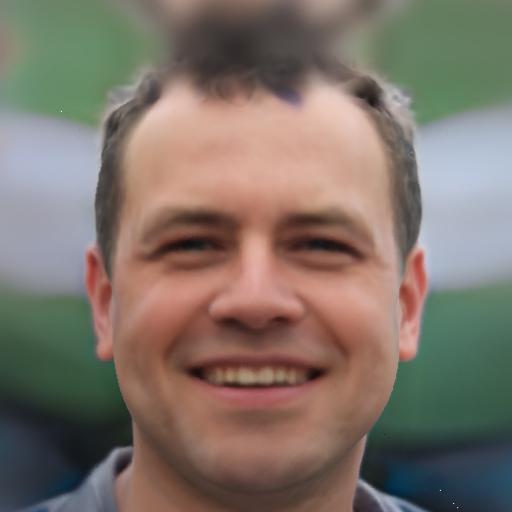} %
\caption{$d=1.9$}
\end{subfigure}
\begin{subfigure}[b]{0.19\linewidth}
\includegraphics[width=\linewidth]{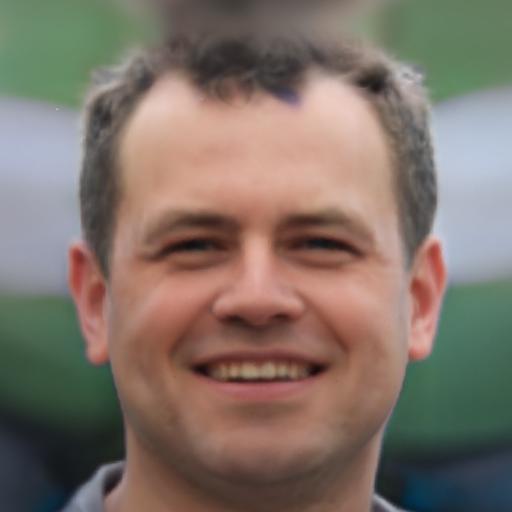} %
\caption{$d=10$ (Ground-Truth)}
\end{subfigure}
\caption{Example of 4 pairs of training data rendered from one textured mesh. The left 4 columns are the input images, and the last column is their ground truth. }
\label{fig:dataset}
\end{center}%
}]

\section{Face Undistortion}

A common problem with face selfies is perspective distortion. This is caused by the camera being too close to the subject, resulting in facial features closer to the camera appearing larger and those farther appearing smaller, thereby creating an unnatural and distorted appearance.

Previous studies have addressed this issue either through single-image optimization~\cite{wang2023disco,shih2019distortion} or training on a combined dataset of real and unrealistic synthetic images~\cite{zhao2019learning}.
For test-time efficiency, we follow the idea of large dataset training. This includes two steps: 
(1) generate high-quality paired dataset with distorted and undistorted face images. 
(2) Train a network on this dataset. 

The goal of the first step is to render a pair of images with small and large camera-subject distance. 
To implement this, we adopt EG3D~\cite{Chan2021}, a state-of-the-art textured 3D head generation method. 
EG3D utilizes a random noise vector and camera parameters to generate tri-planes, which can then be employed for volumetric rendering to produce color images and meshes.
One straightforward idea for pair generation is to fix the random noise vector and adjust the camera parameters  to directly render desired RGB images.  
However, this is not feasible as EG3D is pre-trained on a dataset with a specific camera-subject distance. Consequently, rendering images with out-of-distribution camera-subject distances results in noticeable artifacts.

Instead, we create a textured 3D head mesh using EG3D and render head images at varying distances using rasterization and the Phong shading model.
Specifically,  we employ EG3D to create tri-planes, which are utilized to sample the volume, producing a cube with dimensions $H \times W \times C$ containing density and color values.   The surface of the head (including background) is then extracted as a mesh using the Marching Cubes algorithm~\cite{lorensen1987marching}. For each 3D surface vertex, the vertex color is determined by assigning the color value of the nearest point on the cube.
With the textured mesh in place, we proceed to render images at varying distances using conventional rendering techniques. The camera rotation matrix is fixed, and only the camera distance $d$ is adjusted. To maintain consistent eye positions across different images of the same mesh, the focal length $f$ is computed based on the camera distance, given by:
\begin{equation}
    f = d f_0,
\end{equation}
where $f_0=2.9$ represents the pre-defined focal length for rendering images without invalid pixels (\ie, ensuring that all camera rays can hit the mesh) when $d=1$.    
We use PyTorch3D to render 4 input images with severe distortion by setting $d$ to 1, 1.3, 1.6, and 1.9. Additionally, a shared ground-truth image is rendered with $d$ set to 10. For better alignment, all rendered images are processed using the FFHQ face alignment technique proposed by \cite{karras2019style}.
Fig. \ref{fig:dataset} shows 4 training pairs derived from a single textured mesh. In total, we generate 10,000 textured meshes, each yielding four training pairs, resulting in a dataset comprising 40,000 training pairs.

The next step is to train an undistortion network using the rendered dataset. For this, we adapt an existing method called facevid2vid~\cite{wang2021facevid2vid}. This method uses a source image and a driving image to synthesize a talking-head image with appearance and head pose derived from the source and driving images respectively.
For our task, we made two modifications: (1) Both the source and driving images are the image with severe distortion, and the output image is the undistorted image, which will be supervised by our rendered ground-truth. 
 (2) Instead of using shared estimators, we use different estimators (same architecture, different weights) to compute driving keypoints. This enables the network to predict the driving keypoints used for undistortion.
 Finally,  the facevid2vid consists of a couple of face feature extractors that can be applied to any face image regardless of  the downstream task.  In order to harness this power,  we choose to fine-tune the pretrained model on our dataset instead of training from scratch. 

\begin{figure}[!t]
\centering
\captionsetup[subfigure]{labelformat=empty}
\begin{subfigure}[b]{0.32\linewidth}
\includegraphics[width=\linewidth]{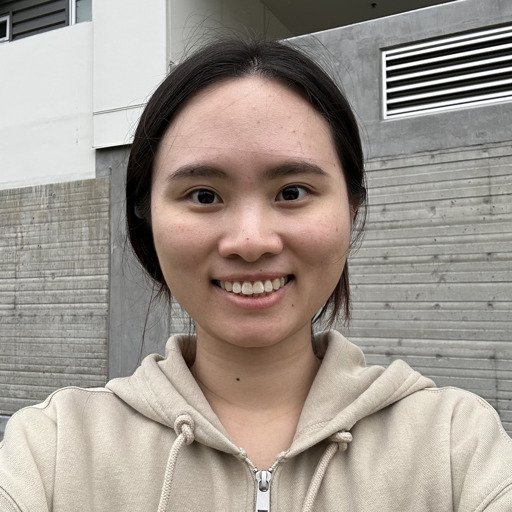} %
\end{subfigure}
\begin{subfigure}[b]{0.32\linewidth}
\includegraphics[width=\linewidth]{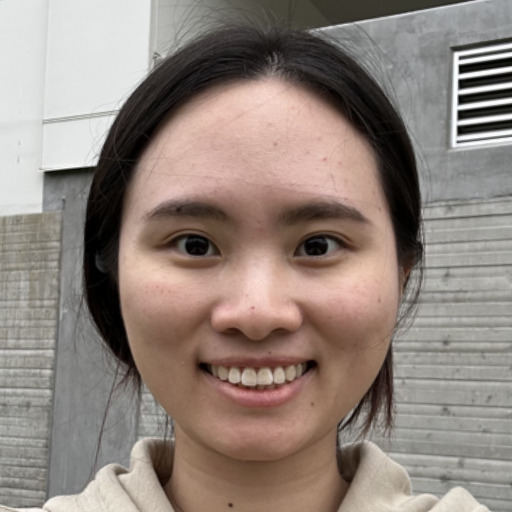} %
\end{subfigure}
\begin{subfigure}[b]{0.32\linewidth}
\includegraphics[width=\linewidth]{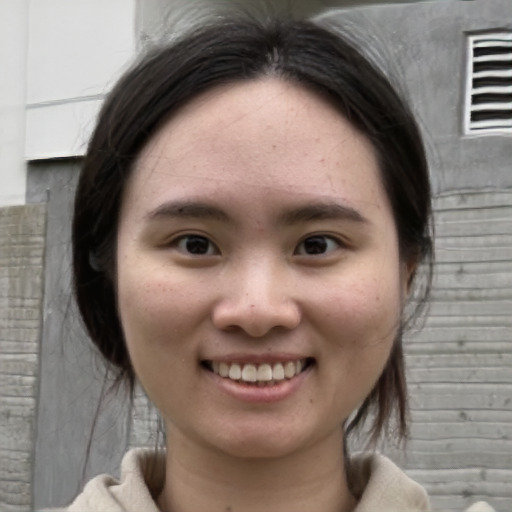} %
\end{subfigure}
\\
\begin{subfigure}[b]{0.32\linewidth}
\includegraphics[width=\linewidth]{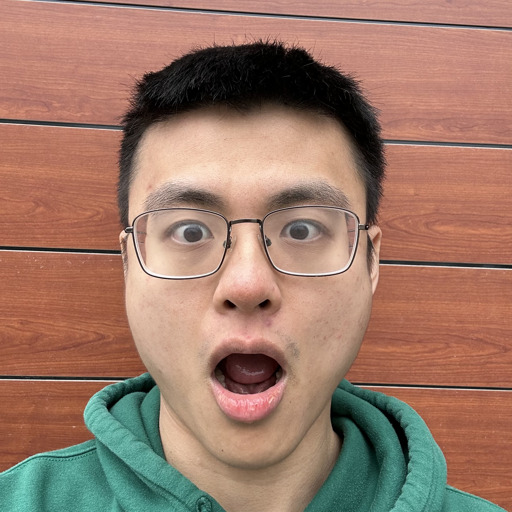} %
\end{subfigure}
\begin{subfigure}[b]{0.32\linewidth}
\includegraphics[width=\linewidth]{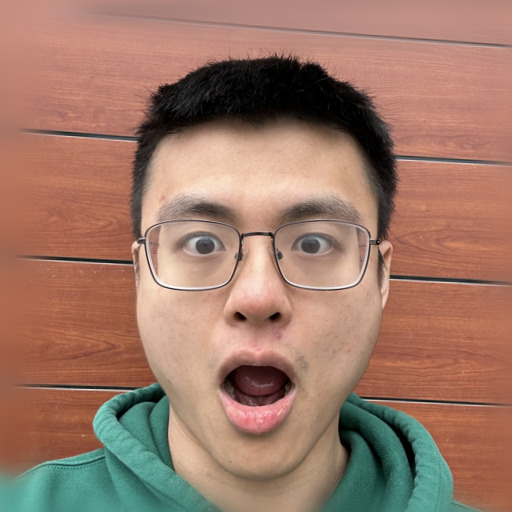} %
\end{subfigure}
\begin{subfigure}[b]{0.32\linewidth}
\includegraphics[width=\linewidth]{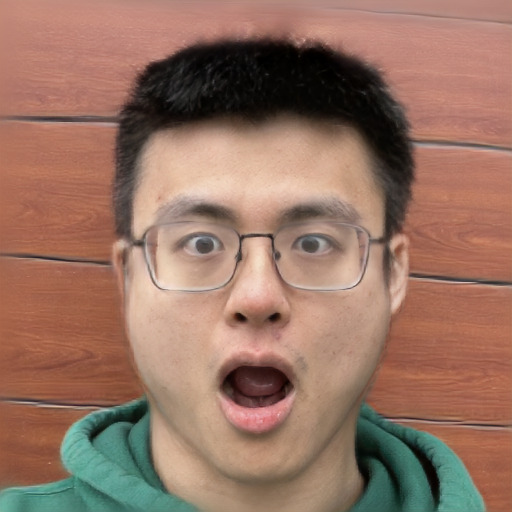} %
\end{subfigure}
\\
\begin{subfigure}[b]{0.32\linewidth}
\includegraphics[width=\linewidth]{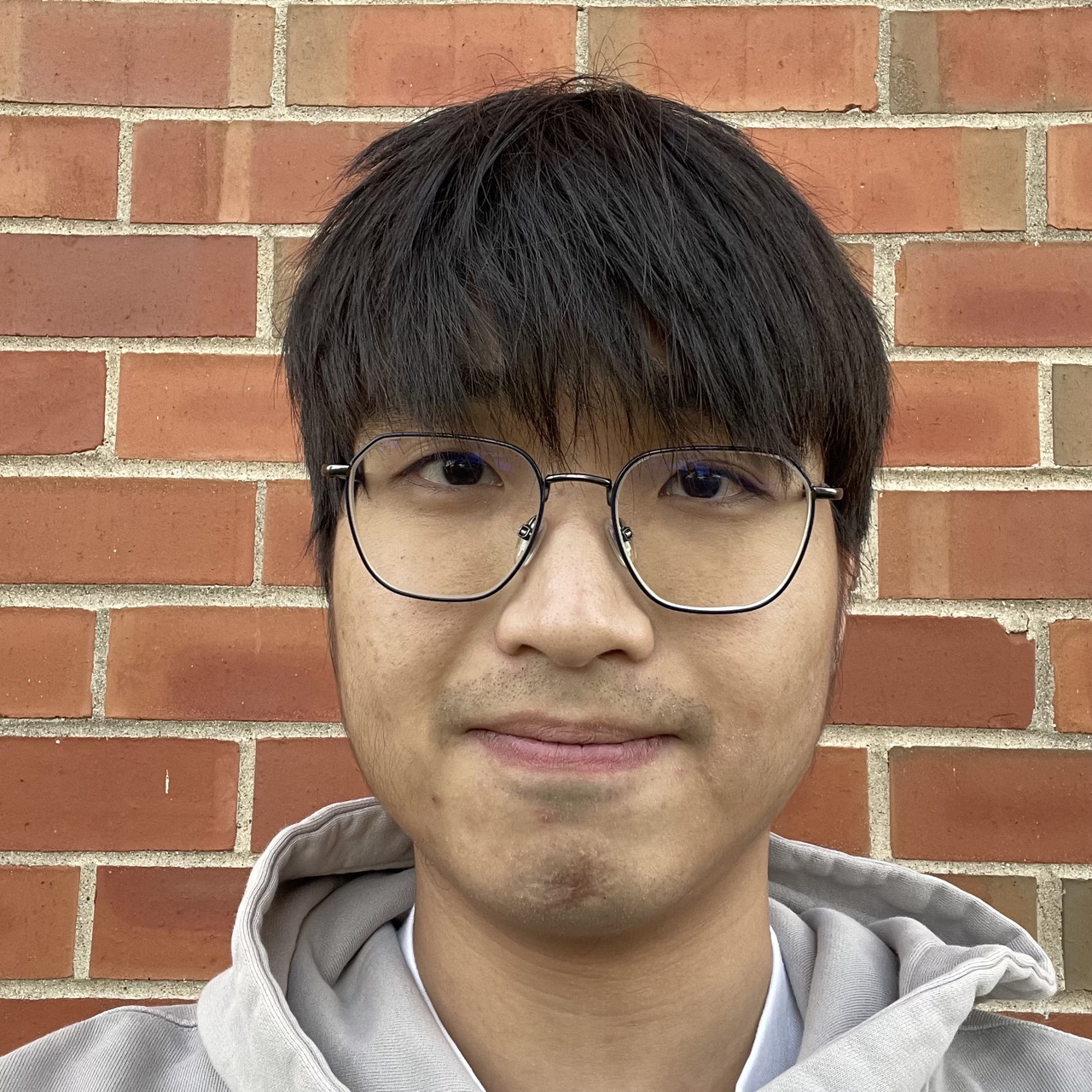} %
\caption{Face Selfie}
\end{subfigure}
\begin{subfigure}[b]{0.32\linewidth}
\includegraphics[width=\linewidth]{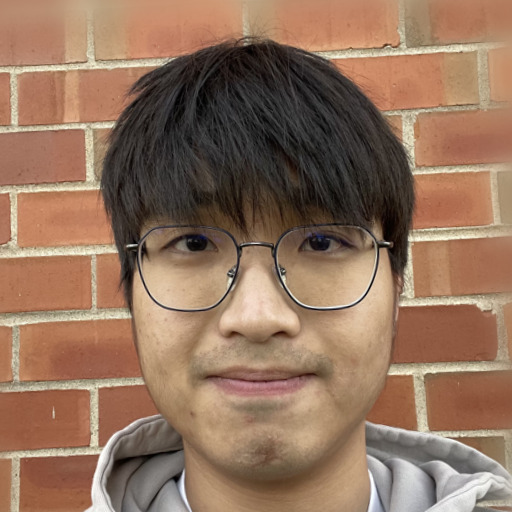} %
\caption{Aligned Face Selfie}
\end{subfigure}
\begin{subfigure}[b]{0.32\linewidth}
\includegraphics[width=\linewidth]{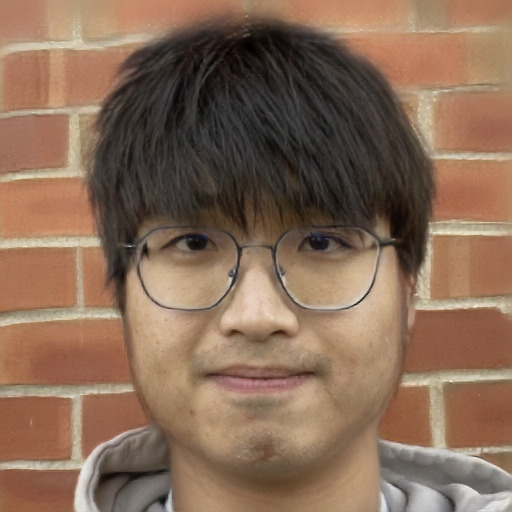} %
\caption{Undistored  Face Selfie}
\end{subfigure}
\caption{Results of our trained perspective undistortion network. Given a face selfie (left), we first align it (middle), and then correct the perspective distortion (right).  }
\label{fig:undis_out}
\end{figure}

During inference, given a face selfie $I_f$, we initially align it and subsequently utilize the fine-tuned network for perspective undistortion.
Fig. \ref{fig:undis_out} shows results of perspective undistortion on the face selfies.  Following \cite{wang2021facevid2vid}, we set learning rate as 2e-4, and batch size as 8 for training.


\section{Implementation Details}

We present the implementation details, and the code will be publicly available after acceptance. Following Stable Diffusion~\cite{lee2022high}, all images in our pipeline are square and share a consistent resolution of 512.

\subsection{Dataset Generation}

We define one training pair as $\{(S', I'_{gt} \cdot M', M'), I'_{gt}\}$, where $S'=\{I'_f, I'_u, I'_\ell, I'_s\}$ is a set of four synthetic selfies for face, upper body, lower body, and shoes respectively. $I'_{gt}$ is the ground-truth full-body image, and $M'$ is the mask indicating the region to be inpainted. 

\begin{figure}[!t]
\centering
\captionsetup[subfigure]{labelformat=empty}
\begin{subfigure}[b]{0.32\linewidth}
\includegraphics[width=\linewidth]{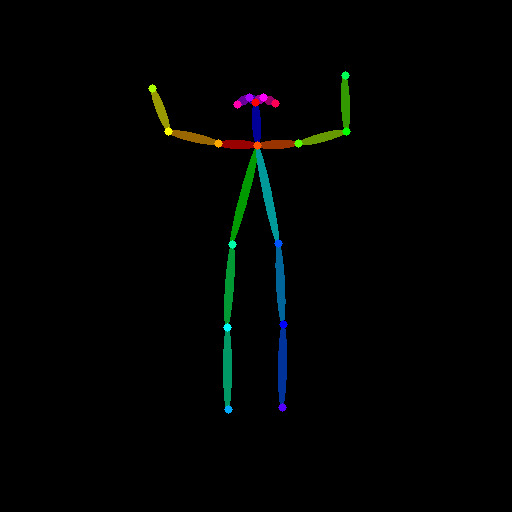} %
\end{subfigure}
\begin{subfigure}[b]{0.32\linewidth}
\includegraphics[width=\linewidth]{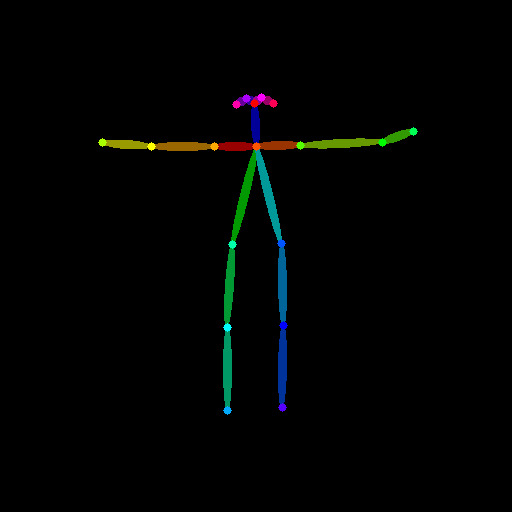} %
\end{subfigure}
\begin{subfigure}[b]{0.32\linewidth}
\includegraphics[width=\linewidth]{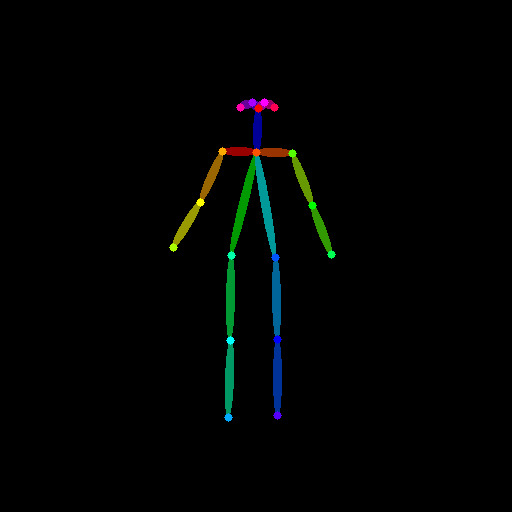} %
\end{subfigure}
\caption{Examples of OpenPose skeleton images we use to generate ground truth full body image. }
\label{fig:pose}
\end{figure}

We employ RealisticVision~\cite{Civitairv51} as the pretrained Stable Diffusion with OpenPose ControlNet v1.1~\cite{zhang2023adding} to generate $I'_{gt}$.  The guidance scale is set to 7.5, the denoising step is 20, and the ControlNet scale is 1.0. OpenPose Skeleton images, used for guidance, are detected from a subset of the Human Bodies in the Wild dataset~\cite{Shapy:CVPR:2022}.
Fig. \ref{fig:pose} illustrates three examples of these pose images. The text prompt used is ``a [gender], [place], [upper], [lower], [shoes], standing, front-facing, RAW photo, full body shot, 8k uhd, high quality, film grain". Here, [gender] can be man or woman, [place] includes common indoor and outdoor locations such as beach, park, street, restaurant, cafe, shopping mall, \etc. [upper] comprises shirt, hoodie, sweater, jacket, \etc, while [lower] includes jeans, leggings, shorts, \etc, and [shoes] covers sneakers, heels, boots, flats, etc. Additionally, we use CodeFormer~\cite{zhou2022codeformer} to enhance the details of face regions in the generated $I'_{gt}$.  

After obtaining the face-refined $I'_{gt}$, we utilize the human parsing network~\cite{yang2023humanparsing} to generate the semantic map of $I'_{gt}$. This map is then employed to extract the bounding box of the person. The bounding box is scaled up following the strategy outlined in \cite{yang2022paint} to obtain $M'$.

\begin{figure}[!t]
\centering
\captionsetup[subfigure]{labelformat=empty}
\begin{subfigure}[b]{0.32\linewidth}
\includegraphics[width=\linewidth]{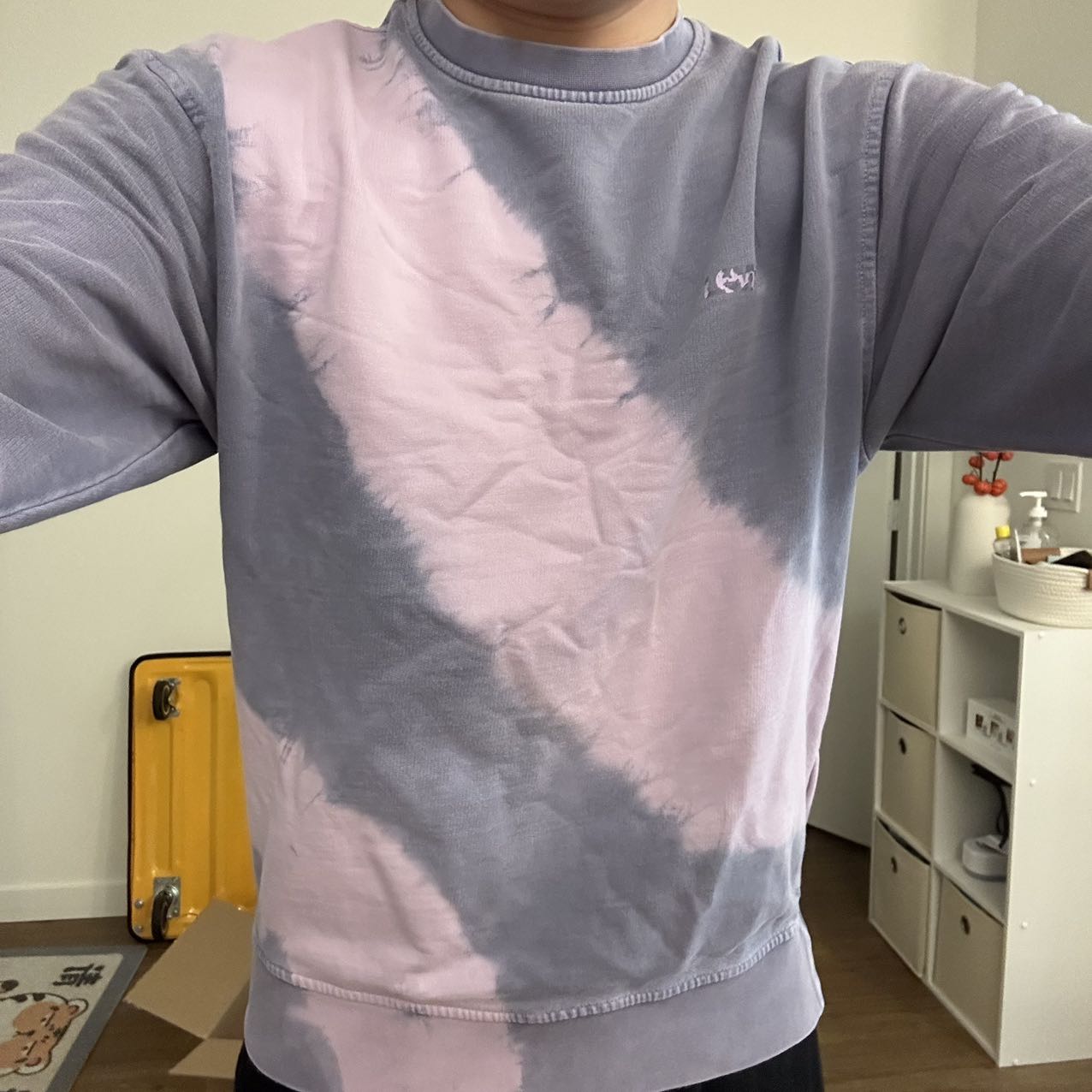} %
\caption{Upper Body}
\end{subfigure}
\begin{subfigure}[b]{0.32\linewidth}
\includegraphics[width=\linewidth]{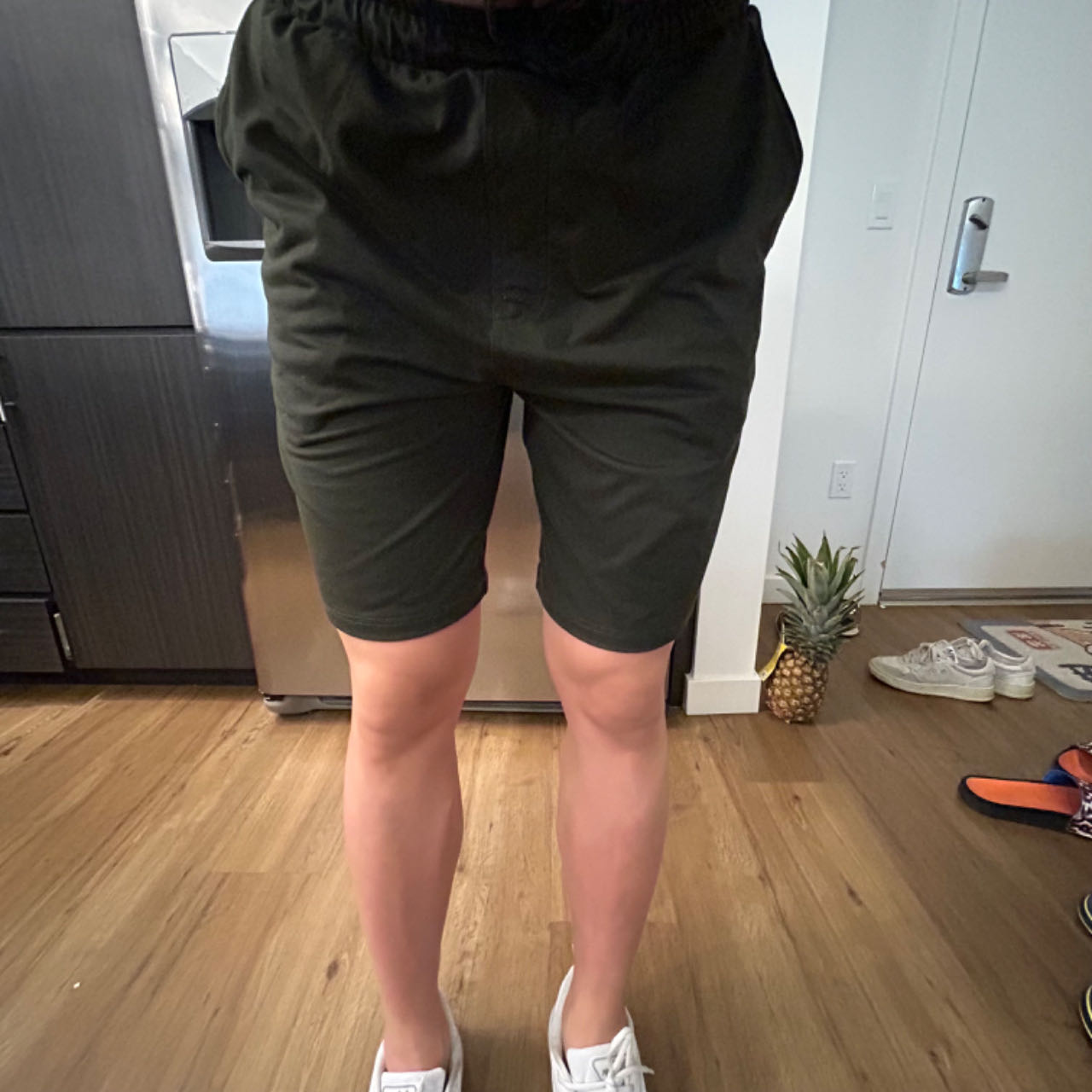} %
\caption{Lower Body}
\end{subfigure}
\begin{subfigure}[b]{0.32\linewidth}
\includegraphics[width=\linewidth]{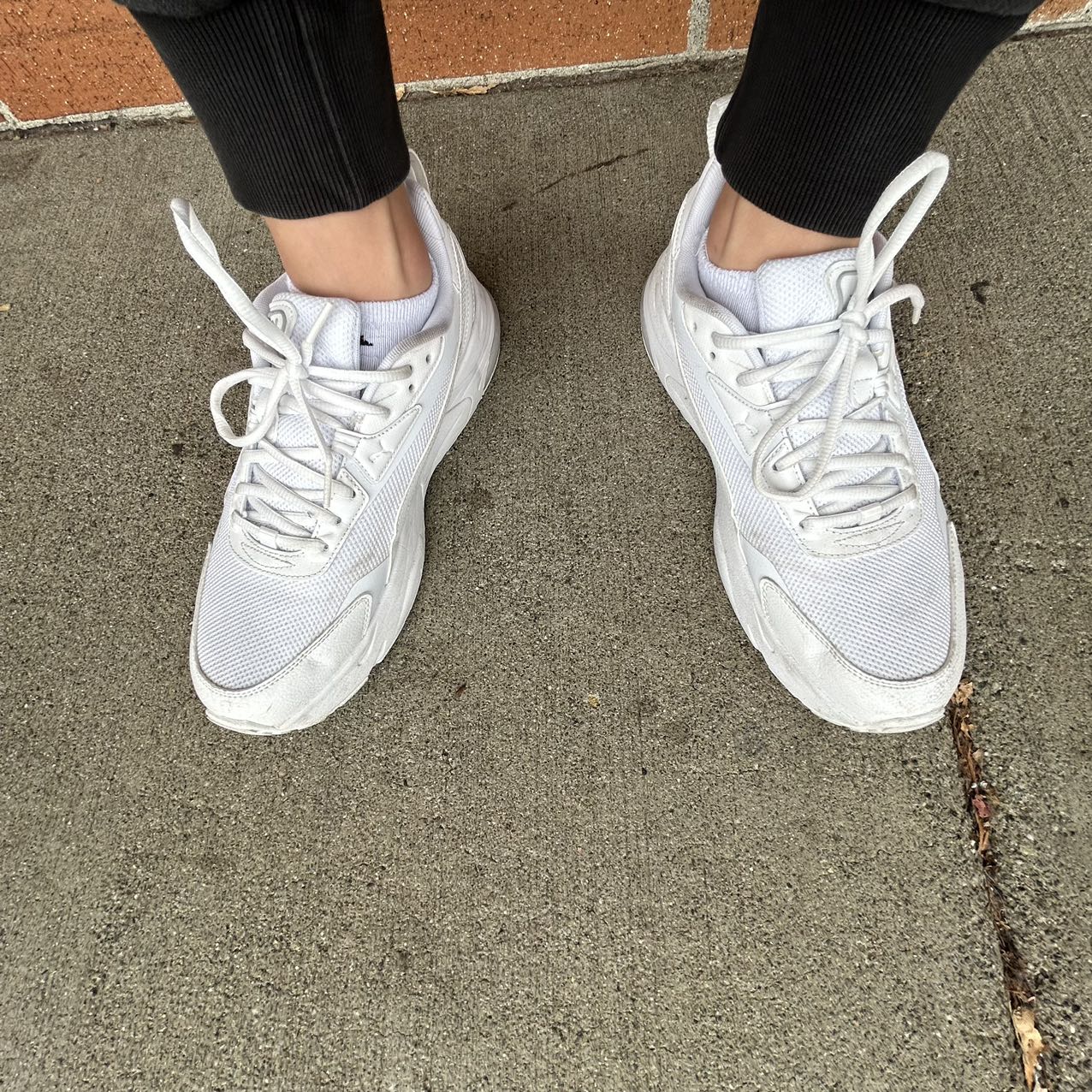} %
\caption{Shoes}
\end{subfigure}
\caption{Examples of in-the-wild selfies used to detect typical keypoints. }
\label{fig:wild_selfies}
\end{figure}

For the real selfies utilized in detecting typical keypoints for selfie simulation, we gather sets of upper body, lower body, and shoes selfies, each comprising 10 examples. Fig. \ref{fig:wild_selfies} illustrates one example from each pre-captured selfie category.


In total, we create a training dataset consisting of 39,816 pairs, encompassing diverse individuals, clothing types, poses, and backgrounds.

\subsection{Training}

We initialize all network weights using the pretrained model provided by Paint-By-Example~\cite{yang2022paint}, except for the adapted linear layer $L$ (zero-initialized).  For training, we set the learning rate as 1e-5 and batch size as 12.  We train the model for 9 epochs, taking around 36 hours on 3 NVIDIA A100 GPUs.

\subsection{Automatic Target Pose Selection}

We develop an automatic selection strategy to help obtain $I_r$ from the users' photo collection  $\Phi$.
The selection criteria are based on the similarity between the clothing types in the input selfies and a candidate image in $\Phi$.
This is because the more similar the clothing type is, the more accurately the body shape (in this particular type of outfit) can be extracted from $I_r$. 

Specifically, we begin by utilizing the pretrained human parsing model~\cite{yang2023humanparsing} to obtain the semantic map of selfies $I_u$. 
We filter out semantic labels that occupy less than 21 pixels and labels that do not belong to upper cloth (\eg, pants, face). This results in a set of upper body labels, denoted as $P_u$. The same process is applied to obtain lower body labels $P_\ell$ and shoes labels $P_s$. 
For a full-body reference photo in $\Phi$, we use the same network to obtain its semantic map. We filter out semantic labels that occupy less than 5 pixels and labels that do not belong to the upper body, lower body, and shoes. The resulting set is denoted as $P_r$. The matching score of each photo in $\Phi$ is computed by $P_r \cap  (P_u \cup P_\ell \cup P_s)$. Then we rank the candidate references based on the matching score, with higher scores indicating better matches.
Once reference photo $I_r$ is selected, we obatin inpainting mask $M$ using the bounding box of $I_r$, scaled up by 1.1 times by default.

\begin{figure}[!t]
\centering
\begin{subfigure}[b]{0.24\linewidth}
\includegraphics[width=\linewidth]{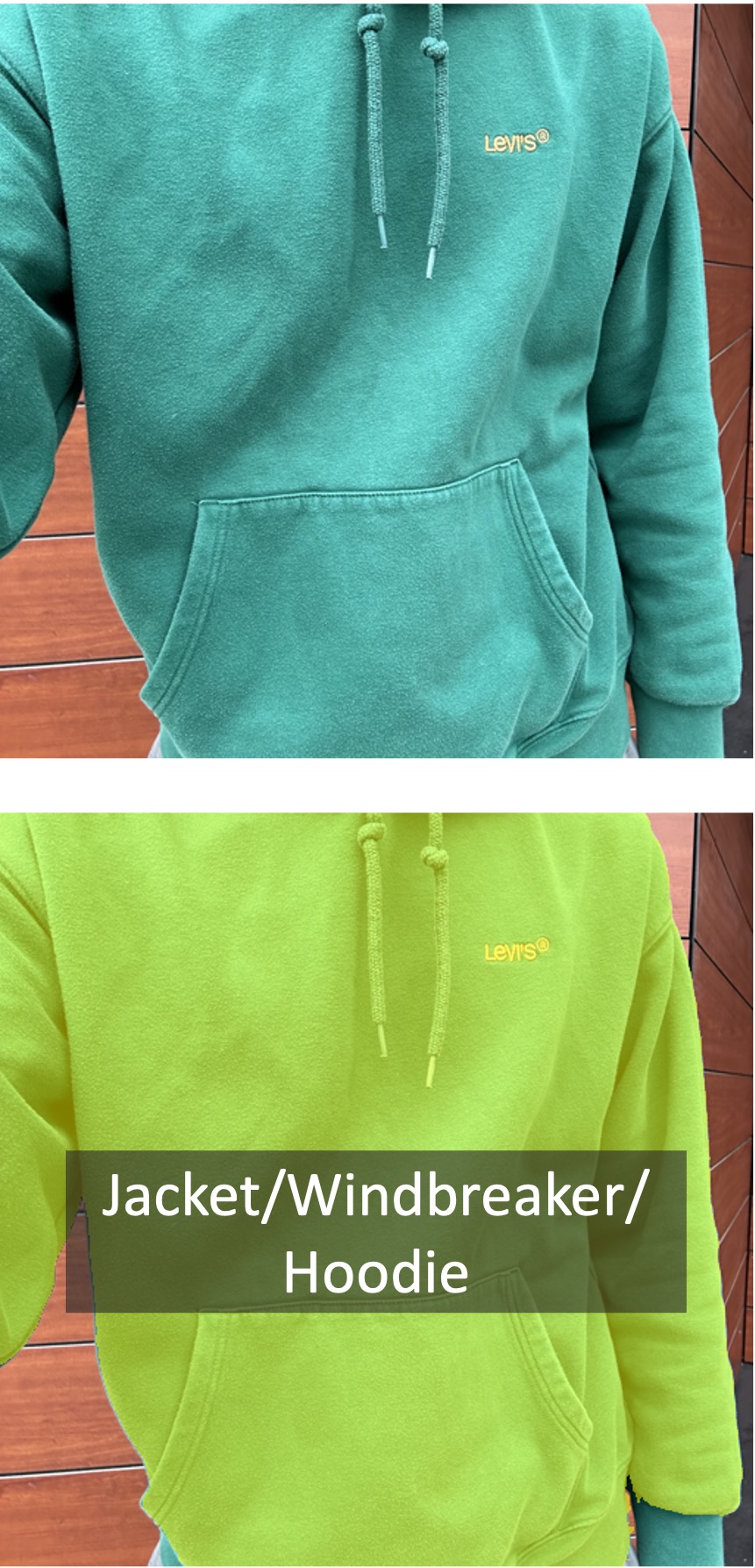} %
\caption{Upper Body}
\end{subfigure}
\begin{subfigure}[b]{0.24\linewidth}
\includegraphics[width=\linewidth]{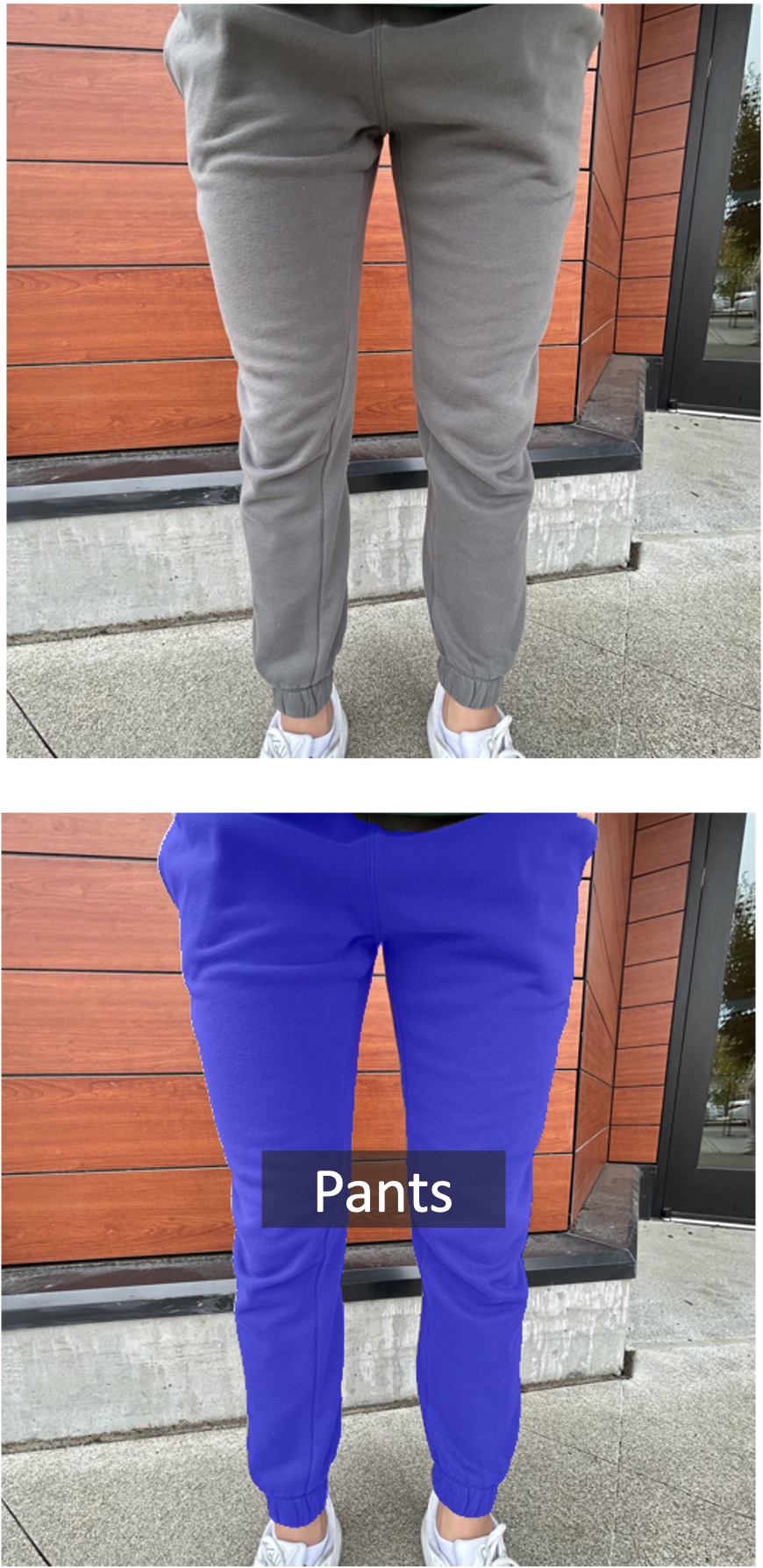} %
\caption{Lower Body}
\end{subfigure}
\begin{subfigure}[b]{0.24\linewidth}
\includegraphics[width=\linewidth]{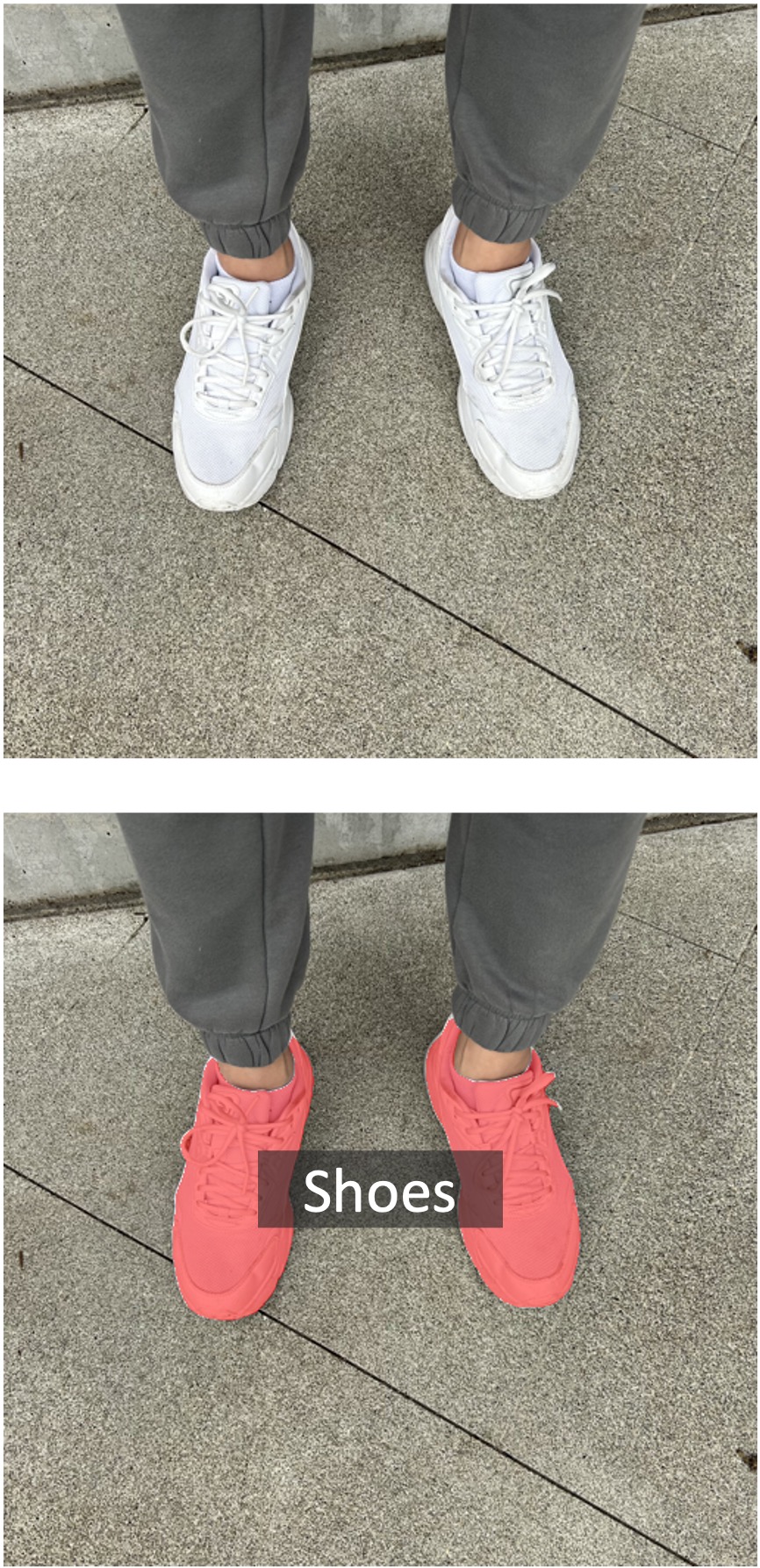} %
\caption{Shoes}
\end{subfigure}
\begin{subfigure}[b]{0.24\linewidth}
\includegraphics[width=\linewidth]{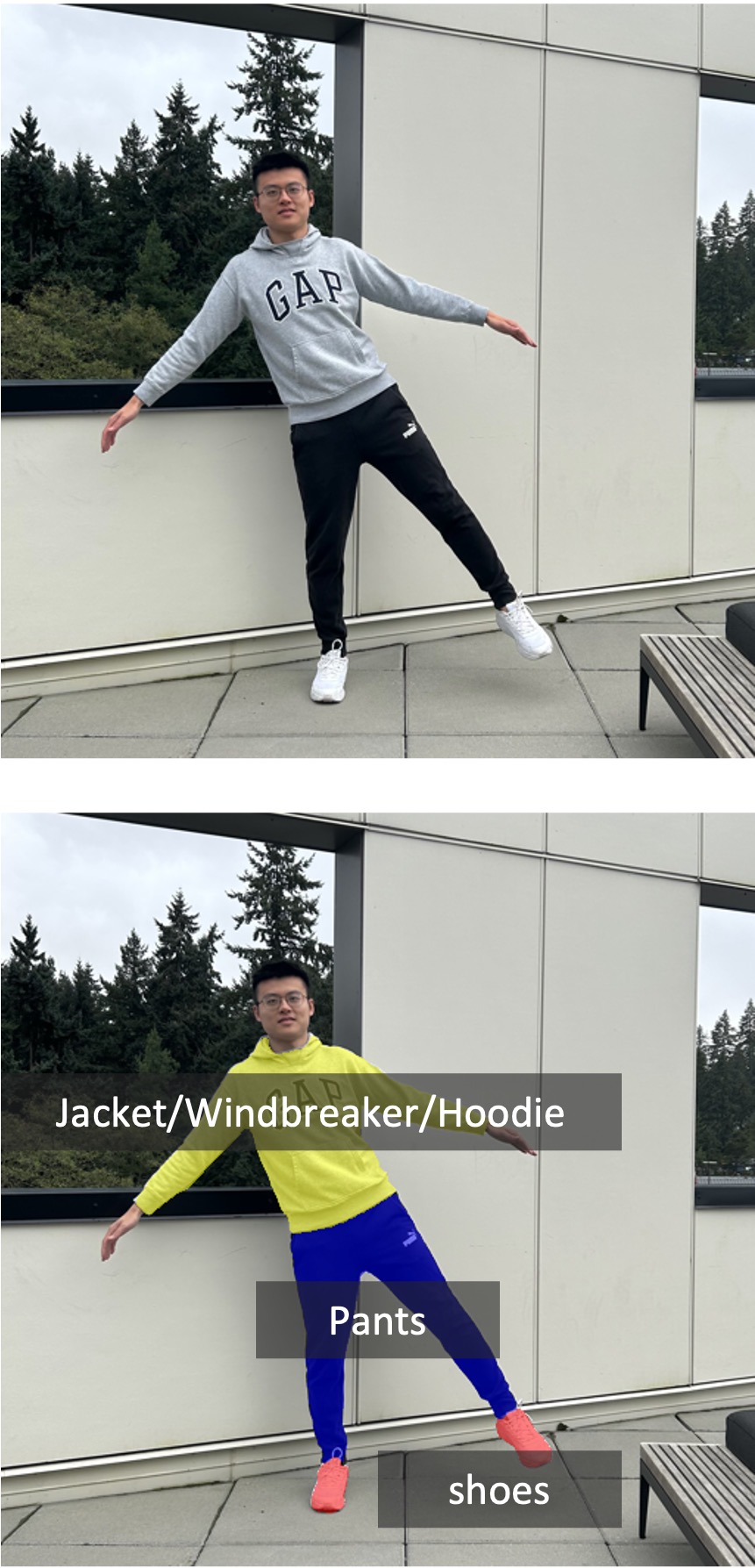} %
\caption{Ref Photo}
\end{subfigure}
\caption{Visualization of automatic target pose selection. The top row shows the input selfies and candidate reference photos, and the bottom row shows their segmentation results.  }
\label{fig:auto_pose}
\end{figure}

Fig. \ref{fig:auto_pose} (a) to (d) visualizes an example of the semantic map of selfies and the selected reference photo.

\subsection{Pose-Guided Generation}

For generation, we set $T=50$ and use DDIM scheduler for denoising.  The ControlNet scale is set to 1.0. In cases where the target pose involves spreading arms, the mask  $M$ might sometimes become too large, potentially leading to the failure of preserving background content in  $I_b$. To help alleviate this issue, we apply the following strategy during the denoising process. We dilate the foreground mask (the finer mask containing only the human body) by 21 pixels, denoted as  $\bar{M}$.  Then, at each denoising timestep $t$, we compute the denoised latent as:
\begin{equation}
z_{t-1}=
\begin{cases}
z_{t-1}^f,   \quad & \text{if } t \leq s T \\
z_{t-1}^f \cdot \bar{M} + z_{t-1}^b \cdot (1-\bar{M}), \quad & \text{if } t > s T\end{cases},
\label{soft_blending}
\end{equation}
where $z^f_{t-1}$ is the foreground latent obtained using the selfie-conditioned inpainting model (following the same process discussed in the main paper).  $z^b_{t-1}$ is the background latent obtained by adding noise to $I_b$ by $t-1$ steps using DDIM scheduler.   We set $s=0.4$. This enables the generation of details in the surrounding area (\eg, shadows) based on the inpainting model in later timesteps (smaller $t$), while reasonably preserving background content in the earlier timesteps  (larger $t$).

\subsection{Fine-Tuning}

We generate a ``ground truth'' for fine-tuning by resizing and placing a randomly selected selfie image from the set $S$ into the mask region $M$ of the background image $I_b$. 
The full-body inpainting prior learned by the trained model is used to determine where and how to place the selected selfie image into the masked $I_b$.

Specifically, we first generate full body selfie $I_n$ without any pose as condition. This is achieved by using the same process as Pose-Guided Generation but omitting the modified ControlNet.
Suppose the upper body selfie $I_u$ is selected for augmentation. We extract the bounding box of the upper body in $I_n$ based on the semantic map of $I_n$ detected by the human parsing network~\cite{yang2023humanparsing}.   Then we resize $I_u$ to have the same height as this bounding box. The resize operation keeps the aspect ratio of $I_u$ unchanged to avoid using an image with the wrong scale.
Finally, we paste this resized $I_u$ to the masked $I_b$, ensuring the center of resized $I_u$ and the bounding box are the same. 

In practice, we generate 20 different $I_n$ as the candidate pool for augmentation. We then repeat the above augmentation process (resizing and pasting) 200 times, each time with $I_n$ randomly chosen from the candidate pool, resulting in a dataset of 200 augmented images. 
For fine-tuning, we set the learning rate to 5e-6 and the batch size to 4. The model is fine-tuned for 400 steps, taking around 10 minutes on a single NVIDIA A40 GPU.

\subsection{Appearance Refinement}

To train the DreamBooth (with two concepts) using only one image for each concept, we need to augment them to avoid overfitting.
Specifically, we randomly resize the face selfie $I_f$ from a resolution of 350 to 450 and apply random zero-padding to create the augmented image with a resolution of 512. The same operation is performed for the shoes selfie $I_s$, but with resizing resolution from 400 to 500. 
We generate 50 augmented images for $I_f$ and $I_s$ respectively.  
Then, we train a DreamBooth with two concepts using these two kinds of augmented images.
Specifically, we set the training text prompt for face  and shoes as ``a sks face'' and ``a hta shoes'' respectively.  We set the learning rate as 5e-6, batch size as 4, and fine-tune the RealisticVision Stable Diffusion model for 300 epochs, taking around 5 minutes on a single NVIDIA A40 GPU.

To perform refinement, we use the pretrained human parsing model~\cite{yang2023humanparsing} to obatin the face (shoes) region in $I_o$, and use SDEdit based on trained DreamBooth to edit the face (shoes).  We then paste the cropped, edited image back onto $I_o$.  The same process is applied (using the pretrained Stable Diffusion model) for the hands regions (left and right hands) since hands are often invisible or incomplete in selfies.




\section{Experiments}

\subsection{Dataset}
We provide details on the data capture process for our pipeline. In a specific site, the user is requested to capture five square images: face, upper body, lower body, shoes, and background.
Firstly, the user takes a face and upper body photos using the front camera, holding the camera with either one or two hands. Subsequently, the user switches to the rear camera to capture the lower body, shoes, and background photos. The entire process usually takes less than 20 seconds.

To obtain the real photo as the "ground truth," we have another person take a full-body photo of the user in a desired pose. This process should ensure that the real photos maintain the same clothing, nearly identical facial expressions, and background.

\subsection{Results}
\begin{figure*}[!h]
\centering
{\includegraphics[width=0.20\linewidth]{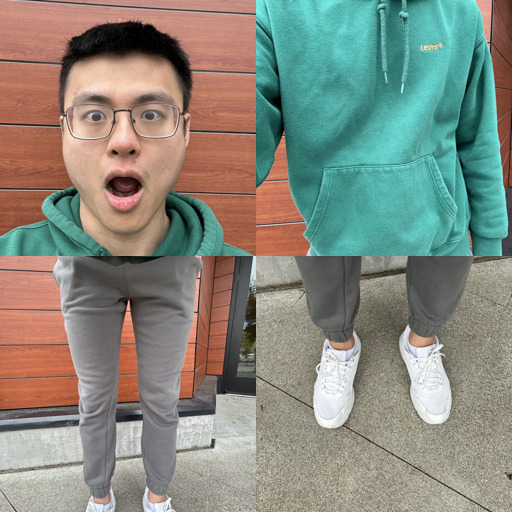}}
{\includegraphics[width=0.20\linewidth]{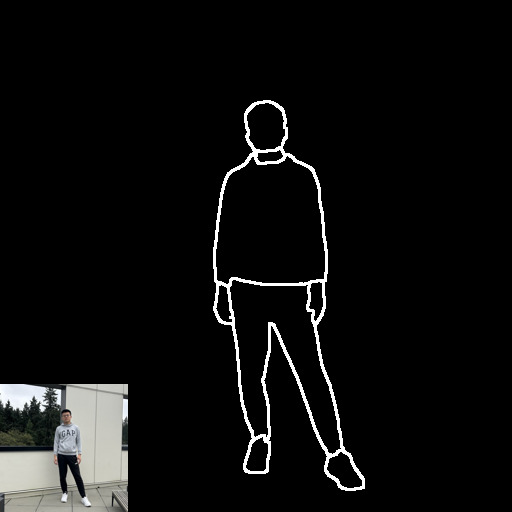}}
{\includegraphics[width=0.20\linewidth]{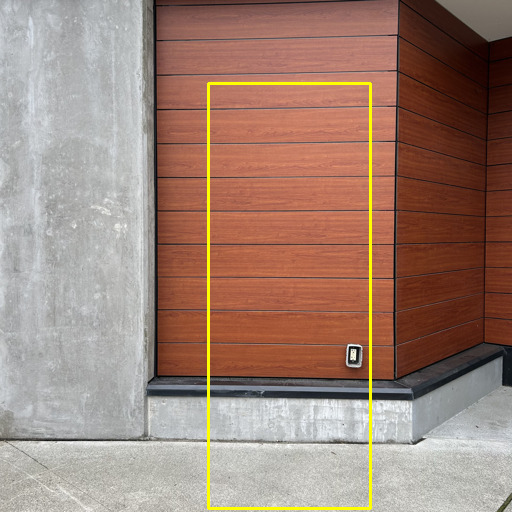}}
{\includegraphics[width=0.20\linewidth]{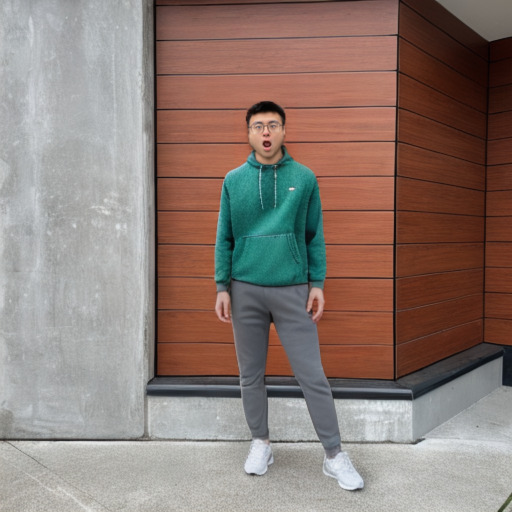}}
\\
{\includegraphics[width=0.20\linewidth]{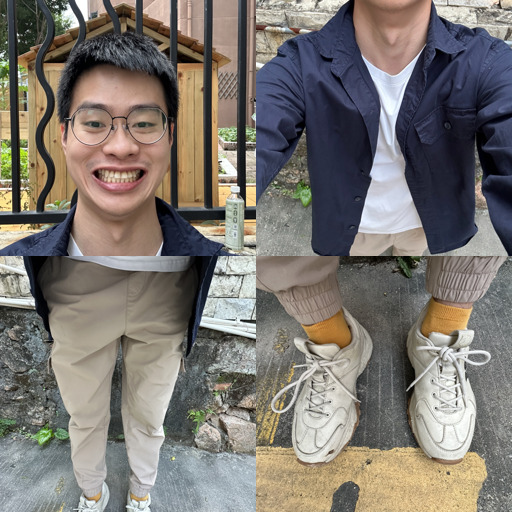}}
{\includegraphics[width=0.20\linewidth]{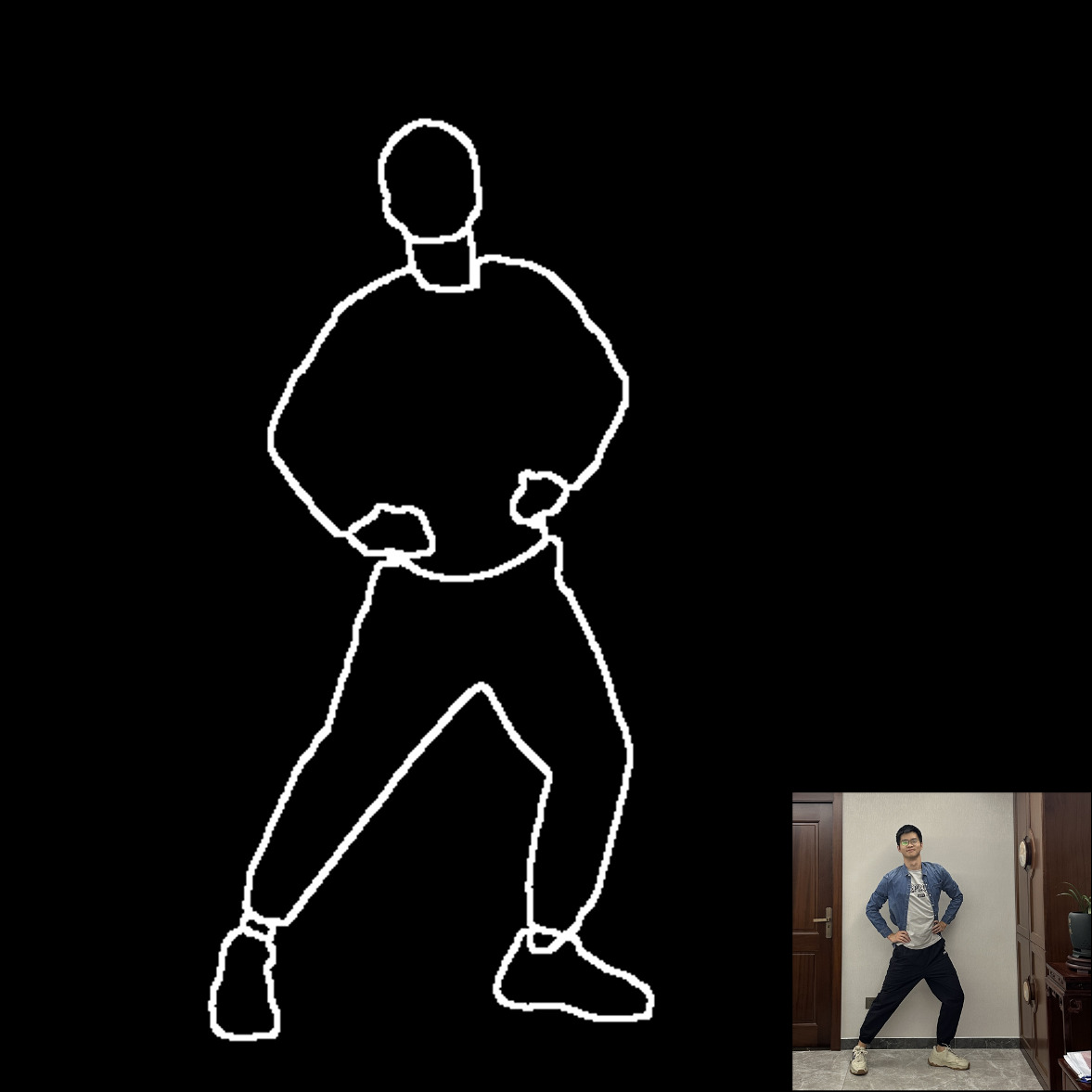}}
{\includegraphics[width=0.20\linewidth]{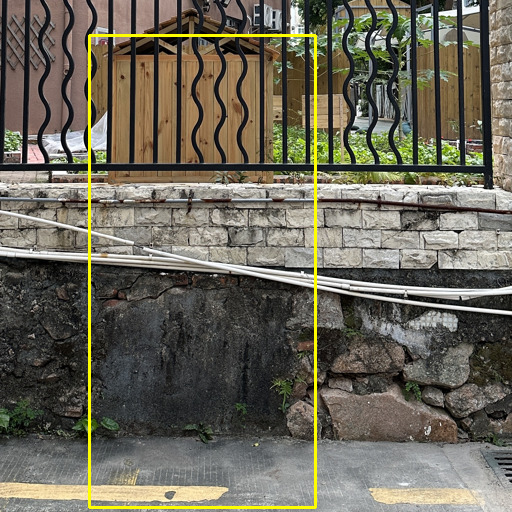}}
{\includegraphics[width=0.20\linewidth]{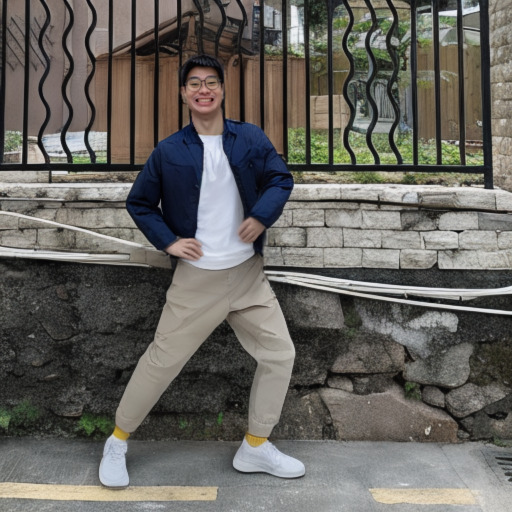}}
\\
{\includegraphics[width=0.20\linewidth]{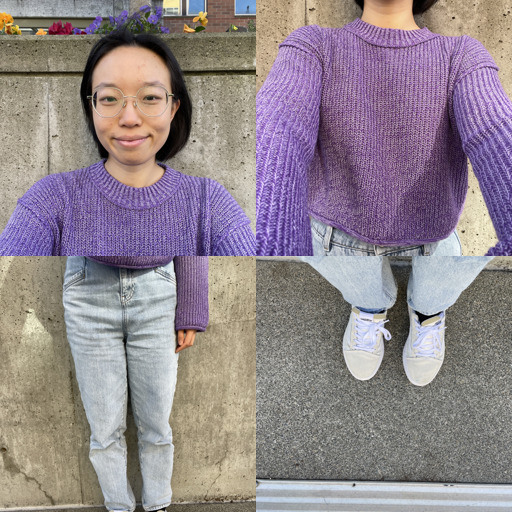}}
{\includegraphics[width=0.20\linewidth]{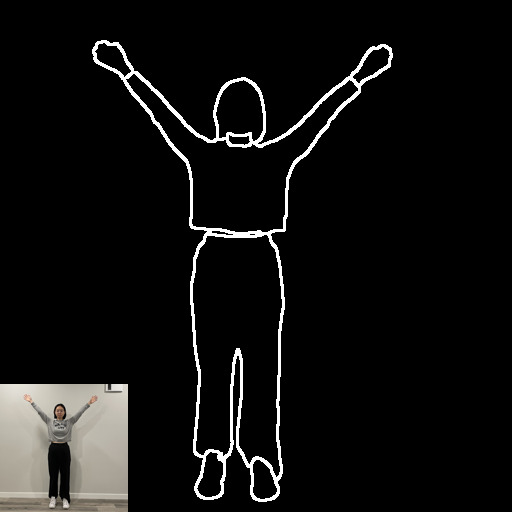}}
{\includegraphics[width=0.20\linewidth]{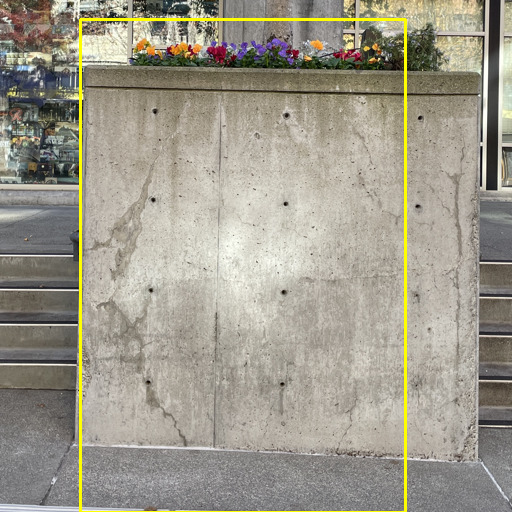}}
{\includegraphics[width=0.20\linewidth]{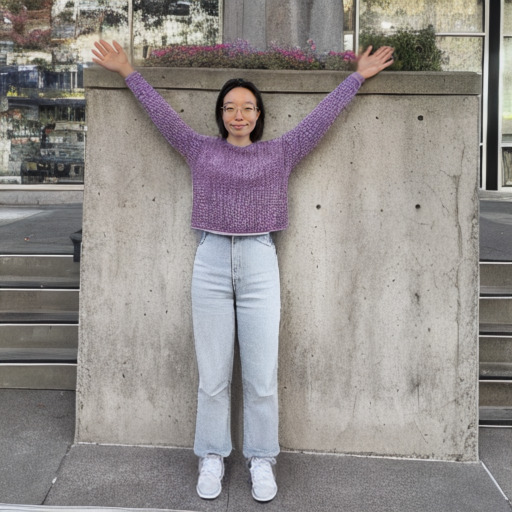}}
\\
{\includegraphics[width=0.20\linewidth]{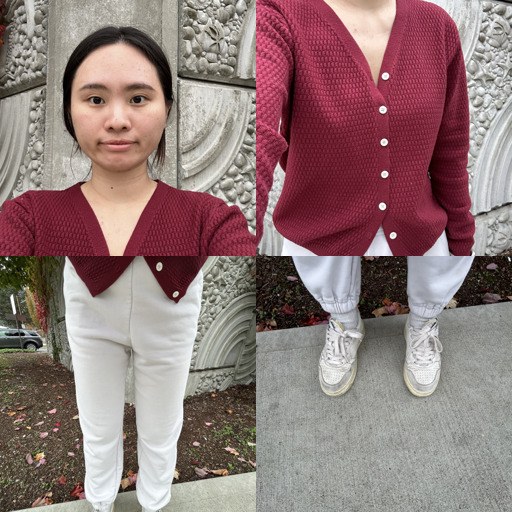}}
{\includegraphics[width=0.20\linewidth]{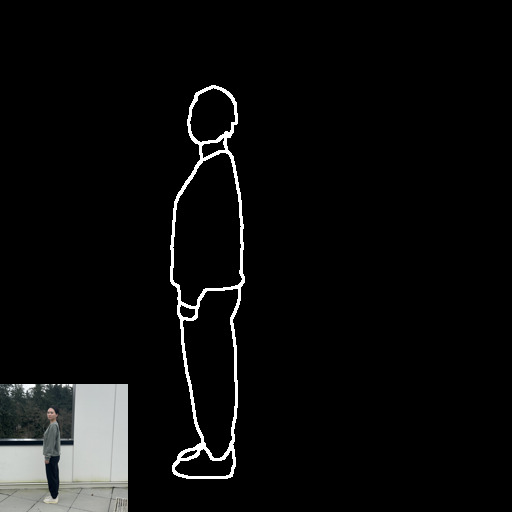}}
{\includegraphics[width=0.20\linewidth]{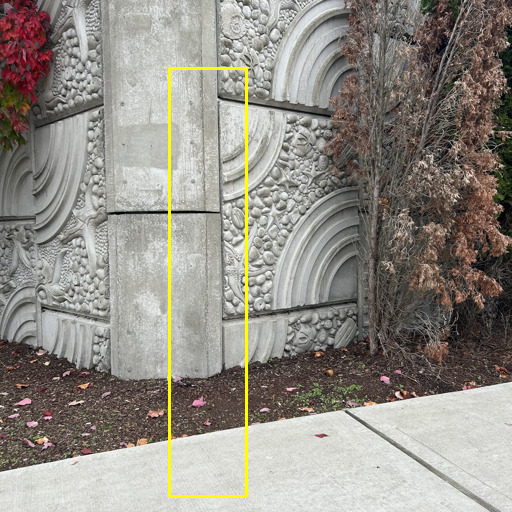}}
{\includegraphics[width=0.20\linewidth]{Total Selfie CVPR camera 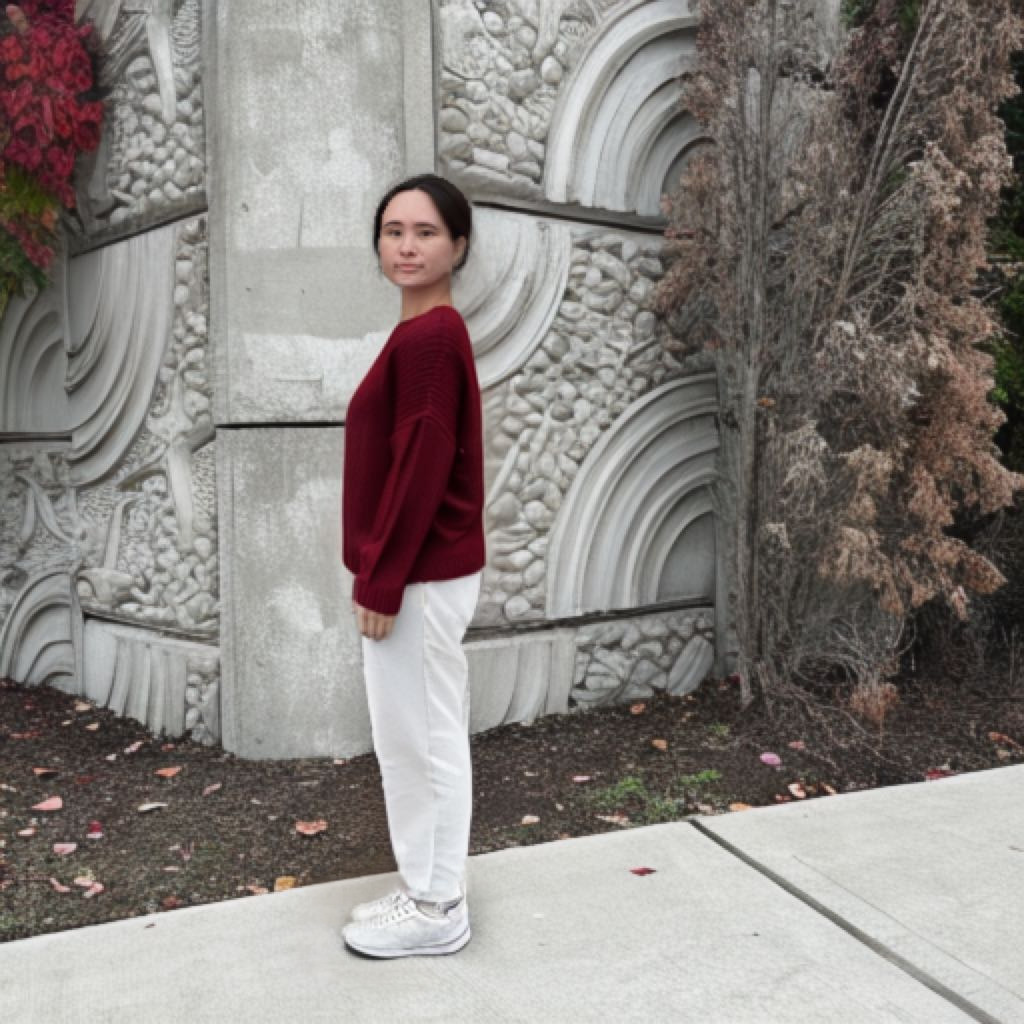}}
\\
{\includegraphics[width=0.20\linewidth]{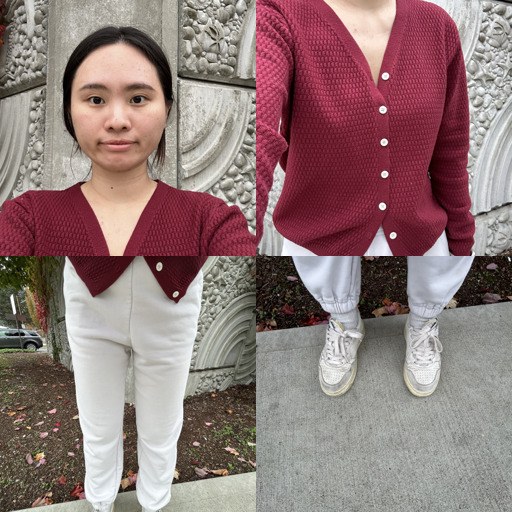}}
{\includegraphics[width=0.20\linewidth]{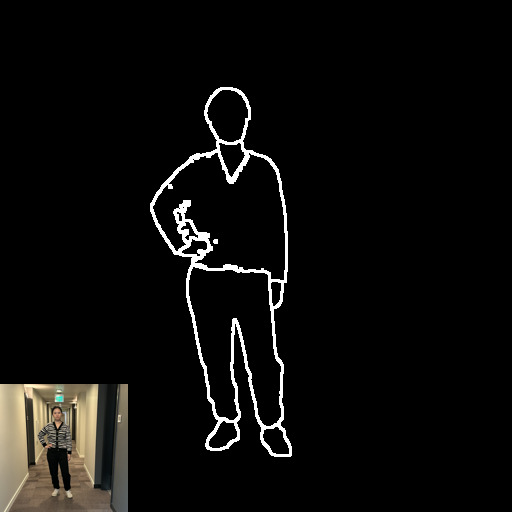}}
{\includegraphics[width=0.20\linewidth]{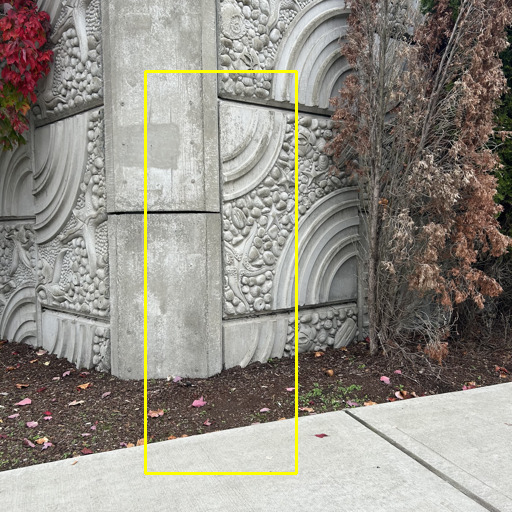}}
{\includegraphics[width=0.20\linewidth]{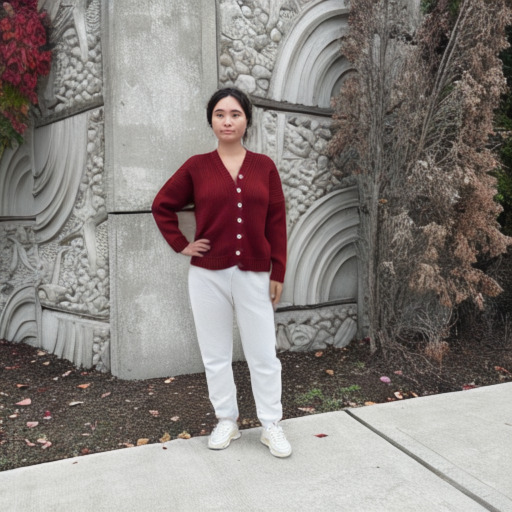}}
\\
  \subcaptionbox{ Input Selfies}%
{\includegraphics[width=0.20\linewidth]{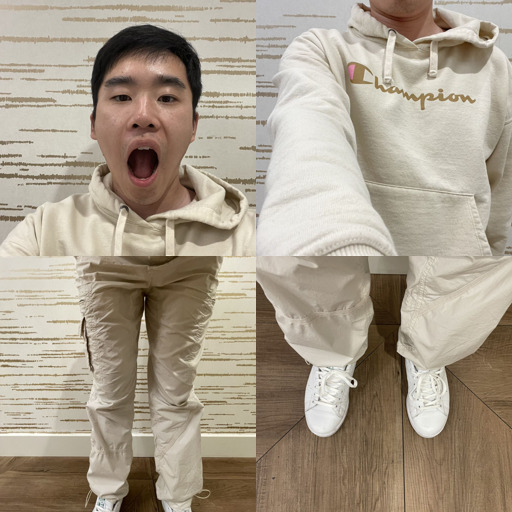}}
    \subcaptionbox{Target Pose }%
{\includegraphics[width=0.20\linewidth]{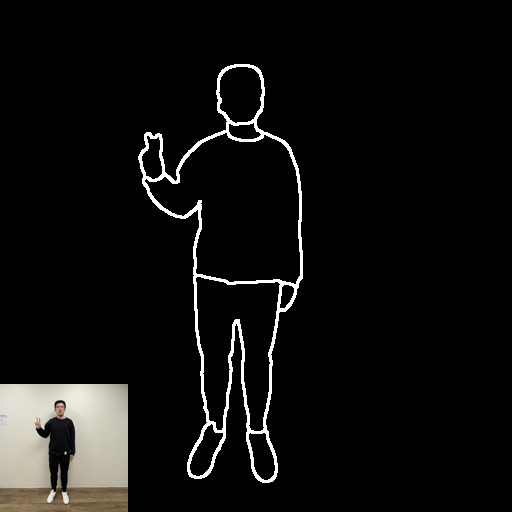}}
  \subcaptionbox{ Masked Background}%
{\includegraphics[width=0.20\linewidth]{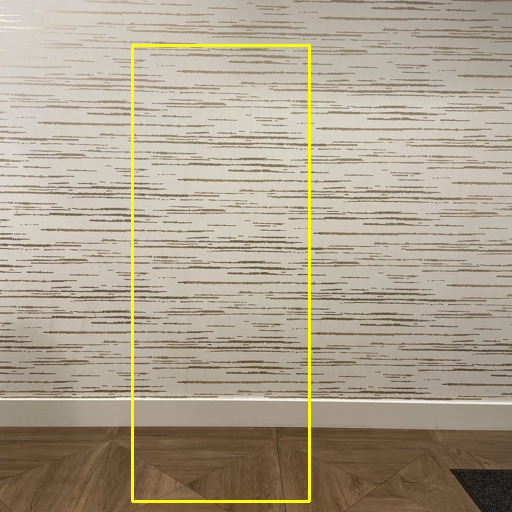}}
    \subcaptionbox{Total Selfie}%
{\includegraphics[width=0.20\linewidth]{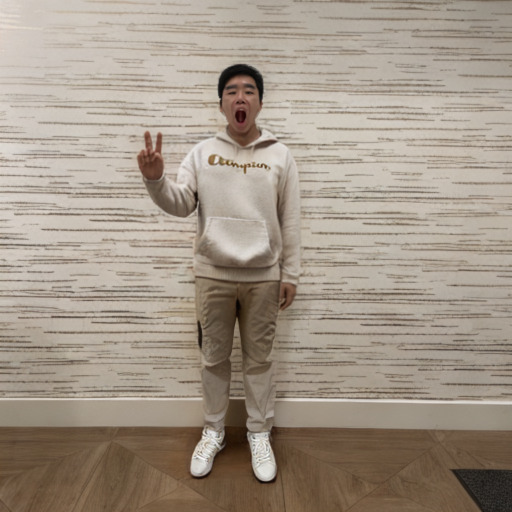}}
\caption{Results. The second column shows the Canny Edge images detected from reference images (shown as insets). Regions inside yellow box of (c) are the masked regions.
Total Selfie generates realistic, full-body images of different individuals with diverse poses and expressions against a variety of backgrounds, while preserving facial expression and clothing. 
}
  \label{fig:our_results_supp}
\end{figure*}

We show more results of Total Selfie in Fig. \ref{fig:our_results_supp}.  Total Selfie can produce high-quality full-body shots in diverse backgrounds, poses, outfits, and expressions, all while maintaining reasonable shading and composition.

\subsection{Ablation Study}
\begin{figure*}[!h]
\centering
{\includegraphics[width=0.13\linewidth]{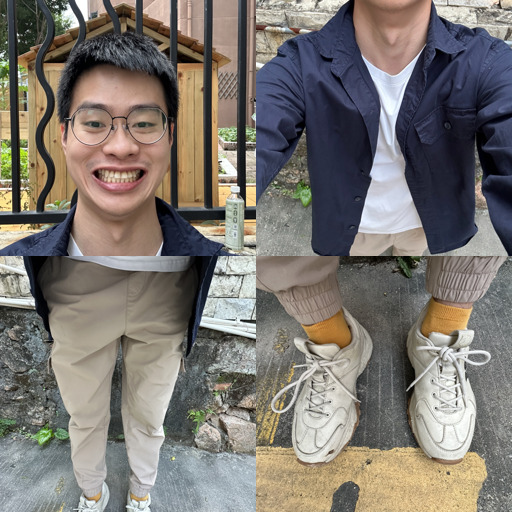}}
{\includegraphics[width=0.13\linewidth]{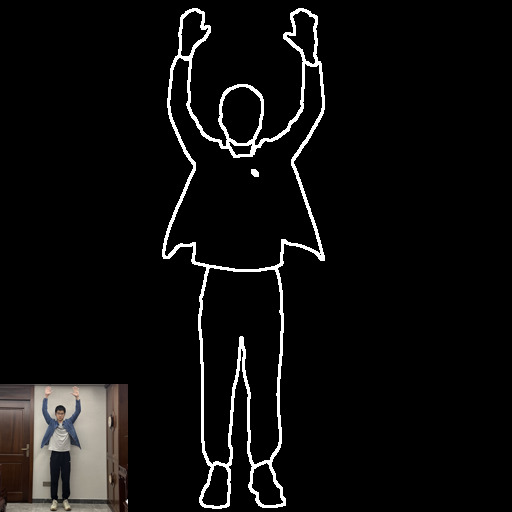}}
{\includegraphics[width=0.13\linewidth]{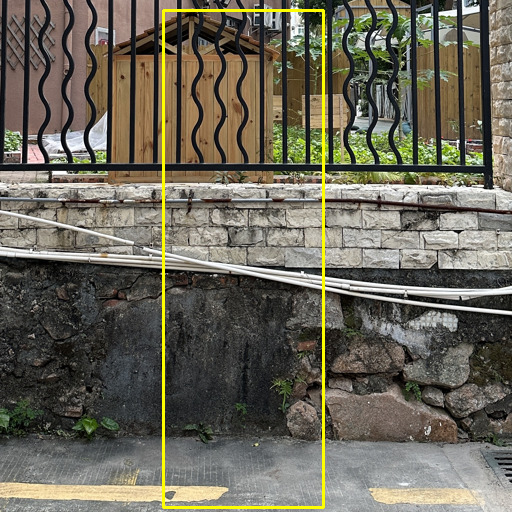}}
{\includegraphics[width=0.13\linewidth]{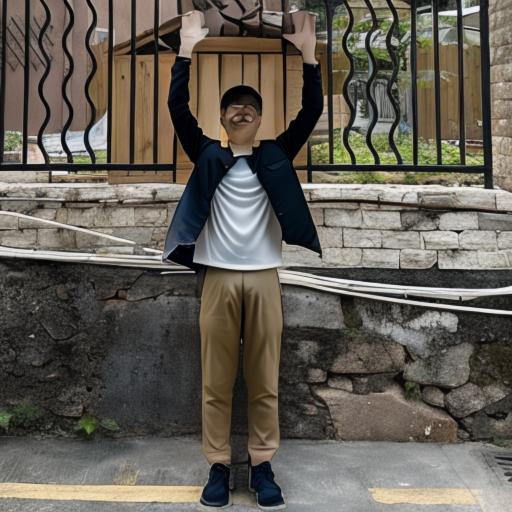}}
{\includegraphics[width=0.13\linewidth]{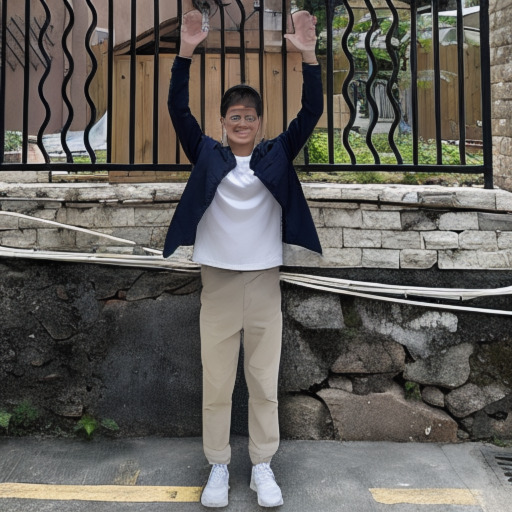}}
{\includegraphics[width=0.13\linewidth]{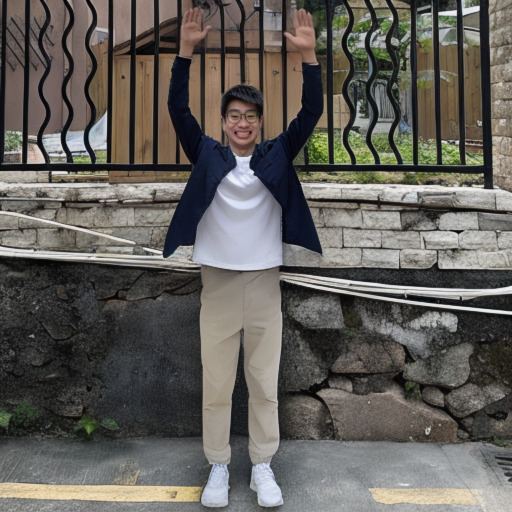}}
{\includegraphics[width=0.13\linewidth]{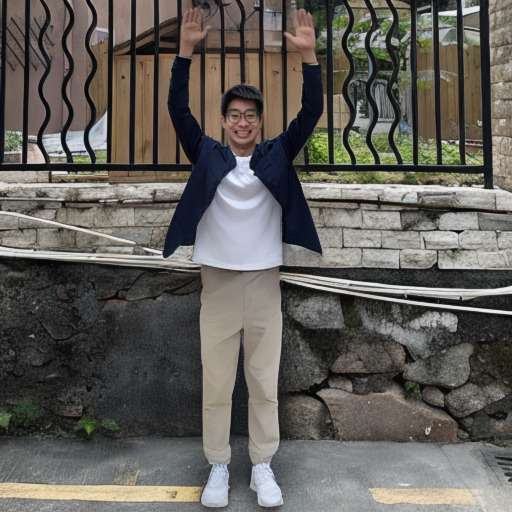}}
\\
{\includegraphics[width=0.13\linewidth]{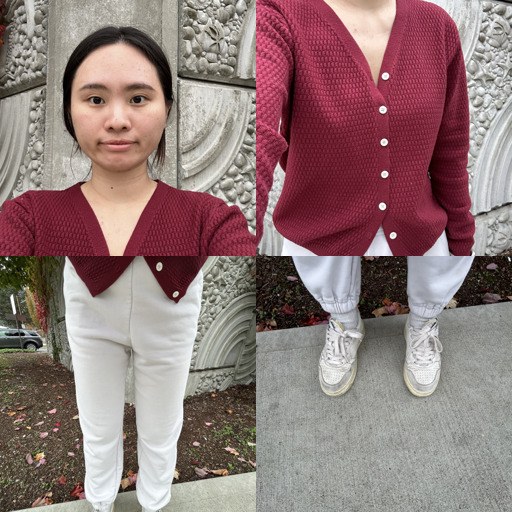}}
{\includegraphics[width=0.13\linewidth]{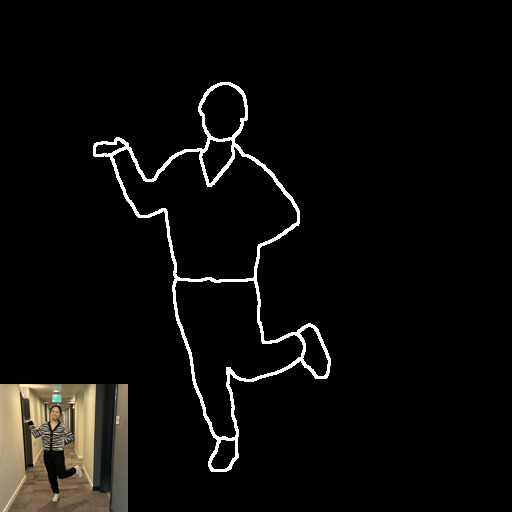}}
{\includegraphics[width=0.13\linewidth]{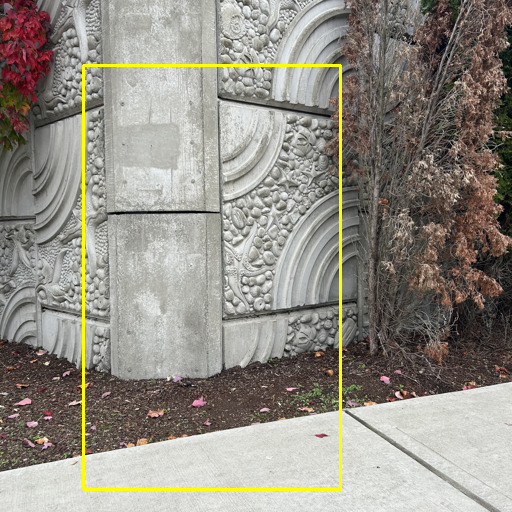}}
{\includegraphics[width=0.13\linewidth]{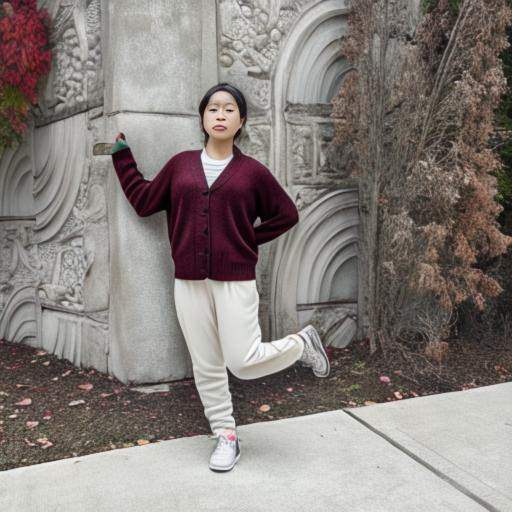}}
{\includegraphics[width=0.13\linewidth]{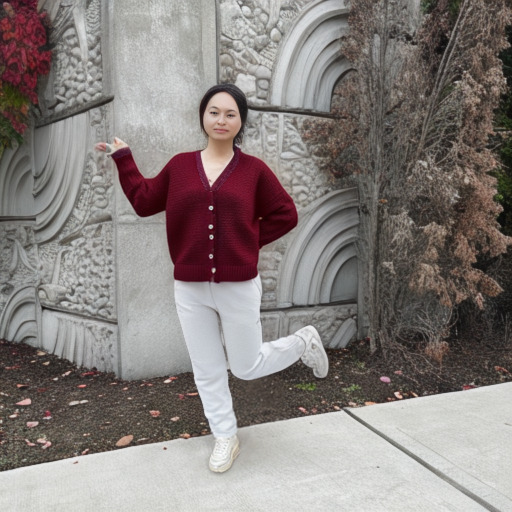}}
{\includegraphics[width=0.13\linewidth]{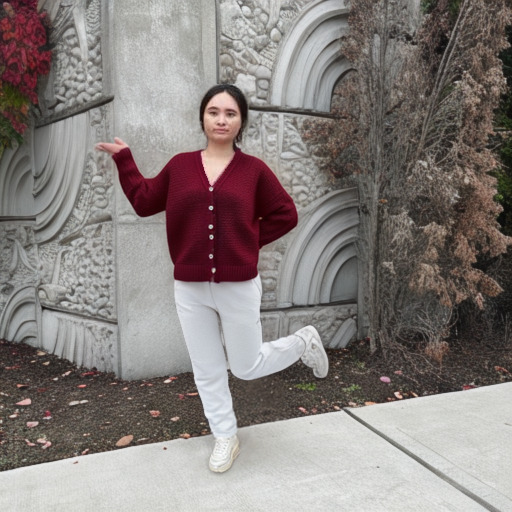}}
{\includegraphics[width=0.13\linewidth]{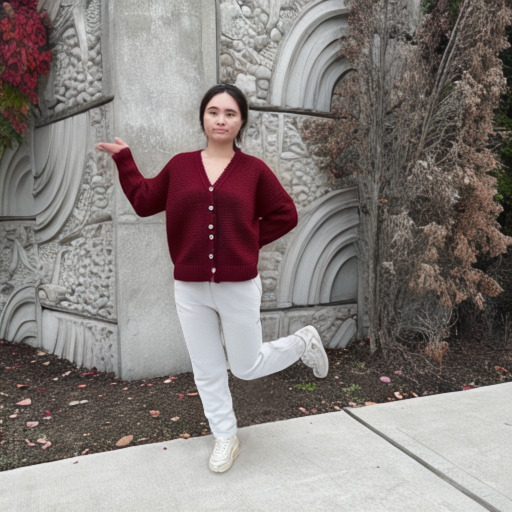}}
\\
  \subcaptionbox{ Input Selfies}%
{\includegraphics[width=0.13\linewidth]{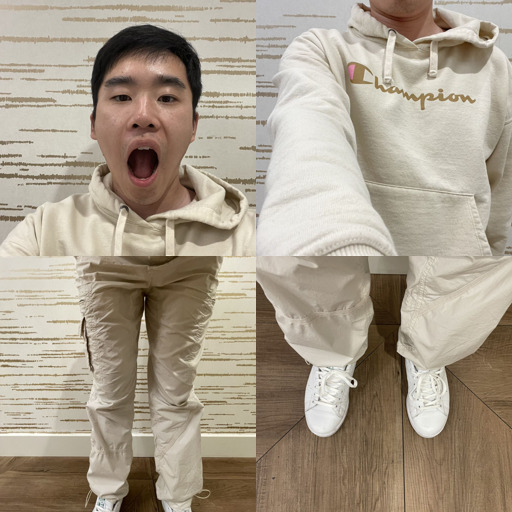}}
    \subcaptionbox{Target Pose }%
{\includegraphics[width=0.13\linewidth]{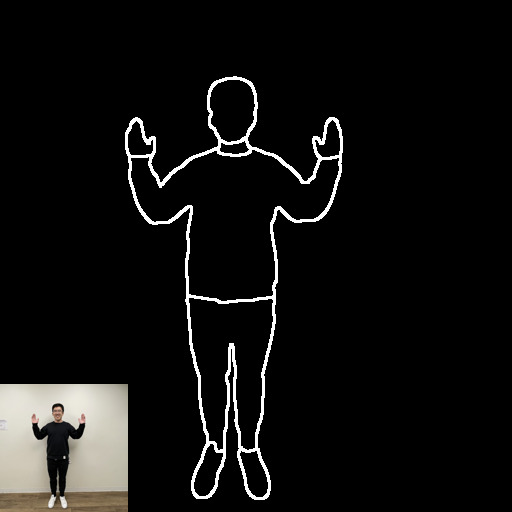}}
    \subcaptionbox{Background }%
{\includegraphics[width=0.13\linewidth]{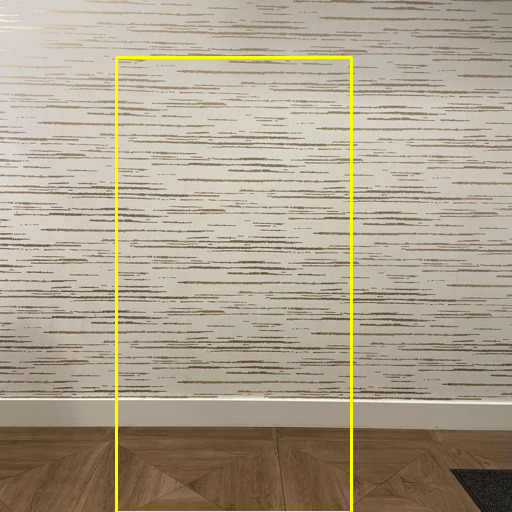}}
  \subcaptionbox{Ours-FT-AR}%
{\includegraphics[width=0.13\linewidth]{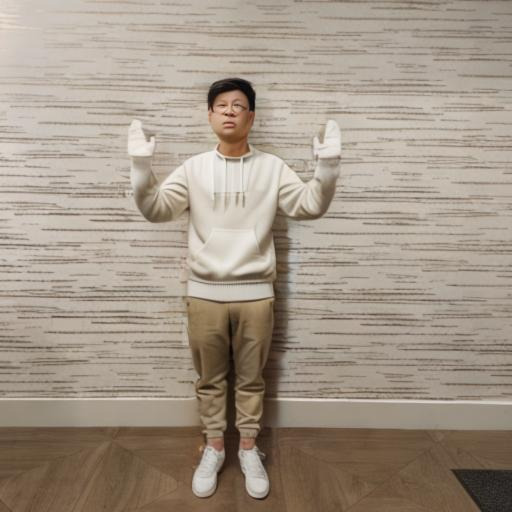}}
  \subcaptionbox{Ours-AR}%
{\includegraphics[width=0.13\linewidth]{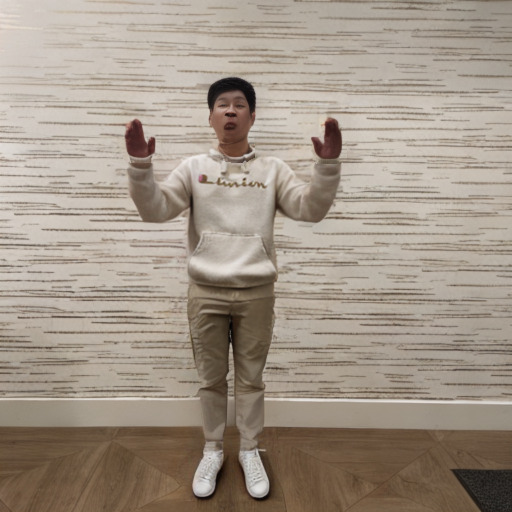}}
  \subcaptionbox{Ours-FU}%
{\includegraphics[width=0.13\linewidth]{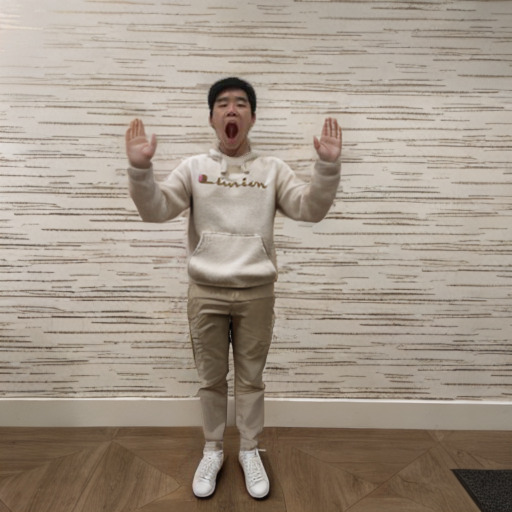}}
  \subcaptionbox{Ours}%
{\includegraphics[width=0.13\linewidth]{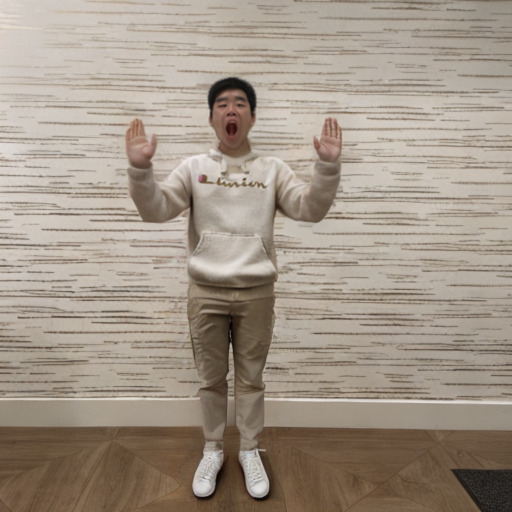}}
\caption{
Results for different modules of our pipeline. The Canny Edge image in (b) is detected from the reference image, inset. Regions inside the bounding box (c) are to be inpainted. Generating without fine-tuning and appearance refinement (d) produces an inaccurate outfit and identity. Through fine-tuning, the pipeline (e) generates the correct outfit with reasonable shading but with the wrong identity. Without face undistortion, (f) generates a face with more perspective distortion (\ie, exaggerated facial features), zoom in for details. In contrast, the full pipeline (g) yields high-quality full-body selfies. 
}
  \label{fig:ablation_supp}
\end{figure*}

Fig. \ref{fig:ablation_supp} shows additional results of the ablation study, further demonstrating the effectiveness of our final design.

\subsection{Baseline Comparison}

\begin{figure*}[!t]
\centering
{\includegraphics[width=0.12\linewidth]{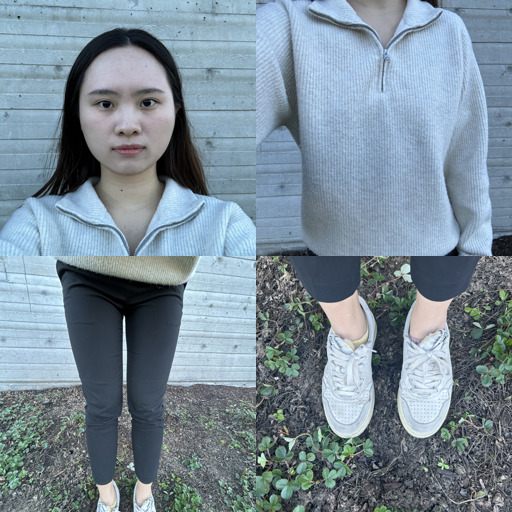}}
{\includegraphics[width=0.12\linewidth]{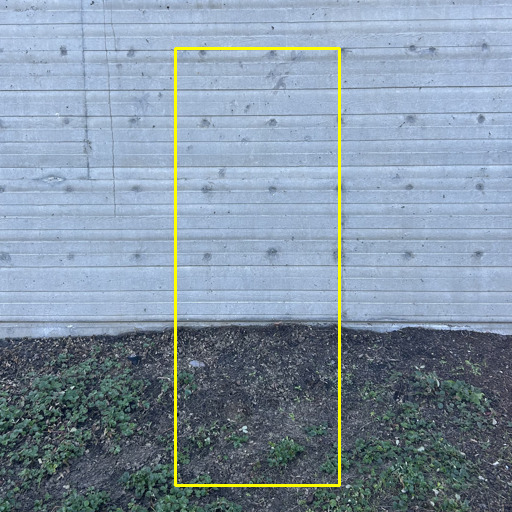}}
{\includegraphics[width=0.12\linewidth]{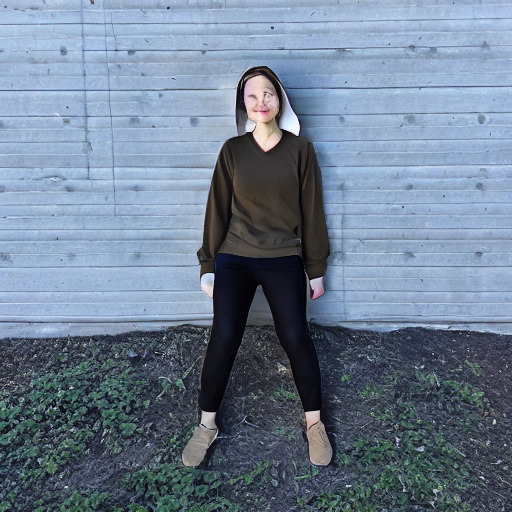}}
{\includegraphics[width=0.12\linewidth]{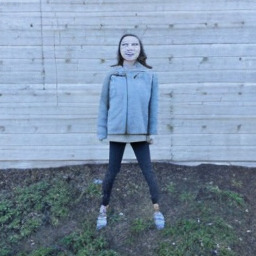}}
{\includegraphics[width=0.12\linewidth]{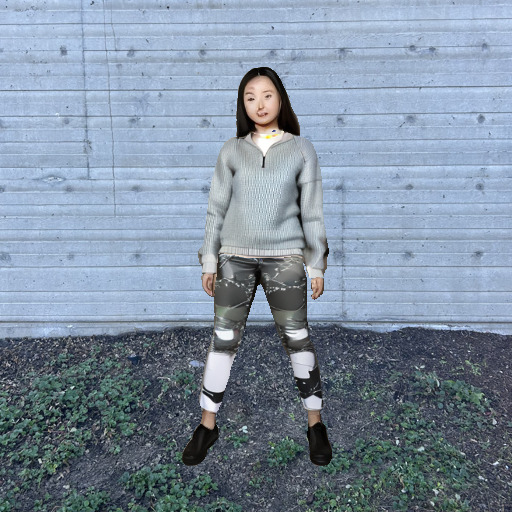}}
{\includegraphics[width=0.12\linewidth]{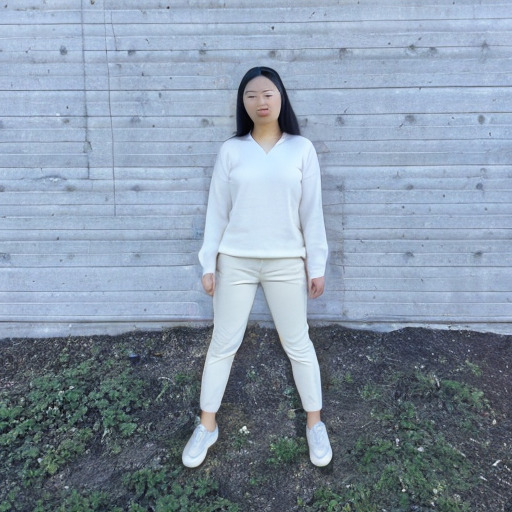}}
{\includegraphics[width=0.12\linewidth]{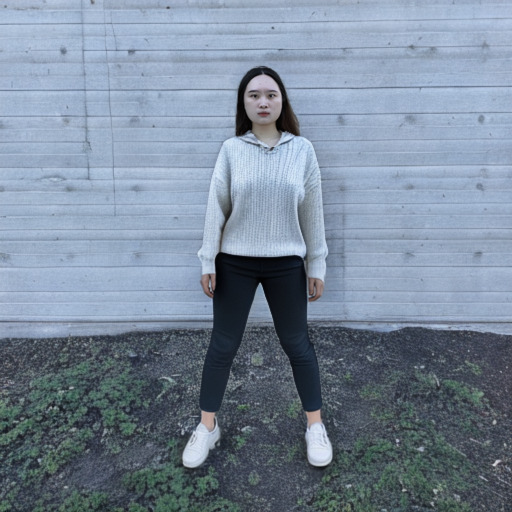}}
{\includegraphics[width=0.12\linewidth]{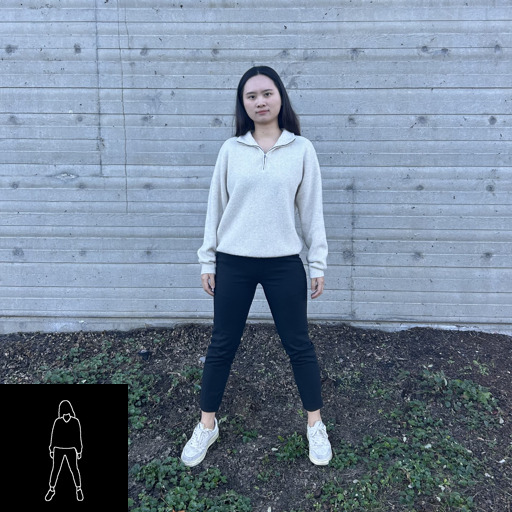}}
\\
{\includegraphics[width=0.12\linewidth]{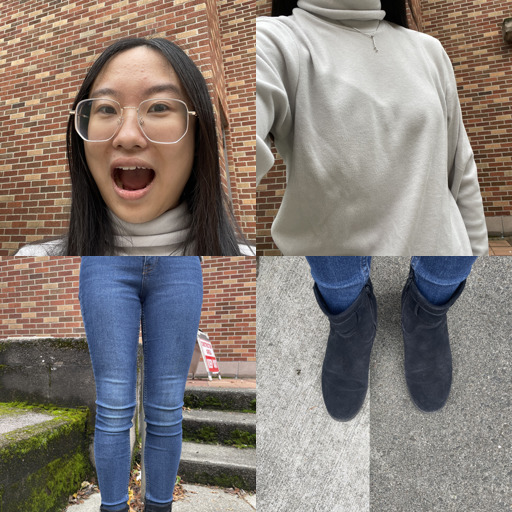}}
{\includegraphics[width=0.12\linewidth]{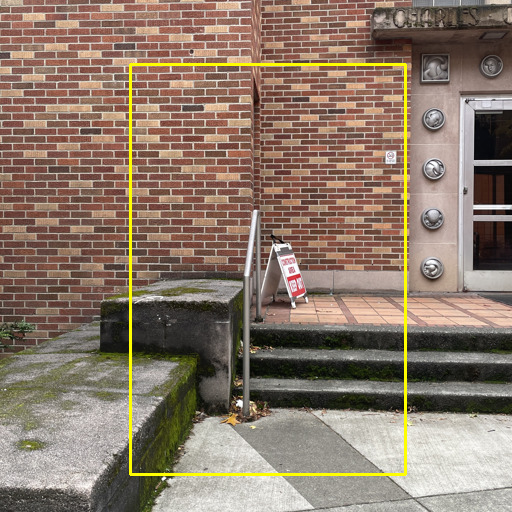}}
{\includegraphics[width=0.12\linewidth]{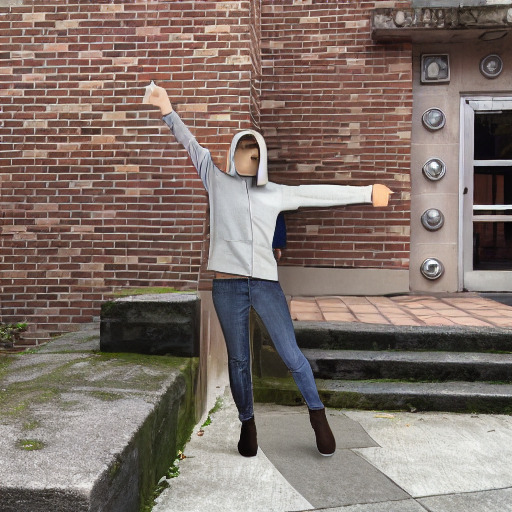}}
{\includegraphics[width=0.12\linewidth]{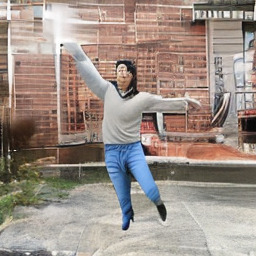}}
{\includegraphics[width=0.12\linewidth]{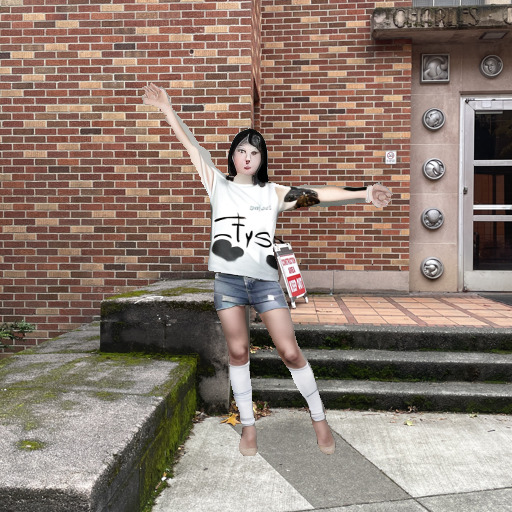}}
{\includegraphics[width=0.12\linewidth]{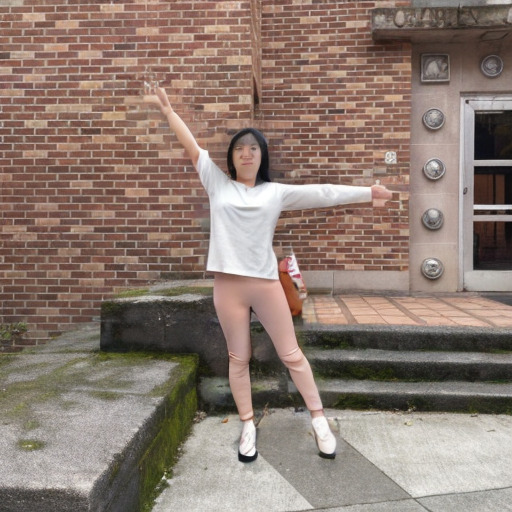}}
{\includegraphics[width=0.12\linewidth]{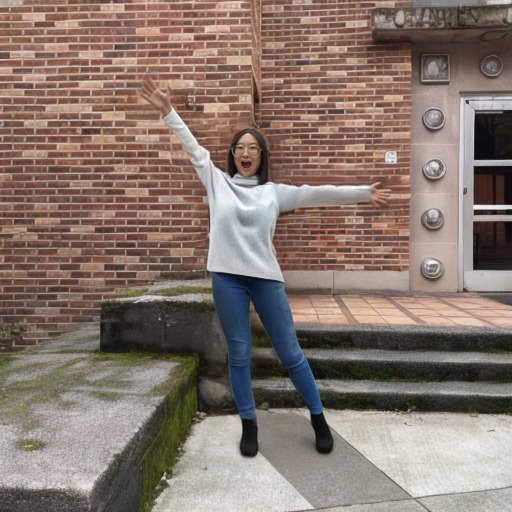}}
{\includegraphics[width=0.12\linewidth]{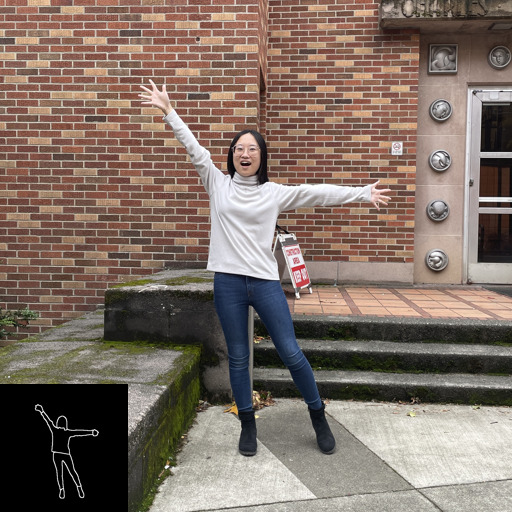}}
\\
{\includegraphics[width=0.12\linewidth]{fig/experiments/comparison/yuqun_ex2/selfies_combined.jpg}}
{\includegraphics[width=0.12\linewidth]{fig/experiments/comparison/yuqun_ex2/scene_bbox.jpg}}
{\includegraphics[width=0.12\linewidth]{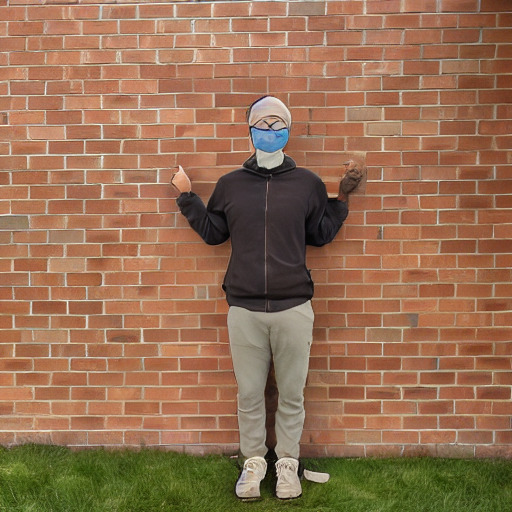}}
{\includegraphics[width=0.12\linewidth]{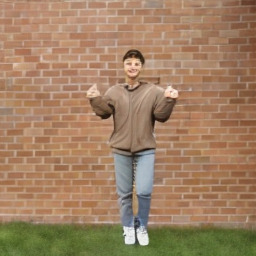}}
{\includegraphics[width=0.12\linewidth]{fig/experiments/comparison/yuqun_ex2/ladi_vton.jpg}}
{\includegraphics[width=0.12\linewidth]{fig/experiments/comparison/yuqun_ex2/dreambooth.jpg}}
{\includegraphics[width=0.12\linewidth]{fig/experiments/comparison/yuqun_ex2/ours.jpg}}
{\includegraphics[width=0.12\linewidth]{fig/experiments/comparison/yuqun_ex2/gt_canny.jpg}}
\\
  \subcaptionbox{ Input Selfies}%
{\includegraphics[width=0.12\linewidth]{fig/experiments/comparison/yuhan_ex1_pose1/selfies_combined.jpg}}
  \subcaptionbox{ Background}%
{\includegraphics[width=0.12\linewidth]{fig/experiments/comparison/yuhan_ex1_pose1/scene_bbox.jpg}}
  \subcaptionbox{PBE}%
{\includegraphics[width=0.12\linewidth]{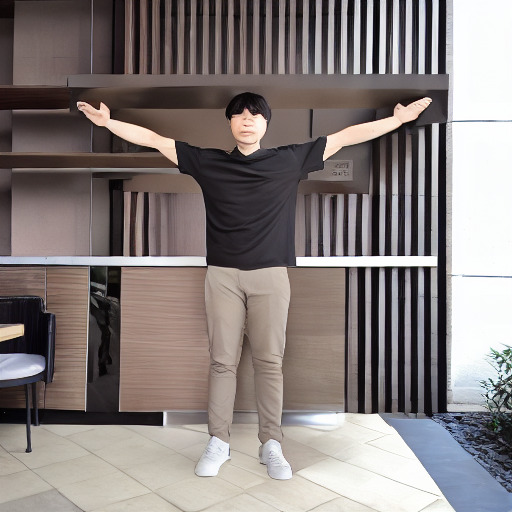}}
  \subcaptionbox{DisCo}%
{\includegraphics[width=0.12\linewidth]{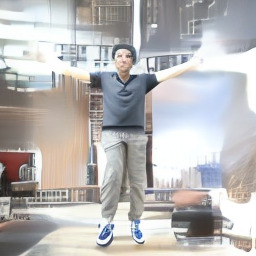}}
    \subcaptionbox{LaDI-VTON}%
{\includegraphics[width=0.12\linewidth]{fig/experiments/comparison/yuhan_ex1_pose1/ladi_vton.jpg}}
    \subcaptionbox{DreamBooth }%
{\includegraphics[width=0.12\linewidth]{fig/experiments/comparison/yuhan_ex1_pose1/dreambooth.jpg}}
    \subcaptionbox{Ours}%
{\includegraphics[width=0.12\linewidth]{fig/experiments/comparison/yuhan_ex1_pose1/ours.jpg}}
    \subcaptionbox{Real Photo}%
{\includegraphics[width=0.12\linewidth]{fig/experiments/comparison/yuhan_ex1_pose1/gt_canny.jpg}}
\caption{
Qualitative comparison with all baselines. For all methods (except for DisCo), we used the Canny Edge of the real photo as the target pose (inset of (h)). For DisCo, we used OpenPose Skeleton of the real photo as the target pose. Our pipeline clearly outperforms baselines in terms of photorealism and faithfulness (zoom in for details, including faces and shoes). Note that, while the selfies, background image, and real photo were captured in the same session, variations in lighting conditions, auto exposure, white balance, and other factors may result in intensity and color tone differences.
}
  \label{fig:supp_comparison}
\end{figure*}

Fig. \ref{fig:supp_comparison} shows a comparison with all baselines. Total Selfie can produce high-quality full-body shots in diverse backgrounds, poses, outfits, and expressions, all while maintaining reasonable shading and composition. 

For the baseline DreamBooth, we use the prompt: ``photo of a full body person, [V] face, wearing [X] top, [Y] bottom, [Z] shoes'', where tokens [$\cdot$] are unique identifiers used to train DreamBooth for a specific concept (body part).

\clearpage
{
    \small
    \bibliographystyle{ieeenat_fullname}
    \bibliography{main}
}
